\documentclass{article}

\usepackage{microtype}
\usepackage{graphicx}
\usepackage{subcaption}
\usepackage{booktabs} % for professional tables

\usepackage{hyperref}
\usepackage{enumitem}

\usepackage[preprint]{icml2026}

\usepackage{amsfonts}       % blackboard math symbols
\usepackage{nicefrac}       % compact symbols for 1/2, etc.
\usepackage{microtype}      % microtypography
\usepackage{xcolor}         % colors
\usepackage{wrapfig}
\usepackage{graphicx}
\usepackage{makecell}
\usepackage{amsmath}
\usepackage{amssymb}
\usepackage{mathtools}
\usepackage{amsthm}

\usepackage{bm}

\usepackage{ulem}
\usepackage[capitalize,noabbrev]{cleveref}
\usepackage{threeparttable}
\usepackage{multicol}
\usepackage{multirow}
\usepackage{amsfonts}
\usepackage{amssymb}
\usepackage{fdsymbol}
\theoremstyle{plain}
\newtheorem{theorem}{Theorem}[section]

\theoremstyle{definition}
\newtheorem{definition}[theorem]{Definition}

\theoremstyle{remark}

\usepackage[textsize=tiny]{todonotes}

\begin{document}

\twocolumn[
  \icmltitle{Beyond All-to-All: Causal-Aligned Transformer with Dynamic Structure Learning for Multivariate Time Series Forecasting}
% Beyond All-to-All: Robust Causal Decomposition for Interpretable Multivariate Time Series Forecasting
% Causal-LLM: Empowering Large Language Models with Dynamic Causal Adapters for Multivariate Time Series Forecasting

  % It is OKAY to include author information, even for blind submissions: the
  % style file will automatically remove it for you unless you've provided
  % the [accepted] option to the icml2026 package.

  % List of affiliations: The first argument should be a (short) identifier you
  % will use later to specify author affiliations Academic affiliations
  % should list Department, University, City, Region, Country Industry
  % affiliations should list Company, City, Region, Country

  % You can specify symbols, otherwise they are numbered in order. Ideally, you
  % should not use this facility. Affiliations will be numbered in order of
  % appearance and this is the preferred way.
  \icmlsetsymbol{equal}{*}

  \begin{icmlauthorlist}
    \icmlauthor{Xingyu Zhang}{ucas,iscas}
    \icmlauthor{Hanyun Du}{ucas,iscas}
    \icmlauthor{Zeen Song}{ucas,iscas}
    \icmlauthor{Siyu Zhao}{ucas}
    \icmlauthor{Changwen Zheng}{ucas,iscas}
    \icmlauthor{Wenwen Qiang}{ucas,iscas}
    %\icmlauthor{}{sch}
    %\icmlauthor{}{sch}
  \end{icmlauthorlist}

  \icmlaffiliation{ucas}{University of Chinese Academy of Sciences}
  \icmlaffiliation{iscas}{Institute of Software, Chinese Academy of Sciences, China}

  \icmlcorrespondingauthor{Wenwen Qiang}{qiangwenwen@iscas.ac.cn}

  % You may provide any keywords that you find helpful for describing your
  % paper; these are used to populate the "keywords" metadata in the PDF but
  % will not be shown in the document
  \icmlkeywords{Machine Learning, ICML}

  \vskip 0.3in
]

% this must go after the closing bracket ] following \twocolumn[ ...

% This command actually creates the footnote in the first column listing the
% affiliations and the copyright notice. The command takes one argument, which
% is text to display at the start of the footnote. The \icmlEqualContribution
% command is standard text for equal contribution. Remove it (just {}) if you
% do not need this facility.

% Use ONE of the following lines. DO NOT remove the command.
% If you have no special notice, KEEP empty braces:
\printAffiliationsAndNotice{}  % no special notice (required even if empty)
% Or, if applicable, use the standard equal contribution text:
% \printAffiliationsAndNotice{\icmlEqualContribution}

\begin{abstract}
  % Most existing multivariate time series forecasting methods adopt an all-to-all paradigm that feeds all variable histories into a unified model to predict their future values without distinguishing their individual roles. However, this undifferentiated paradigm makes it difficult to identify variable-specific causal influences and often entangles causally relevant information with spurious correlations. To address this limitation, we propose an all-to-one forecasting paradigm that predicts each target variable separately. Specifically, we first construct a Structural Causal Model from observational data and then, for each target variable, we partition the historical sequence into four subsegments according to the inferred causal structure: endogenous, direct causal, collider causal, and spurious correlation. The prediction relies solely on the first three causally relevant subsegments, while the spurious correlation subsegment is excluded. Furthermore, we propose Causal Informed Transformer (CDT), a novel forecasting model comprising three components: Endogenous subsegment Prediction Block, Direct Causal subsegment Prediction Block, and Collider Causal subsegment Prediction Block, which process the endogenous, direct causal, and collider causal subsegments, respectively. Their outputs are then combined to produce the final prediction. Extensive experiments on multiple benchmark datasets demonstrate the effectiveness of the CDT.
Most existing multivariate time series forecasting methods adopt an all-to-all paradigm that feeds all variable histories into a unified model to predict their future values without distinguishing their individual roles. However, this undifferentiated paradigm makes it difficult to identify variable-specific causal influences and often entangles causally relevant information with spurious correlations. To address this limitation, we propose an all-to-one forecasting paradigm that predicts each target variable separately. Specifically, we first construct a Structural Causal Model from observational data and then, for each target variable, we partition the historical sequence into four subsegments according to the inferred causal structure: endogenous, direct causal, collider causal, and spurious correlation. Furthermore, we propose the Causal Decomposition Transformer (CDT), which integrates a dynamic causal adapter to learn causal structures initialized by the inferred graph, enabling correction of imperfect causal discovery during training. Furthermore, motivated by causal theory, we apply a projection-based output constraint to mitigate collider induced bias and improve robustness. Extensive experiments on multiple benchmark datasets demonstrate the effectiveness of the CDT.
\end{abstract}

% 摘要部分是我们该怎么建模，做到那三点要求
% 动机部分是我们只能这么建模，一才能因果可解释
% 那介绍部分应该写：预测任务->建模挑战->结构分析->方法
% Step1 从任务出发：为什么MTSF本质上是变量间依赖建模问题
% Step2 挑战+现有方案的问题（点出all-to-all架构的问题，引出我们需要按变量功能进行分解）
% Step3 引出四段划分思想：用天气预测的例子引出“我们不能让这些不同结构的变量混在一起被建模”。
% Step4 SCM 结构推导：为什么这种划分方式是唯一合理的。这是逻辑闭环的关键一环，承接“我们要这么做” → “而且只能这么做”。
%存在问题：1.当前Introduction是从现有Transformer-based模型设计出发说的，而不是从MTSF建模目标出发的。导致因果结构的需求显得是“技术性改进”，而非“任务驱动的必然要求”。
% 

\section{Introduction} \label{sec:1}
% 第一段：介绍背景、引出问题
% 逻辑：背景：多元时间序列预测重要性\to 问题：根据多个变量的历史值预测未来值\to 难点：变量关系复杂\to 影响：未建模相关变量会导致信息损失，限制模型学习正确的变化规律；错误建模无关变量会引入虚假关联，影响模型鲁棒性。\to 关键挑战：如何建模变量间关系。
% 中文：（背景）多元时间序列预测是能源消耗、经济规划、气象预测和交通预测等领域的基础问题。
%第一段逻辑：
%接着现有的，写MTSF的问题(what)的定义
%transformer-based 方法成为了解决MTSF的一种有效方法（how），简单介绍一下怎么做的(输入，结构，输出)。
%简单总结一下Transformer方法的优势（why）。
% （MTSF定义，what）MTSF的目标是，根据多个变量的历史观测值，预测其未来的演变轨迹。。
% （怎么做，how）近年来，基于Transformer的方法成为MTSF的重要研究方向。具体而言，这类方法首先编码多个变量的历史轨迹，通过自注意力机制计算不同时间点、时间片段或变量之间的相似性，捕捉长程依赖和跨变量交互特征，随后基于这些动态相关性生成未来序列的预测结果。
% （why）Transformer方法的优势
% Multivariate Time Series Forecasting (MTSF) is a fundamental problem in various fields, including energy consumption \citep{TSF2ECL}, economic planning \citep{TSF2finance2}, weather prediction \citep{TSF2weather}, and traffic forecasting \citep{TSF2traffic}. The goal of MTSF is to predict the future values of multiple interrelated variables based on their historical observations \citep{box2015time}. Unlike univariate time series forecasting, MTSF must capture not only individual temporal patterns but also the interactions among multiple interdependent variables. This makes it crucial to identify which variables influence the target and how. Failing to distinguish relevant from irrelevant inter-variable dependencies properly could either result in information loss or introduce spurious correlations that degrade forecasting performance. 
Multivariate Time Series Forecasting (MTSF) is a fundamental problem in various fields, including energy consumption \citep{TSF2ECL}, economic planning \citep{TSF2finance2}, weather prediction \citep{TSF2weather}, and traffic forecasting \citep{TSF2traffic}. The goal of MTSF is to predict the future values of multiple interrelated variables based on their historical observations \citep{box2015time}. Unlike univariate forecasting, MTSF must capture not only individual temporal patterns but also the interactions among multiple interdependent variables. This makes it crucial to identify which variables influence the target and how. Failing to distinguish relevant from irrelevant inter-variable dependencies properly could either result in information loss or introduce spurious correlations that degrade forecasting performance.

% 第二段 all2all的问题
% 对不同变量历史权重不等给出直观的例子
Based on the above statement, an MTSF method should be capable of capturing the intrinsic temporal patterns of each variable, correctly identifying how other variables causally influence the target, and eliminating spurious correlations that obscure true dependence. However, most existing MTSF methods overlook this structural heterogeneity, including Transformer-based models \citep{TransformersinTS, tang2023infomaxformer}. These models typically take all variables’ histories without differentiation, and train a single shared model to jointly forecast all targets in one forward pass \citep{timerxl, NSTransformer}. This design, whether channel-independent \citep{PatchTST} or channel-mixed \citep{iTransformer}, makes no distinction in contribution among variables, ignoring the distinct causal roles they may play with respect to the target. Although this all-to-all design is easy to implement, it overlooks an important observation: when focusing on forecasting a specific target variable, the history segments of different variables often play very different roles. For instance, in weather forecasting, temperature and humidity are influenced by wind direction, yet they have no direct causal relationship. Meanwhile, temperature and atmospheric pressure jointly affect precipitation, forming a collider structure: $\text{temperature} \to \text{precipitation} \gets \text{pressure}$ \citep{wilks2011statistical}. When forecasting temperature, different variables influence in distinct ways: (i) temperature’s own past provides an autoregressive signal; (ii) wind direction exerts a direct causal influence by determining the inflow of warm or cold air; (iii) through the collider structure, precipitation activates conditional dependence between temperature and pressure; and (iv) humidity appears correlated with temperature only through their shared cause wind direction, but becomes independent once wind direction is conditioned on. Feeding all of these histories indiscriminately into an all-to-all model conflates true causal drivers with spurious signals, leading to noisy attention weights, entangled parameter learning, and ultimately degraded forecasting performance.

% 第三段 提出all2one策略
%为了解决上述挑战，本文提出了一种新的基于all-to-one的MTSF策略，即逐一对目标变量的未来变化进行预测。这样做的优势在于我们可以显示的建模历史片段的不同部分对目标变量未来的影响。接下来，我们回答怎么建模历史片段和目标变量未来变化之间的作用关系。我们首先考虑的是将历史片段划分为不同的子片段，子片段内不同变量的历史对未来变量的影响是同性的，不同子片段对未来变量的影响各部相同。整体而言，我们将整个历史片段按照变量的不同划分为4类：1）内生性子片段，其由目标变量的历史组成；2）具有直接因果关系的外生性子片段，其是由其他变量的历史组成，但从结构因果模型（SCM）看,这些其他变量和目标变量具有直接格兰杰因果关系。3）虚伪相关性的外生性子片段，其是由其他变量的历史组成，但从结构因果模型（SCM）看,在给定2)中的其他变量时，这些其他变量和目标变量条件独立；4）具有间接因果关系的外生性子片段，其是由其他变量的历史对组成，每一个对包含两个变量的历史，然后每一个对的变量和目标变量构成一个对撞结构，例如，Figure \ref{fig:mb}中的${V_i} \to {V_c} \leftarrow {V_s}$结构.从因果的角度看，往往假设内生变量和外生变量相互独立，因此我们得到1）。然后不同变量间的因果关系包含:直接因果相关、节间因果相关及虚伪相关，因此我们得到2）、3）及4）。与此同时，我们通过理论分析表明，将单独考虑是有助于提升模型的泛化性。接着，对于1）、2）及4）子片段，我们通过一个条件概率的形式对于每个子片段进行建模，条件概率表述为$P（目标变量的未来|子片段中的历史）$。对于3）子片段，由于建模它会引入虚伪相关性，因此我们直接将之剔除。最终，我们将上述所有子片段的结果进行融合，得到最终的结果。
To address these challenges, we adopt an all-to-one forecasting paradigm, where each target variable is predicted individually with an explicit structural decomposition of its historical window. Specifically, guided by the d-separation criterion in structural causal models (SCMs), we partition the multivariate history for each target variable into four sub-segments: (1) Endogenous Subsegment (ES), the target's own history; (2) Direct Causal Subsegment (DCS), histories of variables that are directly connected to variable in the causal graph; (3) Collider Causal Subsegment (CCS), histories involved in collider patterns with the target variable; and (4) Spurious Correlation Subsegment (SCS), histories that become independent of the target variable after conditioning on the relevant segments. 
To operationalize this decomposition, we learn a target specific DAG from observational training data using the Peter-Clark (PC) algorithm~\citep{spirtes2001causation}, from which we derive the direct causal and collider causal variable sets for segment construction. To ensure robustness to inevitable estimation errors, we further introduce a learnable dynamic causal adapter that treats the learned DAG as a structural prior and refines during training.

% (范式\to 具体方法)
% 第四段 方法 
% 逻辑：（承上）基于范式提出方法\to （方法）模型分为三部分\to 每个部分做什么、怎么做\to 效果
% 中文：（承上）基于这个范式，我们提出了一个多变量时间序列预测方法——Causal Informed TransFormer（CDT）。
% （方法）CDT三个部分组成:target-variable Prediction Block(TPB), Direct causal variables Prediction Block(DPB), and Indirect auxiliary variables Prediction Block(IPB)。
% （1 做什么）TPB仅利用目标变量自身的历史序列建模其沿时间的变化规律，并用于未来预测。
% （1 怎么做）具体而言，TPB将同一变量在时间维度上相邻的时间点划分为同一个Patch，并通过Multi-Patch Attention机制计算patch之间的相似性，从而建模变量内沿时间的变化规律。
% （2 做l什么）DPB建模直接因果变量如何影响目标变量的变化。
% （2 怎么做）具体地,DPB采用带掩码的多变量Attention机制，使模型能够捕捉直接因果变量对目标变量的影响。
% （3 做什么）IPB建模间接辅助变量如何影响目标变量的预测。
% （3 怎么做）IPB同样采用带掩码的多变量Attention机制，利用间接辅助变量提供的额外信息，增强对目标变量的预测能力。
% （融合）最后，模型通过动态融合模块整合三部分的预测结果，以优化最终预测并平衡各部分的贡献。
% （效果）在多个基准数据集上的实验结果表明，CDT 在预测精度上优于现有方法，并通过消融实验验证了三个模块及动态融合机制对模型性能的影响。
Building on the above, we propose the Causal Decomposition Transformer (CDT), a target specific forecasting framework that aligns model computation with the causal roles of historical segments of other variables. For each target, CDT constructs three complementary contexts: an endogenous context capturing the target’s own dynamics, a direct causal context summarizing the influence of directly related variables, and a collider causal context accounting for collider induced dependencies. These contexts are processed in a structured manner with attention constraints that respect their causal roles, ensuring that multivariate information is incorporated without entangling spurious signals. Finally, motivated by generalization analysis, CDT applies a causal output refinement to suppress collider induced spurious dependence and improve robustness.

\textbf{Our contributions:} \textbf{1}) We propose an all-to-one forecasting paradigm for MTSF, in which each target variable’s future is predicted individually; \textbf{2}) We propose CDT, which refines the structural prior, incorporates causal contexts, and introduces an output spouse projection motivated by theory to mitigate collider induced spurious dependence; \textbf{3}) Extensive experiments and ablation studies on multiple benchmark datasets demonstrate that CDT achieves superior predictive accuracy, robustness, and interpretability compared to existing methods.

\section{Related Works} 
\paragraph{Multivariate Time Series Forecasting}
With the development of deep learning \citep{lecun_deep_2015}, numerous models have been proposed for MTSF \citep{hu2022time, TransformersinTS}, including CNN-based \citep{zhan2023differential}, RNN-based \citep{how_RNN2TSF,tang2021building}, MLP-based \citep{zhang2024not}, and Transformer-based \citep{zhang2024intriguing} architectures. These approaches are all based on all‑to‑all strategy and can be broadly categorized based on their modeling focus into three groups: temporal-domain, frequency-domain, and variable-domain methods. Temporal domain methods, such as PatchTST \citep{PatchTST} and TimesNet \citep{TimesNet}, focus on intra-variable dependencies by modeling patch-wise or point-wise relations. Frequency domain methods, such as FEDFormer \citep{FEDformer} and FreDF \citep{zhang2024intriguing}, transform sequences into the Fourier domain to capture frequency-specific dynamics. Variable domain methods, such as iTransformer \citep{iTransformer}, model inter-variable dependencies via attention mechanisms, and TimeXer \citep{timexer} proposes to treat variables' endogenous and exogenous signals differently.

% \paragraph{Large Language Models for TSF}
% With the rapid progress of foundation models, LLM-based approaches for TSF have surged. Early works like PromptCast \citep{PromptCast} and LLMTime \citep{LLMTime} treat time series as numerical text tokens, leveraging zero-shot capabilities. Subsequent methods focus on reprogramming or fine-tuning: FPT \citep{FPT} freezes the LLM and fine-tunes the input embedding; Time-LLM \citep{Time-LLM} aligns time series features with the text space via reprogramming layers. Despite their success, most existing methods adopt a Channel-Independent (CI) strategy \citep{Autotim}, processing each variable in isolation. This inherently discards inter-variable dependencies. While some recent works attempt to mix channels \citep{timerxl}, they typically employ all-to-all attention mechanisms, which lack structural inductive bias and struggle to distinguish between causal drivers and spurious correlations.

\paragraph{Modeling Variable Relationships in MTSF}
Capturing the complex interactions among variables is central to MTSF. Existing approaches exhibit primary modeling paradigms: Temporal-based methods \citep{TimesNet, PatchTST} focus on intra-variable temporal patterns by analyzing relationships between time points or segments; Frequency-based methods \citep{zhang2024not, FEDformer} decompose temporal patterns through spectral transformations, but remain limited to single-variable analysis; Variable-based methods \citep{iTransformer} attempt to capture cross-variable interactions through attention mechanisms. However, these methods often perform unconstrained pairwise computations which may conflate causal relationships with spurious correlations, lacking explicit mechanisms to distinguish different types of inter-variable dependencies. The limitations of these approaches reveal a gap: current methods either oversimplify cross-variable correlation or naively aggregate all potential interactions without causal discrimination. Therefore, effectively modeling inter-variable relationships in MTSF remains an open research problem.

\paragraph{Causal Discovery for MTSF}
Causal discovery aims to identify genuine cause-effect relationships from observational data by learning directed acyclic graphs (DAGs) \citep{glymourCausalInferenceStatistics2016,pearlCausality2009, Causal23Gong, lauritzen1989graphical}. Early constraint-based methods, such as the Inductive Causation algorithm \citep{verma1990causal} and the Peter-Clark algorithm \citep{spirtes1991algorithm}, rely on conditional independence tests to reconstruct DAG. Subsequent work improved efficiency \citep{spirtes2001causation} and robustness to latent variables \citep{spirtes2001anytime}. In time series settings, causal discovery faces additional challenges. Extensions such as tsFCI \citep{entner2010causal} adapt FCI to time-lagged data, while Granger causality infers causal relations based on predictive precedence \citep{grangerInvestigatingCausalRelations1969}. More recently, causal discovery has been combined with multivariate time series forecasting (MTSF) mainly through two paradigms: (1) causal Markov–based preprocessing to remove spurious correlations \citep{li2021causal}, and (2) proxy-variable methods that recover latent confounders to infer causal structures \citep{liu2023causal}. Representative models include CausalFormer \citep{CausalFormer}, which jointly learns temporal dynamics and a Granger-causal graph, and Causal-TSF \citep{CausalTSF}, which estimates latent confounders and applies causal interventions during forecasting. However, existing approaches typically treat causal discovery as a preprocessing step, regularizer, or bias-correction module, rather than embedding it directly into model parametrization. Consequently, causal information is not dynamically enforced during forecasting.

\begin{figure*}
   \centering
   \subfloat[\footnotesize ETTh1\label{f1a}]{
    \includegraphics[width=0.15\linewidth]{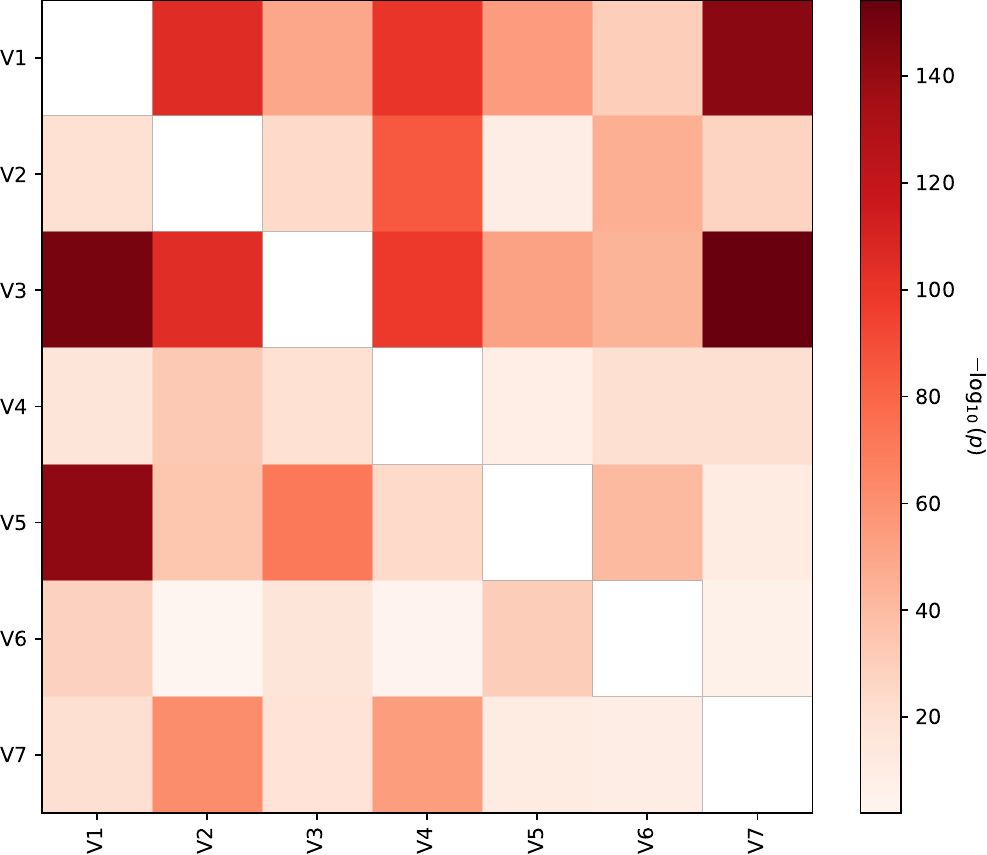}
    }
    \subfloat[\footnotesize ETTm1\label{f1b}]{
        \includegraphics[width=0.15\linewidth]{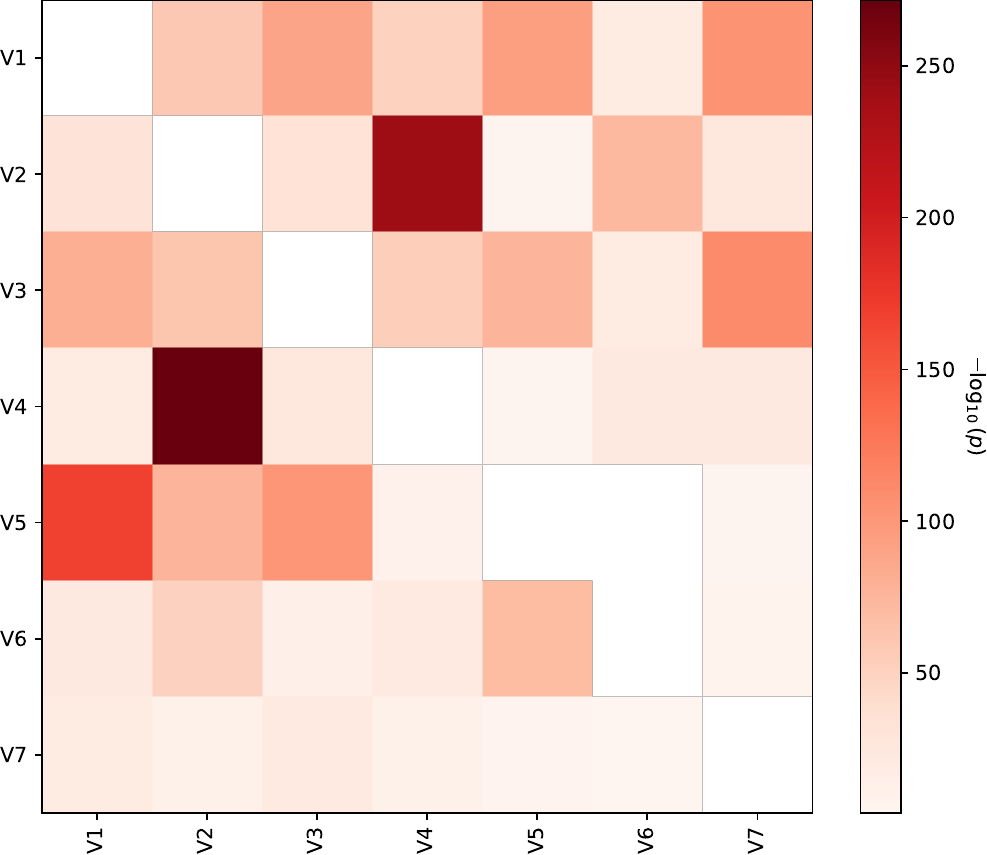}
    }
    \subfloat[\footnotesize Exchange\label{f1c}]{
        \includegraphics[width=0.15\linewidth]{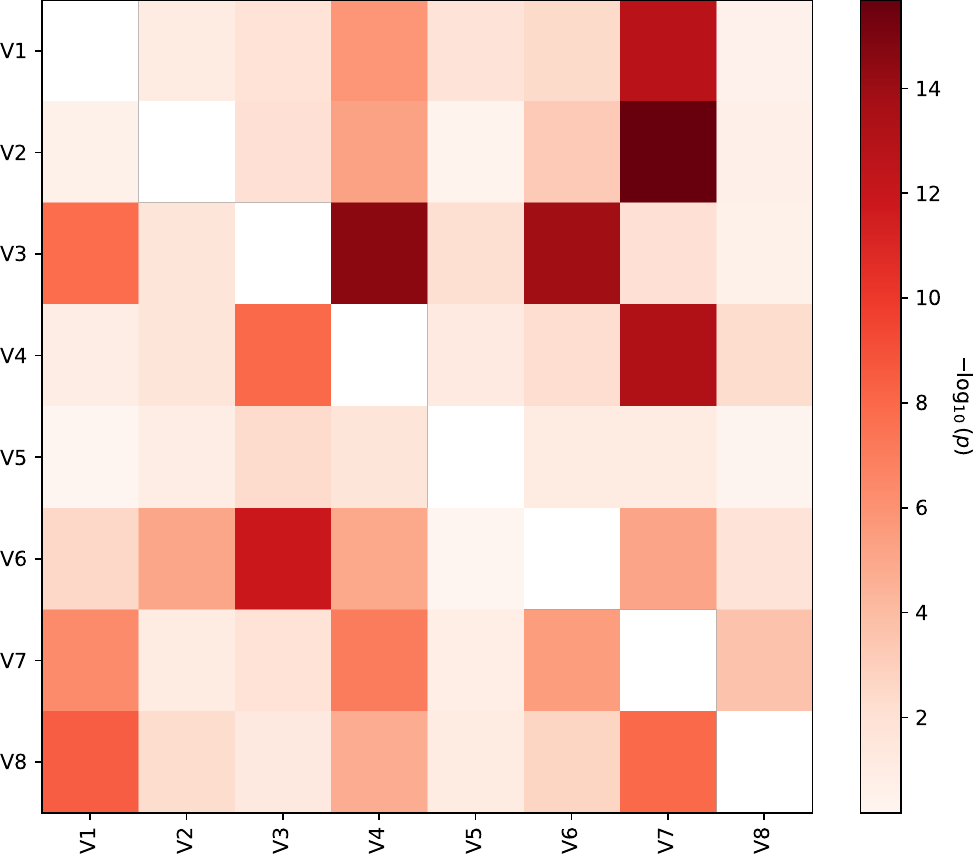}
    }
    \subfloat[\footnotesize Path\label{f1d}]{
        \includegraphics[width=0.341\linewidth]{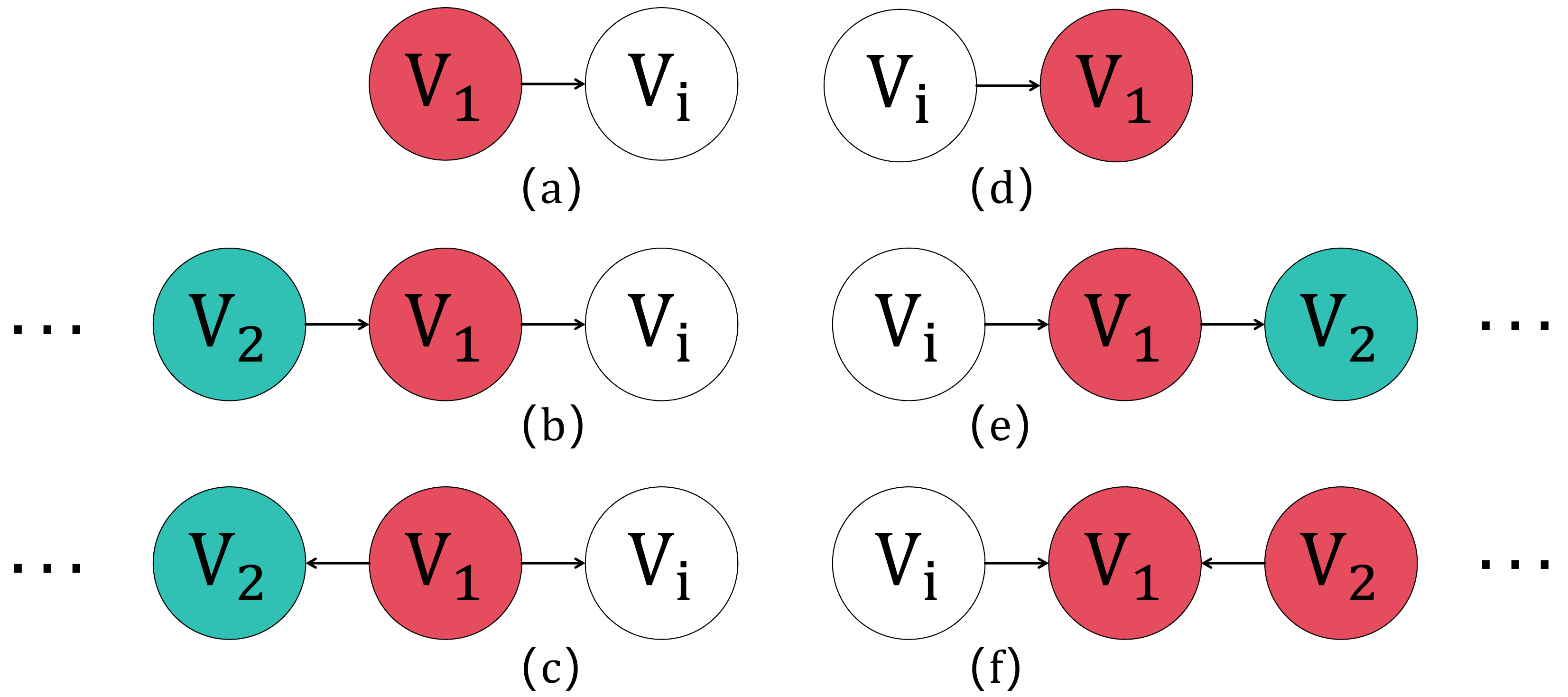}
    }
   \caption{(a)-(c) Visualization of Granger causality across variables in ETTh1, ETTm1, and Exchange datasets. Each heatmap shows the transformed causal strength matrix using $-\log(P)$ values, where a darker color indicates a stronger causal influence from the row variable to the column variable. Diagonal entries are masked. (d) Representative partial SCM commonly encountered in MTSF. White nodes represent the target variable $V_i$, red and green nodes represent causally related variables and spurious correlated variables separately.}
   \label{fig:motivation}
\end{figure*}

\section{Causal Analysis and Motivation} \label{Sec:Motivation}
% In this section, 首先从因果分析的角度回答为什么我们要探索不同历史片段对目标变量的影响。其次，基于所提供的因果分析，我们给出建模MTSF的洞察。
%提出的建模MTSF的策略的动机。其次，从因果的角度，再次给出建模策略中关于其他变量和目标变量关系的分析。接着，详细介绍所提出的建模策略的组成成分。最后，通过理论分析给出建模策略中组成成分的合理性分析。
%在motivation的第二段，我们解释了为什么要考虑不同变量对目标变量的影响。
In this section, we first present some notations. Then, we explain from a causal analysis perspective why it is essential to separate the influence of different historical segments on the target variable. At last, we convert these causal analyses into concrete modeling guidelines for our MTSF framework.

\subsection{Notation and Problem Definition} \label{Sec:Motivation.1}
Let $ X = [ x_1, \cdots, x_T ] \in \mathbb{R}^{T \times D} $ be a historical sequence with $ T $ time steps and $ D $ variables. At each timestamp $ t \in \left\{ {1,\cdots,T} \right\} $, the state is represented as $ x_t = [ V^t_1, \cdots, V^t_D ] \in \mathbb{R}^D $, where $V^t_i \in \mathbb{R}$ is the observed value of the variable $V_i$ at time step $t$. Let $ Y = [ x_{T+1}, \cdots, x_{T+S} ] \in \mathbb{R}^{S \times D} $ be the future sequence with $ S $ time steps. Given a training dataset $ D_{\text{train}} = \{ (X^i, Y^i) \}_{i=1}^K $, where $K$ is the number of training samples, $X^i$ represents the $i$-th historical sequence, and $Y^i$ is its corresponding future sequence. The learning process of MTSF can be formalized as finding an optimal predictor $ f^* $ within a hypothesis space $ \mathcal{F} $, such that $f^*(X) = Y$. Specifically, the forecasting model is learned by solving the following empirical risk minimization problem:
\begin{equation} \label{qweqweffd}
f^* = \arg\min_{f \in \mathcal{F}} \frac{1}{K} \sum\nolimits_{i=1}^{K} \mathcal{L}(Y^i, f(X^i)),
\end{equation}
where $ \mathcal{L} (\cdot)$ denotes the loss function, e.g., the MSE loss. As shown in Equation (\ref{qweqweffd}), the learning process of $f^*$ doesn't constrain correlations among variables in the input.

As discussed in Section \ref{sec:1}, when analyzing model predictions from the perspective of a specific target variable, all-to-all based strategies may inadvertently encode spurious correlations between variables. This can lead to inaccurate forecasting and significantly degrade the model’s generalization ability. This issue is further substantiated by the empirical analysis presented in Section \ref{Sec:Motivation.3}. To address this challenge, the key lies in understanding how each variable's historical values contribute to target variable future evolution. We propose a segmentation strategy based on semantic consistency: the historical sequences of variables that influence the target variable through similar causal mechanisms are grouped into the same subsegment, while segments with distinct mechanisms are separated. In following section, we elaborate on the rationale and theoretical foundations behind this segmentation approach.

\subsection{Why Divide History into Endogenous and Exogenous Components?} \label{Sec:Motivation.2}
% 对于任意一个 MTSF 目标变量，其历史序列可视为动态系统的离散采样，本身已经蕴含了描述该变量自我演化所需的全部信息。我们真正关心的是：其他变量的历史片段究竟如何调制这种演化。若不把 内生片段（目标变量自身的滞后）与 外生片段（其他变量的滞后）明确区分，那么任何观测到的预测改进都会被目标变量的惯性所混淆。结果是——我们只能知道目标变量发生了变化，却无法判断这变化是外生片段驱动的，还是仅仅源于其自身的历史。因此，想要做出清晰的归因分析，必须显式地区分内生与外生历史片段。
For any target variable in MTSF, its own history can be regarded as a discrete sampling of an underlying dynamical system \citep{takens1981detecting}, encoding all the information needed to describe the variable’s intrinsic evolution \citep{timexer}. What we really want to uncover is how the history of other variables influences that evolution. If we do not explicitly separate the endogenous subsegment from the exogenous subsegment, any apparent improvement in prediction will be confounded by the history of the target variable itself. We will observe that the target changes, but cannot determine whether that change is driven by external histories or merely by its own history. Clear causal attribution therefore, demands an explicit distinction between endogenous and exogenous historical subsegments.

\subsection{Why Further Subdivide the Exogenous Segment?} \label{Sec:Motivation.3}
Having explicitly distinguished the endogenous and exogenous subsegments, we next examine the structural heterogeneity within the exogenous subsegment. According to \citep{pearlCausality2009}, causal relationships between variables can be classified as either direct or indirect. From the SCM perspective, a direct causal relationship indicates an immediate connection between two variables, whereas an indirect causal relationship involves one or more intermediate nodes. As noted in Appendix \ref{background_causal}, two variables that are indirectly connected may become independent once we condition on intermediate variables. If we indiscriminately use all variables’ histories to predict the target, variables that are not causally relevant may nonetheless exhibit apparent predictive associations with the target in the learned model, even though these associations do not correspond to genuine causal influence, i.e., spurious correlations. Therefore, the exogenous subsegment admits a meaningful conceptual subdivision into causally relevant components and spurious correlation components, which motivates a more fine-grained treatment rather than indiscriminate modeling.

To empirically illustrate the necessity of treating the exogenous subsegment in a more fine-grained manner, we conduct an exploratory analysis on the ETTh1, ETTm1, and Exchange-rate datasets. We employ Granger causality analysis, a widely used statistical test in time series analysis, to examine predictive dependencies between variables, rather than to establish structural causal relations. Specifically, if past values of one variable improve the forecasting accuracy of a target variable under the Granger framework, we regard its historical observations as predictively informative for that target. Conversely, variables whose histories do not exhibit such predictive utility are treated as weakly informative in this analysis. In contrast, modeling the history of such variables may introduce potentially spurious associations without improving predictive accuracy, motivating their separation from predictively relevant components. 

Granger causality analysis outputs a $P$-value indicating statistical significance. To better visualization, we apply a $-\log(P)$ transformation and present the results as a heatmap in Figure \ref{f1a}-\ref{f1c}. In the heatmap, the cell at row $m$ and column $n$ represents causal influence from the $m$-th variable to the $n$-th variable, with darker colors corresponding to stronger influences. Diagonal elements are filled with a uniform color to exclude self-influence. As Figure~\ref{fig:motivation} shows, some variables exhibit strong predictive influence on others, while many variable pairs display negligible predictive dependency. This observation highlights the pronounced sparsity and heterogeneity of inter-variable influences, underscoring the necessity of distinguishing genuinely informative dependencies from non-informative ones to mitigate spurious correlations and improve predictive performance. In contrast, most existing methods either treat variables independently or aggregate them indiscriminately, lacking explicit mechanisms to model such heterogeneous dependency roles, which limits their ability to exploit structured inter-variable information for forecasting.

\subsection{How should the exogenous segment be further subdivided?} \label{Sec:Motivation.4}
To precisely distinguish genuine causal relationships from spurious ones, we systematically analyze all relevant causal pathways connected to the target variable $V_i$ and construct a local SCM centered around it. Based on the causal analysis in Appendix \ref{background_causal}, the relationship between a non-target variable and the target $V_i$ in Fig.~\ref{f1d} can be treat separately:

First, $V_1$ can be an ancestor of $V_i$, including the direct-parent case and more general upstream structures such as chains and forks (e.g., $V_2 \to V_1 \to V_i$ or $V_2 \gets V_1 \to V_i$, corresponding to Paths (a)–(c)). In all these cases, $V_1$ lies on a directed path into $V_i$, and conditioning on all remaining variables $\mathcal{Z} = \{V_1,\dots,V_{i-1},V_{i+1},\dots,V_D\}$ leaves $V_1$ dependent on $V_i$, while any other $V_j \in \mathcal{Z}\setminus\{V_1\}$ becomes conditionally independent of $V_i$. 

Second, $V_1$ can be a descendant of $V_i$, including direct children and any downstream chain (corresponding to Paths (d)–(e)). This situation is symmetric: $V_1$ lies on a directed path out of $V_i$, and after conditioning on $\mathcal{Z}$, only $V_1$ remains conditionally dependent on $V_i$, whereas all other $V_j \in \mathcal{Z}\setminus\{V_1\}$ are independent of $V_i$.

Finally, $V_1$ can act as a collider on a path from $V_i$ to another variable $V_2$ (Path (f)), e.g., $V_i \to V_1 \gets V_2 - \cdots - V_D$. In this case, conditioning on $\mathcal{Z}$ makes both $V_1$ and its spouse $V_2$ dependent on $V_i$, while any $V_j \in \mathcal{Z}\setminus\{V_1,V_2\}$ is conditionally independent of $V_i$. 
Formal proofs of the conditional independencies asserted for each path in Appendix \ref{proof4Group}.

Without loss of generality, consider any SCM defined over a set of variables $\{V_1, \cdots, V_D\}$. For any target variable $V_i$ and the local SCM relevant to $V_i$ can be represented by the combination of elements in $\{${Path a}, $\cdots$, {Path f}$\}$. Then, we can identify a subset of variables that are conditionally dependent on the $V_i$ and eliminate other independent variables. Specifically, the simplified SCM includes: 1) \textbf{Direct Parents}: Variables that have a direct causal influence on $V_i$, denoted as $V_p \to V_i$;
2) \textbf{Direct Children}: Variables that are directly influenced by $V_i$, denoted as $V_i \to V_k$ and each $V_k$ is not a collider;
3) \textbf{Collider Structures}: Variables that form collider structure involving $V_i$, e.g., $V_i \to V_c \gets V_s$, where $V_c$ is the collider and $V_s$ denotes the spouse variables.

Ultimately, for each variable, we partition the exogenous segment into three subsegments: 1) Direct Causal subsegment (DCS): including all variables that are direct parents or direct children of the target variable, representing direct causal affect on target variable; 2) Collider Causal subsegment (CCS): consisting of variables that, together with the target, form collider patterns such as $V_i \to V_c \leftarrow V_s$, where $V_c$ is the collider node; 3) Spurious Correlation subsegment (SCS): comprising every remaining variable that is not part of the direct parent, direct child, or collider structures; these variables do not reflect genuine causality with the target variable. In the next section, we further elaborate on why causal relationships should be distinguished specifically between DCS and CCS to enhance predictive generalization.

\subsection{Why Are Causal Relationships Divided into DCS and CCS?}\label{Sec:Motivation.5}
While both DCS and CCS are causally relevant to the target, we treat CCS as a separate component because its collider induced dependencies behave differently in prediction: they do not directly affect the target, but can affect generalization if not properly constrained. As we show below, isolating CCS enables us to enforce a conditional independence constraint that reduces the generalization gap.

Consider any variable pair $(V_{c,j},V_{s,j})$ that constitutes a collider structure $ V_i \to  V_{c,j} \leftarrow V_{s,j} $ within the set $\{V_{c,j},V_{s,j}\}_{j=1}^m$,  where $m$ denotes the number of colliders. Both $V_{c,j}$ and $V_{s,j}$ consist of only one variable. Then, the optimal predictor $f^*_{\rm IP}$ under MSE loss for the future values of $V_i$ is defined as:
\begin{equation}\label{optimal_regressor}
%\resizebox{0.99\linewidth}{!}{$
  V^{T:T+S}_i = f^*_{\rm IP}(V^{0:T}_{c,j}, V^{0:T}_{s,j})= \mathbb{E}[V^{T:T+S}_i \vert V^{0:T}_{c,j}, V^{0:T}_{s,j}].
%$}
\end{equation}
For notational simplicity, we denote the history $V^{0:T}_{c,j}$ and $V^{0:T}_{s,j}$ simply as $V_{c,j}$ and $V_{s,j}$ respectively, and $V^{T:T+S}_i$ simply as $V_i$. Thus, we have: 
% \begin{equation}\label{equal_optimal_regressor}
%\resizebox{0.88\linewidth}{!}{$
  $f^*_{\rm IP}(V^{0:T}_{c,j}, V^{0:T}_{s,j}) \cong f^*_{\rm IP}(V_{c,j}, V_{s,j})  =  \mathbb{E}[V_i \vert V_{c,j}, V_{s,j}]$.
%$}
% \end{equation}
Collider structure implies independence relationship $ V_i \Vbar V_{s,j} $, we can obtain:
\begin{equation}\label{expectation=0}
%\resizebox{0.88\linewidth}{!}{$
    \begin{aligned}
    \mathbb{E}[f^*_{\rm IP}(V_{c,j}, V_{s,j}) \vert V_{s,j}] &= \mathbb{E}[\mathbb{E}[V_i \vert V_{c,j}, V_{s,j}] \vert V_{s,j}] 
     \\&= \mathbb{E}[V_i \vert V_{s,j}] = \mathbb{E}[V_i],
    \end{aligned}
%$}
\end{equation}

where the second equality follows from the tower property \citep{pearl2022graphoids}. Let $\mathcal{S}_{V_c}=\{V_{c,1},\cdots,V_{c,m}\}$ and $\mathcal{S}_{V_s}=\{V_{s,1},\cdots,V_{s,m}\}$. Since each path $ V_i \to V_{c,j} \leftarrow V_{s,j}$, $j=1,\cdots,m$, forms a separate collider structure and these structures do not intersect, the corresponding independence relations hold. $V_i \Vbar \mathcal{S}_{V_s}$ means every $V_{s,j} \in \mathcal{S}_{V_s}$ is $V_i \Vbar V_{s,j}$, and $V_i \nVbar \mathcal{S}_{V_s} \mid \mathcal{S}_{V_c}$ means condition on $\mathcal{S}_{V_c}$, $\forall V_{s,j} \in \mathcal{S}_{V_s}$ exists $V_i \nVbar V_{s,j} \mid \mathcal{S}_{V_c}$. For all variable pairs in $\{V_{c,j},V_{s,j}\}_{j=1}^m$, Equation (\ref{expectation=0}) equals to:
\begin{equation}\label{allexpectation=0}
%\resizebox{0.88\linewidth}{!}{$
    \begin{aligned}
    &\mathbb{E}[f^*_{\rm IP}(\mathcal{S}_{V_c}, \mathcal{S}_{V_s}) \vert \mathcal{S}_{V_s}] \\
    &= \mathbb{E}[f^*_{\rm IP}(V_{c,1}, V_{s,1},\cdots,V_{c,m}, V_{s,m}) \vert V_{s,1},\cdots,V_{s,m}] \\ 
    &= \mathbb{E}[\mathbb{E}[V_i \vert V_{c,1}, V_{s,1},\cdots,V_{c,m}, V_{s,m}] \vert V_{s,1},\cdots,V_{s,m}] \\
    &= \mathbb{E}[V_i \vert V_{s,1},\cdots,V_{s,m}] = \mathbb{E}[V_i],
    \end{aligned}
%$}
\end{equation}

Without loss of generality, we assume that $\mathbb{E}[V_i]=C$, here $C$ is a constant. This implies that:  
\begin{equation}\label{ZCE}
% \resizebox{0.88\linewidth}{!}{$
    f^*_{\rm IP} \in \mathcal{F}_{\Psi} = \left\{ f \in \mathcal{F} \, \big| \, \mathbb{E}[f(\mathcal{S}_{V_c}, \mathcal{S}_{V_s}) \mid \mathcal{S}_{V_s}] - C=0 \right\},
% $}
\end{equation}
where $ \mathcal{F}$ is denoted as $ L^2(V) $, a space of the square-integrable functions. Let $ \Phi: L^2(V) \to  L^2(V) $ denote the following conditional expectation operator:
\begin{equation}\label{operatorE}
%\resizebox{0.88\linewidth}{!}{$
    \Phi f(\mathcal{S}_{V_c}, \mathcal{S}_{V_s}) = \mathbb{E}[f(\mathcal{S}_{V_c}, \mathcal{S}_{V_s}) \mid \mathcal{S}_{V_s}]-C.
%$}
\end{equation}
Based on Equation (\ref{ZCE}) and Equation (\ref{operatorE}), the space $ L^2(V) $ can be decomposed orthogonally as
% \begin{equation}\label{directsum}
    $L^2(V) = \text{Preimage}(\Phi) \oplus \text{Kernel}(\Phi)$,
% \end{equation}
where $\text{Kernel}(\Phi) = \mathcal{F}_{\Psi}$ denotes the kernel (null space) of \( \Phi \), while $\text{Preimage}(\Phi)$ denotes the preimage space (inverse image) of \( \Phi \). Based on Equation (\ref{optimal_regressor}) and (\ref{ZCE}), we obtain that $f^*_{\rm IP}$ lies in $\text{Kernel}(\Phi)$. Then, for any $ f \in L^2(V) $, define the projection $\Psi$ as:
\begin{equation}\label{decompose}
    \Psi f = f - \Phi f,
\end{equation}
where $ \Psi = I - \Phi $ and $I$ is an identity mapping. Then, $\Psi$ orthogonal projects $f$ into $\text{Kernel}(\Phi)$. Thus, we want $\Psi f$ as our ideal prediction function. Then, we can obtain the following theorem, which demonstrates that $\Psi f$ can improve generalization of $f$:

\begin{theorem}\label{generalize}
    (Generalization Gap Reduction) For any predictor $ f \in L^2(V) $, we can obtain
    % \begin{equation}
        $\Delta(f, \Psi f) = \| \Phi f \|_{L^2(V)}^2 \geq 0$,
    % \end{equation}
    where $\Delta(f, \Psi f)$ denotes the generalization gap, which is defined by
    % \begin{equation} 
$\Delta (f,\Psi f) = \mathbb{E}[{(V_i - f(\mathcal{S}_{V_c},\mathcal{S}_{V_s}))^2}] - \mathbb{E}[{(V_i - \Psi f(\mathcal{S}_{V_c},\mathcal{S}_{V_s}))^2}]$.
    % \end{equation}
\end{theorem}
See the proof in Appendix \ref{proof}. From a causal perspective, $\Phi f$ captures the portion of $f$ that is spuriously correlated with $\mathcal{S}_{V_s}$. In other words, the operator $\Phi$ extracts those components of $f$ that vary systematically with $\mathcal{S}_{V_s}$ but offer no real predictive benefit for $V_i$. Under collider induced independence ($V_i \Vbar \mathcal{S}_{V_s}$), any apparent correlation with $\mathcal{S}_{V_s}$ reflects noise or sampling artifacts. Consequently, incorporating this dependence not only fails to improve prediction but can also inflate variance by fitting irrelevant fluctuations. By contrast, the projection $\Psi f$ removes these spurious elements-effectively filtering out parts of $f$ that do not help predict $V_i$. Eliminating such irrelevant dependencies tightens the generalization bound \citep{mohriFoundationsMachineLearning2018} because reducing the hypothesis space naturally curbs overfitting. This observation clarifies the reasoning behind \textbf{Theorem \ref{generalize}}, and shows how constraining $f$ to the kernel space of $\Phi$ directly mitigates the generalization gap. The above theoretical insights naturally motivate the decomposition strategy employed in our final predictive model.

\begin{figure*}
    \centering
    \includegraphics[width=0.8\linewidth]{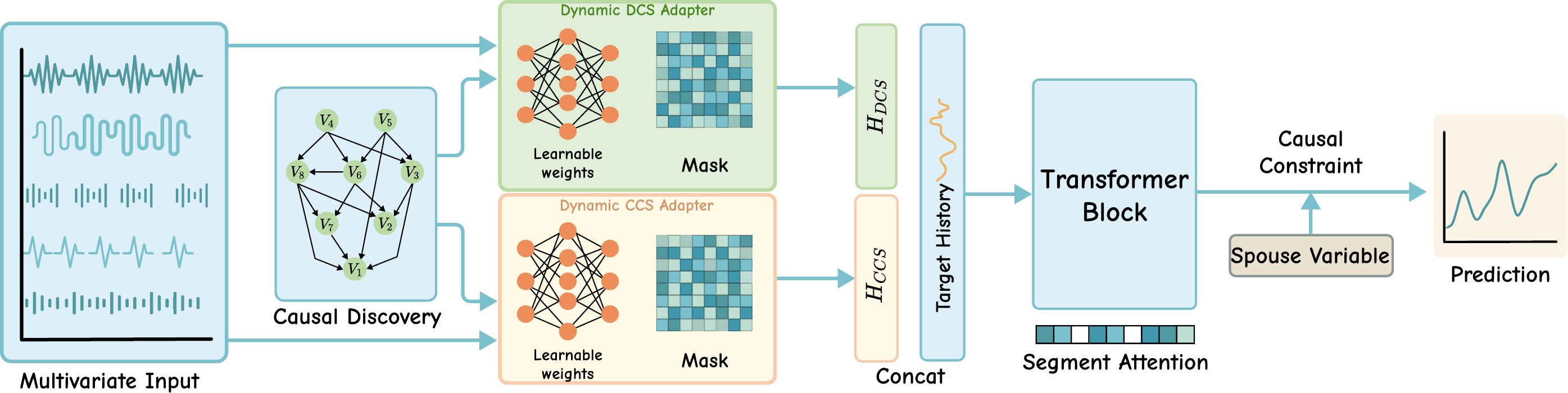}
    \caption{Architecture of the Causal Decomposition Transformer (CDT). The framework integrates three key stages: Structural Initialization, where a DAG prior guides the Dynamic Causal Adapter; Transformer Block, where a segment attention mechanism processes serialized causal contexts; and Spouse Projection, which enforces conditional independence constraints on the output.}
    \label{fig:framework}
    \vspace{-0.2 in}
\end{figure*}

\textbf{Motivation}. Drawing on the analyses in Sections \ref{Sec:Motivation.2}-\ref{Sec:Motivation.5}, we partition the historical influence of all variables on the target variable into three components: the ES, DCS, and CCS. This decomposition motivates an MTSF architecture that assigns a dedicated modeling module to each component and then combines their outputs to yield the final prediction.

\section{The Proposed Method} \label{Sec:Method}
In this section, we propose Causal Decomposition Transformer (CDT), which (i) employs a PC-learned DAG as a structural prior, (ii) refines it via a learnable dynamic causal adapter into target specific relevance weights, and (iii) enforces the SCM decomposition in Section~\ref{Sec:Motivation} through a segment constrained causal attention mask. Finally, motivated by Theorem~\ref{generalize}, CDT applies a spouse projection to suppress collider induced spurious dependence.

\paragraph{Causal Discovery}
We first apply the Peter-Clark (PC) algorithm~\citep{spirtes2001causation} on the observational training data to obtain an initial directed acyclic graph (DAG), represented by an adjacency matrix $A_{\text{init}}\in\{-1,0,1\}^{D\times D}$. Because PC relies on causal sufficiency and finite-sample conditional independence tests, we treat $A_{\text{init}}$ not as a ground truth but as a structural prior. It initializes learnable relevance logits that are refined end-to-end. Following Section~\ref{Sec:Motivation.4}, for each target variable $V_i$, we extract the initial sets of direct parents $\mathcal{S}^P_i$, direct children $\mathcal{S}^K_i$, and colliders $\mathcal{S}^C_i$. We denote the initial direct-causal set as $\mathcal{S}^D_i := \mathcal{S}^P_i\cup\mathcal{S}^K_i$.

\paragraph{Dynamic Causal Adapter}
Although the PC algorithm is theoretically grounded in the SCM framework and the d-separation criterion~\citep{pearlCausality2009} (Details in Appendix~B), in practice its output can be imperfect due to finite sample conditional independence tests and distributional shifts. To make the decomposition robust to such imperfections, we propose a Dynamic Causal Adapter (DCA) that treats the DAG as a structural prior rather than a fixed ground truth.

DCA replaces hard attention masks with target specific relevance weights derived from learnable logits. For each target variable $V_i$, DCA aggregates information from all variables into two causal-context representations: the direct causal subsegment $H_{\text{DCS}}^i$ and the collider causal subsegment $H_{\text{CCS}}^i$.

Concretely, we introduce two learnable logit matrices $W_{\text{DCS}}, W_{\text{CCS}}\in\mathbb{R}^{D\times D}$. These are initialized based on $A_{\text{init}}$: entries consistent with $\mathcal{S}^D_i$ (or $\mathcal{S}^C_i$) are initialized with larger logits $\alpha$, while unrelated entries receive finite negative values (e.g., $-\beta$ with $\beta>0$) to suppress them initially while allowing gradients to recover false negatives during training.
Let $\mathrm{Enc}(\cdot)$ be a Multi-Layer Perceptron (MLP) that maps each variable history $X_j$ to a sequence embedding $H_j=\mathrm{Enc}(X_j)\in\mathbb{R}^{T\times d}$. We apply two projection heads $g_d(\cdot)$ and $g_c(\cdot)$ to form channel-specific features. For target $V_i$, we compute soft relevance weights $\pi_{\text{DCS}}^{(i)}(j)=\sigma(W_{\text{DCS}}[j,i])$ and $\pi_{\text{CCS}}^{(i)}(j)=\sigma(W_{\text{CCS}}[j,i])$, where $\sigma$ is the sigmoid function, and aggregate:
\begin{equation}
\begin{aligned}
H_{\text{DCS}}^i &= \Sigma_{j=1}^{D} \pi_{\text{DCS}}^{(i)}(j)\cdot g_d(H_j),\\
H_{\text{CCS}}^i &= \Sigma_{j=1}^{D} \pi_{\text{CCS}}^{(i)}(j)\cdot g_c(H_j).
\end{aligned}
\end{equation}
Variables in the spurious segment $\mathcal{S}^S_i$ receive low prior logits and are further discouraged by prior regularization, so their contributions are typically negligible after training.

\begin{table*}[t]
  % \vspace{-5pt}
  \renewcommand{\arraystretch}{0.9} %行间距
  \centering
      \caption{MTSF results with prediction lengths $S\in\left \{96, 192, 336, 720\right \}$ and fixed lookback length $T = 96$. The best results in \textbf{bold}, the second \underline{underlined}, and “–” denote metrics not reported in the original papers. The lower MSE/MAE indicates a more accurate prediction result. The full results in Table \ref{tab:main_result}.}
 \vspace{-0.1in}
       \vspace{0.1in}
 \resizebox{\linewidth}{!}{
      \begin{tabular}{cc|cc|cc|cc|cc|cc|cc|cc|cc|cc|cc|cc|cc|cc|cc|cc|cc}
    \toprule
    \multicolumn{2}{c}{\multirow{1}{*}{Models}} & 
    \multicolumn{2}{c}{\rotatebox{0}{CDT}} &
    \multicolumn{2}{c}{\rotatebox{0}{Numerion}} &
    \multicolumn{2}{c}{\rotatebox{0}{DistDF}} &
    \multicolumn{2}{c}{\rotatebox{0}{GTR}} &
    \multicolumn{2}{c}{\rotatebox{0}{TimePro}} &
    \multicolumn{2}{c}{\rotatebox{0}{SEMPO}} &
    \multicolumn{2}{c}{\rotatebox{0}{TFPS}} &
    \multicolumn{2}{c}{\rotatebox{0}{{iTransformer}}} &
    \multicolumn{2}{c}{\rotatebox{0}{{PatchTST}}} &
    \multicolumn{2}{c}{\rotatebox{0}{{RLinear}}}  &
    \multicolumn{2}{c}{\rotatebox{0}{{Crossformer}}}  &
    \multicolumn{2}{c}{\rotatebox{0}{{TiDE}}} &
    \multicolumn{2}{c}{\rotatebox{0}{{{TimesNet}}}} &
    \multicolumn{2}{c}{\rotatebox{0}{{DLinear}}}&
    % \multicolumn{2}{c}{\rotatebox{0}{{SCINet}}} &
    \multicolumn{2}{c}{\rotatebox{0}{{FEDformer}}} &
    % \multicolumn{2}{c}{\rotatebox{0}{{Stationary}}} &
    \multicolumn{2}{c}{\rotatebox{0}{{Autoformer}}} \\
    \multicolumn{2}{c}{} &
    \multicolumn{2}{c}{{\textbf{(Ours)}}} &
    \multicolumn{2}{c}{{(\citeyear{Numerion}})} &
    \multicolumn{2}{c}{{(\citeyear{DistDF}})} &
    \multicolumn{2}{c}{{(\citeyear{GTR}})} &
    \multicolumn{2}{c}{{(\citeyear{timepro}})} &
    \multicolumn{2}{c}{{(\citeyear{SEMPO}})} &
    \multicolumn{2}{c}{{(\citeyear{TFPS}})} &
    \multicolumn{2}{c}{{(\citeyear{iTransformer})}} & 
    \multicolumn{2}{c}{{(\citeyear{PatchTST})}} & 
    \multicolumn{2}{c}{{(\citeyear{RLinear})}} &
    \multicolumn{2}{c}{{(\citeyear{Crossformer})}}  & 
    \multicolumn{2}{c}{{(\citeyear{TiDE})}} & 
    \multicolumn{2}{c}{{(\citeyear{TimesNet})}} & 
    \multicolumn{2}{c}{{(\citeyear{DLinear})}}& 
    % \multicolumn{2}{c}{{\citep{SCINet}}} &
    \multicolumn{2}{c}{{(\citeyear{FEDformer})}} &
    % \multicolumn{2}{c}{{\citep{NSTransformer}}} &
    \multicolumn{2}{c}{{(\citeyear{Autoformer})}} \\
    \cmidrule(lr){3-4} \cmidrule(lr){5-6}\cmidrule(lr){7-8} \cmidrule(lr){9-10}\cmidrule(lr){11-12}\cmidrule(lr){13-14} \cmidrule(lr){15-16} \cmidrule(lr){17-18} \cmidrule(lr){19-20}  \cmidrule(lr){21-22} \cmidrule(lr){23-24} \cmidrule(lr){25-26} \cmidrule(lr){27-28} \cmidrule(lr){29-30} \cmidrule(lr){31-32} \cmidrule(lr){33-34}
    \multicolumn{2}{c}{Metric}  & {MSE} & {MAE}  & MSE & MAE & MSE & MAE & MSE & MAE & {MSE} & {MAE}  & {MSE} & {MAE}  & {MSE} & {MAE}  & {MSE} & {MAE} & {MSE} & {MAE} & {MSE} & {MAE} & {MSE} & {MAE} & {MSE} & {MAE} & {MSE} & {MAE} & {MSE} & {MAE} & {MSE} & {MAE} & {MSE} & {MAE} \\
    \toprule

    \multicolumn{2}{c}{{\rotatebox{0}{{ETTm1}}}}
    &   \textbf{{0.365}} & \textbf{{0.375}} & \underline{0.370} & \underline{0.378} & {0.378} & {0.394} & {0.378} & {0.394} & 0.391 & {0.400} & 0.503 & {0.466} & 0.395 & {0.406} & {0.407} & {0.410} &{0.387} &\underline{0.400} & {0.414} & {0.407} & {0.513} & {0.496} & {0.419} & {0.419} &{{0.400}} &{{0.406}}  &{{0.403}} &{{0.407}} &{0.448} &{0.452}  &{0.588} &{0.517} \\ 
    \midrule
    
    \multicolumn{2}{c}{\rotatebox{0}{{ETTm2}}}
    & \textbf{{0.268}} & \underline{{0.316}} & \underline{0.270} & \textbf{0.315} & {0.277} & {0.321} & {0.277} & {0.321}  & {0.281} & {0.326}  & {0.286} & {0.341}  & {0.276} & {0.321}  & {{0.288}} & {{0.332}} &\underline{0.281} &{0.326} & {0.286} & {0.327} & {0.757} & {0.610} & {0.358} & {0.404} &{{0.291}} &{{0.333}} &{0.350} &{0.401} &{0.305} &{0.349} &{0.327} &{0.371} \\ 
    \midrule

    \multicolumn{2}{c}{\rotatebox{0}{{{ETTh1}}}}
    & \textbf{{0.406}} & \textbf{{0.415}} & {0.414} & \underline{0.417} & {0.430} & {0.429} & {0.430} & {0.429} & {0.438} & {0.438}  & \underline{0.410} & {0.430}  & {0.448} & {0.443}  & {{0.454}} & {{0.447}} &0.469 &0.454 & {0.446} & {0.434} & {0.529} & {0.522} & {0.541} & {0.507} &{0.458} &{{0.450}} &{{0.456}} &{{0.452}} &{{0.440}} &{0.460} &{0.496} &{0.487}  \\ 
    \midrule

    \multicolumn{2}{c}{\rotatebox{0}{{ETTh2}}}
    & \underline{{0.358}} & \underline{{0.386}}& {0.364} & {0.388} & {0.367} & {0.393} & {0.372} & {0.400} & {0.377} & {0.403}  & \textbf{0.341} & \textbf{0.391}  & {0.380} & {0.403}  & {{0.383}} & {{0.407}} &0.387 &0.407 & {{0.374}} & {{0.398}} & {0.942} & {0.684} & {0.611} & {0.550}  &{{0.414}} &{{0.427}} &{0.559} &{0.515} &{{0.437}} &{{0.449}} &{0.450} &{0.459} \\  
    \midrule

    \multicolumn{2}{c}{\rotatebox{0}{{Weather}}} 
    & \textbf{{0.239}} & \textbf{{0.267}} & {0.246} & \underline{0.271} & {0.248} & {0.275} & {0.248} & {0.275} & {0.251} & {0.276}  & {0.248} & {0.287}  & \underline{0.241} & \underline{0.271}  & {{0.258}} & {{0.279}} &0.259 &0.281 &0.272 &0.291 & {0.259} & {0.315} & {0.271} & {0.320} &{{0.259}} &{{0.287}} &{0.265} &{0.317} &{0.309} &{0.360} &{0.338} &{0.382} \\
    \midrule
    \multicolumn{2}{c}{\rotatebox{0}{{ECL}}} 
    & \textbf{{0.165}} & \textbf{{0.258}} & {0.181} & {0.267} & {0.172} & {0.267} & \underline{0.166} & \underline{0.260} & {0.169} & {0.262}  & {0.196} & {0.295}  & {0.183} & {0.280}  & {{0.178}} & {{0.270}} & {0.205} & {{0.290}} & {0.219} & {0.298} & {0.244} & {0.334} & {0.251} & {0.344} &{{0.192}} &{0.295} &{0.212} &{0.300} &{0.214} &{0.327} &{0.227} &{0.338} \\  
    \midrule

    \multicolumn{2}{c}{\rotatebox{0}{{Exchange}}}
    & \textbf{{0.344}} & \textbf{{0.395}}& {0.358} & \underline{0.399} & {-} & {-} & {-} & {-} & \underline{0.352} & \underline{0.399} & - & -  & {0.395} & {0.414}  & {0.360} & {{0.403}} &0.367 &0.404 & {0.378} & {0.417} & {0.940} & {0.707} & {0.370} & {0.413} & {0.416} & {0.443} & {{0.354}} & {0.414} & {0.519} & {0.429} & {0.613} & {0.539} \\ 
    \midrule
    
    \multicolumn{2}{c}{\rotatebox{0}{{Traffic}}} 
    & \textbf{{0.411}} & \textbf{{0.274}} & {0.468} & {0.281} & \underline{0.417} & \underline{0.279} & {0.470} & {0.280} & {-} & {-}  & {0.466} & {0.344}  & - & -  & {{0.428}} & {{0.282}} & {{0.481}} & {{0.304}} & {0.626} & {0.378}& {0.550} & {{0.304}} & {0.760} & {0.473} &{{0.620}} &{{0.336}} &{0.625} &{0.383} &{{0.610}} &{0.376} &{0.628} &{0.379} \\   

    \bottomrule
  \end{tabular}}
  \label{tab:main_result_avg}
\end{table*}
 \vspace{-0.1in}

\paragraph{Segment Constrained Attention}
For each target $V_i$, we serialize the SCM decomposition by concatenating the three segments $S_i = [H_{\text{DCS}}^i \,;\, H_{\text{CCS}}^i \,;\, H_{\text{ES}}^i]$, where $H_{\text{ES}}^i$ denotes the endogenous representation of the target history. To strictly enforce the causal roles defined in Section~\ref{Sec:Motivation}, CDT applies a Transformer with a segment constrained causal mask $M$. Specifically, for any time step $t$, the attention mechanism enforces the following visibility rules: (i) DCS Independence: A token at time $t$ in the DCS subsegment can only attend to DCS tokens at time steps $\tau \le t$; (ii) CCS Independence: A token at time $t$ in the CCS subsegment can only attend to CCS tokens at time steps $\tau \le t$; (iii) Endogenous Integration: A token at time $t$ in the ES segment (the reasoning anchor) attends to tokens from all three subsegments at time steps $\tau \le t$. This design ensures that cross-segment information flows exclusively into the endogenous stream for prediction, preventing spurious information leakage between the causal contexts. Our framework is model agnostic and can be applied to frozen LLM backbones; see Appendix \ref{tab:llm_ablation}.

\paragraph{Causal Constrained Output Projection}
Let $\hat{Y}_{\text{raw}}^i\in\mathbb{R}^{S}$ be the raw prediction for target $V_i$. Motivated by Theorem~\ref{generalize}, we project the prediction onto the kernel space of the collider spouse dependency. 
Analogous to the DCA, we introduce a learnable logit matrix $W_{\text{SP}}\in\mathbb{R}^{D\times D}$ to derive spouse weights $\pi_{\text{SP}}^{(i)}(j)=\sigma(W_{\text{SP}}[j,i])$. The logits are initialized from the PC prior (marking spouse candidates $\mathcal{S}^S_i$) but remain refineable. The spouse context is aggregated as $H_{\text{SP}}^i=\sum_{j=1}^{D}\pi_{\text{SP}}^{(i)}(j)\cdot g_s(H_j)$. We then subtract the estimated conditional expectation $\hat{Y}^i=\hat{Y}_{\text{raw}}^i-\Phi(\hat{Y}_{\text{raw}}^i, H_{\text{SP}}^i)$, where $\Phi(\cdot)$ approximates $\mathbb{E}[\hat{Y}_{\text{raw}}^i\mid H_{\text{SP}}^i]$.

\paragraph{Training Objective}
The total loss function $\mathcal{L}$ combines the forecasting error with regularization terms for the three learnable causal structures (DCS, CCS, and Spouse) $\mathcal{L} = \mathcal{L}_{\text{MSE}} + \lambda \mathcal{L}_{\text{Reg}}$.
Here, $\mathcal{L}_{\text{MSE}}$ is the Mean Squared Error of the forecast and $\mathcal{L}_{\text{Reg}} = \mathcal{L}_{\text{prior}}(W_{\text{DCS}}, A_{\text{DCS}}) + \mathcal{L}_{\text{prior}}(W_{\text{CCS}}, A_{\text{CCS}}) + \mathcal{L}_{\text{prior}}(W_{\text{SP}}, A_{\text{SP}})$. The prior losses $\mathcal{L}_{\text{prior}}(W, A)$ enforce both sparsity and alignment with the PC initialized DAG. Specifically, we formulate this as a binary cross-entropy loss between the learned probabilities $\sigma(W)$ and the binary masks derived from the PC prior DAG, guiding the optimization to respect the initial causal discovery while allowing for gradient-based refinement.

\section{Experimental Results}\label{Experiment}
In this section, we conduct extensive experiments on various benchmarks to evaluate the effectiveness of CDT. Due to space limitations, detailed discussions on the theoretical alignment with Pearl's framework, structural visualizations, and scalability to LLMs are presented in the Appendix. These supplementary results further validate the soundness and versatility of the CDT framework.

\paragraph{Datasets} We conduct long-term forecasting experiments on eight real-world datasets, including ECL, the ETT series (h1, h2, m1, m2), Traffic, Exchange and Weather, which is widely used in recent MTSF studies \cite{AutoTimes}. Details in Appendix \ref{appendixdatasets}

\paragraph{Baselines} We compare against fifteen representative baselines, including Transformer-based (DistDF \cite{DistDF}, iTransformer \cite{iTransformer}, PatchTST \cite{PatchTST}, Autoformer \cite{Autoformer}, FEDformer \cite{FEDformer}, Crossformer \cite{Crossformer}, TimePro \cite{timepro}, SEMPO \cite{SEMPO}, TFPS \cite{TFPS}), linear (Numerion \cite{Numerion}, GTR \cite{GTR}, DLinear \cite{DLinear}, TiDE \cite{TiDE}, RLinear \cite{RLinear}), and TCN-based (TimesNet \cite{TimesNet}) models.

\begin{table}[t]
\renewcommand{\arraystretch}{0.55} % 稍微增加一点行间距，提升可读性
\setlength{\tabcolsep}{4pt} % 调整列间距
\centering
\caption{Ablation study of CDT components on ETTh1, ETTm1, Weather, and ECL. All results are averaged over five seeds, with input length $T{=}96$ and prediction horizons $S\in\{96,192,336,720\}$. Lower is better. \textbf{Bold} indicates the best performance.}
\label{tab:cdt_ablation_components}
\resizebox{0.95\linewidth}{!}{
\begin{tabular}{lcccccccc}
\toprule
\multirow{2}{*}{\textbf{Model Setting}} & \multicolumn{2}{c}{\textbf{ETTh1}} & \multicolumn{2}{c}{\textbf{ETTm1}} & \multicolumn{2}{c}{\textbf{Weather}} & \multicolumn{2}{c}{\textbf{ECL}} \\
\cmidrule(lr){2-3}\cmidrule(lr){4-5}\cmidrule(lr){6-7}\cmidrule(lr){8-9}
& MSE & MAE & MSE & MAE & MSE & MAE & MSE & MAE \\
\midrule
\textbf{CDT} & \textbf{0.406} & \textbf{0.415} & \textbf{0.365} & \textbf{0.375} & \textbf{0.239} & \textbf{0.267} & \textbf{0.168} & \textbf{0.261} \\
\midrule
% \multicolumn{9}{l}{\textit{Effect of Structural Prior \& Adapter}} \\
Non-learnable Prior (Static) & 0.418 & 0.424 & 0.374 & 0.382 & 0.246 & 0.275 & 0.175 & 0.268 \\
Random Init (No Prior) & 0.425 & 0.431 & 0.381 & 0.389 & 0.252 & 0.281 & 0.180 & 0.274 \\
\midrule
% \multicolumn{9}{l}{\textit{Effect of SCM Decomposition}} \\
w/o DCS (Direct Causal) & 0.442 & 0.448 & 0.403 & 0.410 & 0.271 & 0.298 & 0.194 & 0.285 \\
w/o CCS (Collider Causal) & 0.415 & 0.423 & 0.378 & 0.385 & 0.248 & 0.276 & 0.176 & 0.269 \\
w/o Segment Constraints & 0.421 & 0.428 & 0.385 & 0.392 & 0.255 & 0.283 & 0.182 & 0.275 \\
\bottomrule
\end{tabular}}
\end{table}

\paragraph{Main Results}
We evaluate the proposed method on standard MTSF benchmark datasets. Specifically, we compare our method with other approaches under the long-term forecasting standard setting, which follows \cite{iTransformer}. The results are reported in Table \ref{tab:main_result_avg}. The results show that our method outperforms current popular MTSF approaches on most datasets and under most settings.

\begin{figure*}[htpb]
    \centering
    \includegraphics[width=0.8\linewidth]{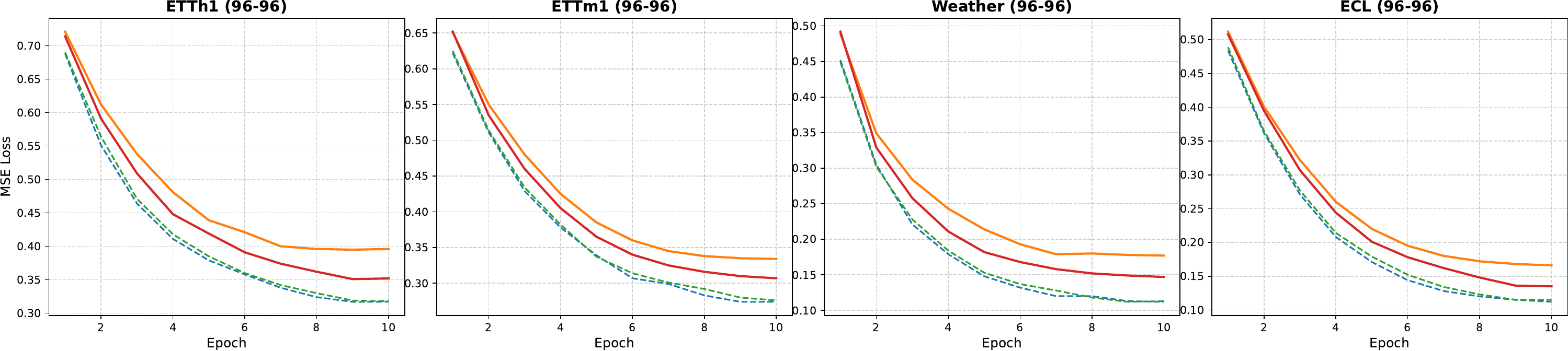}
    \caption{Comparison of training (dashed) and testing (solid) MSE loss trajectories on four datasets. The model w/ Projection (Blue) exhibits a significantly smaller generalization gap compared to the variant w/o Projection (Orange/Red), validating that the projection operator effectively filters spurious collider correlations that lead to overfitting.}
    \label{fig:spouse_gap}
\end{figure*}

\subsection{Ablation Study}

\paragraph{Ablation Study on Module} 
We verify CDT variants on ETTh1, ETTm1, Weather, and ECL. We fix $T=96$ and report averaged MSE/MAE over $S\in\{96,192,336,720\}$. We compare: (i) Full CDT; (ii) Static Prior: freezing relevance logits after PC initialization without refinement; (iii) Random Init: learning logits from scratch without causal priors; (iv/v) w/o DCS/CCS: removing direct causal or collider causal subsegments; and (vi) Global Attention: replacing the segment constrained mask with a standard causal mask. Results in Table~\ref{tab:cdt_ablation_components} demonstrate that Full CDT consistently outperforms all variants. Specifically, Static Prior underperforms Full CDT, and Random Init degrades further, demonstrating both the necessity of DCA based error correction and the value of structural priors for cold start. Removing DCS causes the largest error increase, while removing CCS or segment constraints also leads to clear deterioration, confirming the importance of direct parents, collider contexts, and preventing spurious cross-segment mixing. Although Appendix~\ref{app:pc_semantics} establishes the graphical semantics of PC/CPDAG, CDT treats the discovered structure only as a warm-start prior; its practical impact is therefore assessed via ablations (Table~\ref{tab:cdt_ablation_components}) and perturbation tests (Fig.~\ref{fig:perturb}).

 \vspace{-0.2in}

\paragraph{Ablation Study on spouse projection}
We conduct an ablation study on ETTh1, ETTm1, Weather, and ECL ($T=S=96$). We compare the full CDT against a variant without the projection module. As shown in Figure~\ref{fig:spouse_gap}, enabling spouse projection consistently narrows the generalization gap ($\Delta \mathcal{L} = \mathcal{L}_{\text{test}}-\mathcal{L}_{\text{train}}$) and yields lower test loss, while leaving training loss largely unchanged. This result provides empirical support for Theorem~\ref{generalize}: by suppressing collider induced spurious dependencies, which are fit during training but fail to generalize, the projection operator explicitly improves out-of-distribution robustness.

\subsection{Robustness Analysis}
\begin{figure}
    \centering
    \includegraphics[width=0.9\linewidth]{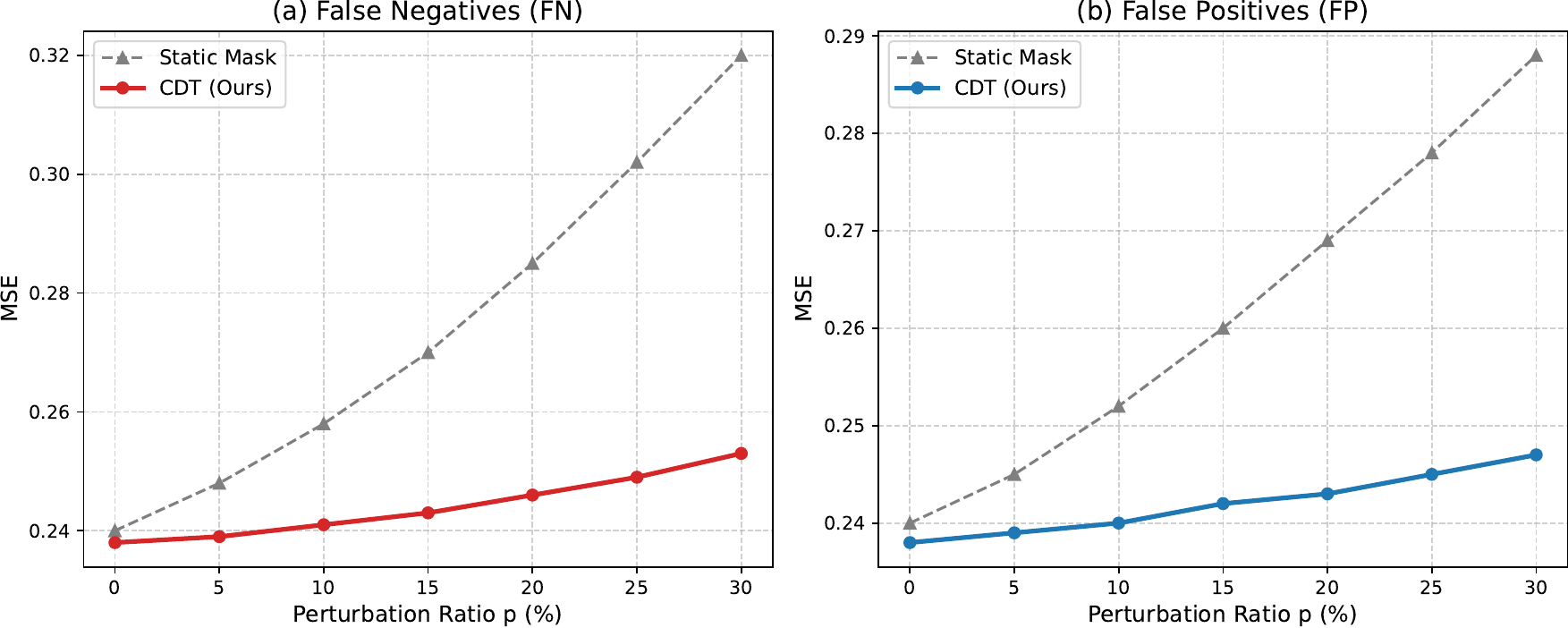}
    \caption{MSE/MAE vs. Perturbation Percentage on Weather.}
    \label{fig:perturb}
\end{figure}

We evaluate performance with structurally perturbed PC-learned DAGs on the Weather dataset (input/predict$=96$). We inject synthetic noise with perturbation ratios $p \in {0\%, 10\%, 20\%, 30\%}$, considering two cases: (i) False Negatives (FN), where $p\%$ true edges are removed, and (ii) False Positives (FP), where $p\%$ spurious edges are added. We compare CDT (ours) with a Static Prior variant that fixes adjacency weights to the initial prior $A_{\text{init}}$. As shown in Fig.~\ref{fig:perturb}, the Static Prior degrades rapidly as noise increases, especially under FN perturbations (up to $\sim$18\% MSE increase at $p=30\%$), highlighting the brittleness of rigid causal constraints. In contrast, CDT remains highly robust, exhibiting a much flatter degradation trend and maintaining competitive accuracy even under severe FN noise. This demonstrates that DCA can recover missing causal relations through end-to-end gradient refinement, effectively using the noisy DAG as a warm-start rather than a hard constraint.

\begin{table}[t]
\centering
\caption{Comparison of training/inference time and performance.}
\label{tab:efficiency}
\resizebox{0.75\linewidth}{!}{
\begin{tabular}{lccc}
\toprule
Model & Train Time (s) & Test Time (s) & MSE \\
\midrule
CDT & 242 & 3.2 & 0.147 \\
PatchTST & 253 & 3.3 & 0.177 \\
iTransformer & 150 & 2.2 & 0.174 \\
\bottomrule
\end{tabular}}
%\vspace{-10pt}
\end{table}

\subsection{Efficiency Analysis}
We evaluate computational costs on the Weather dataset (input-96/predict-96). As shown in Table~\ref{tab:efficiency}, CDT incurs a moderate overhead compared to baselines due to the additional causal adapter computations. However, thanks to the parallelized implementation, the inference time remains efficient (4.2s). Crucially, this marginal cost yields a significant performance boost (MSE 0.147 vs. 0.177), verifying that the trade-off between complexity and accuracy is favorable. The causal discovery step is performed once offline; its stand-alone runtime is reported in Appendix~\ref{app:cd_runtime}.

 % \vspace{-0.1in}

\section{Conclusion}
This paper presents a novel perspective on MTSF by decomposing historical data into endogenous, direct causal, collider causal, and spurious segments based on SCMs. To operationalize this, we develop the Causal Decomposition Transformer (CDT). Our method makes three key contributions: It introduces an All-to-One strategy that focuses on target specific causal contexts; It implements a Dynamic Causal Adapter that initializes from a DAG, allowing the model to dynamically refine causal structures and correct initial discovery errors; and it applies a Causal Constrained Output Projection to mitigate spurious dependencies arising from collider structures. Empirical results across multiple benchmarks confirm that CDT consistently outperforms strong baselines. Moreover, ablation and robustness studies validate that our dynamic refinement strategy effectively handles noisy causal priors, offering a robust and interpretable solution for MTSF.

\section*{Impact Statements}
This paper presents work whose goal is to advance the field of machine learning and time series forecasting. There are many potential societal consequences of our work, none of which we feel must be specifically highlighted here.

\bibliography{reference}
\bibliographystyle{icml2026}

%%%%%%%%%%%%%%%%%%%%%%%%%%%%%%%%%%%%%%%%%%%%%%%%%%%%%%%%%%%%%%%%%%%%%%%%%%%%%%%
%%%%%%%%%%%%%%%%%%%%%%%%%%%%%%%%%%%%%%%%%%%%%%%%%%%%%%%%%%%%%%%%%%%%%%%%%%%%%%%
% APPENDIX
%%%%%%%%%%%%%%%%%%%%%%%%%%%%%%%%%%%%%%%%%%%%%%%%%%%%%%%%%%%%%%%%%%%%%%%%%%%%%%%
%%%%%%%%%%%%%%%%%%%%%%%%%%%%%%%%%%%%%%%%%%%%%%%%%%%%%%%%%%%%%%%%%%%%%%%%%%%%%%%
\newpage
\appendix
\onecolumn

\section*{Appendix}
The appendix is organized into several sections:
\begin{itemize}
    \item Appendix \ref{notation} provides the notation;
    \item Appendix \ref{background_causal} provides the background in causality;
    \item Appendix \ref{app:pc_semantics} provides the causal semantics;
    \item Appendix \ref{appproof} presents the proof of analysis in Section \ref{Sec:Motivation.4} and Theorem \ref{generalize};
    \item Appendix \ref{appimple} provides the details of implements;
    \item Appendix \ref{appendixdatasets} provider the details of datasets;
    \item Appendix \ref{app_exp} provides more experiment;
    \item Appendix \ref{pc} details the implementation of the causal discovery algorithm PC;
    \item Appendix \ref{compare} prevents the comparison of the results among CDT, iTransformer, and PatchTST.
\end{itemize}

\section{List of Notations}\label{notation}
We list the definitions of all notations used throughout the paper as follows:
\begin{itemize}[label=$\square$]
\item \textbf{Time Series and Forecasting Symbols (Basic Class)}
\begin{itemize}[label=$\bullet$]
    \item $X = [x_1,\dots,x_T]\in\mathbb{R}^{T\times D}$: multivariate historical time series with $T$ look-back steps and $D$ variables.
    \item $Y = [x_{T+1},\dots,x_{T+S}]\in\mathbb{R}^{S\times D}$: multivariate future sequence with prediction horizon $S$.
    \item $x_t = [V_1^t,\dots,V_D^t]\in\mathbb{R}^D$: observations of all variables at time step $t$.
    \item $V_i^t\in\mathbb{R}$: value of the $i$-th variable at time $t$.
    \item $V_i^{t:t+\tau-1}\in\mathbb{R}^{\tau}$: length-$\tau$ subsequence of variable $V_i$.
    \item $X_i\in\mathbb{R}^{T}$: historical sequence of the target variable $V_i$.
    \item $Y_i\in\mathbb{R}^{S}$: future sequence of the target variable $V_i$.
    \item $T$: look-back window length.
    \item $S$: prediction horizon.
    \item $\mathcal{D}_{\text{train}}=\{(X^{(n)},Y^{(n)})\}_{n=1}^{N}$: training dataset.
\end{itemize}

\item \textbf{Causal Graph and Variable Categorization (Causal Class)}
\begin{itemize}[label=$\bullet$]
    \item $G=(V,E)$: causal directed acyclic graph (DAG).
    \item $V=\{V_1,V_2,\dots,V_D\}$: set of variable nodes.
    \item $A_{\text{init}}\in\{-1,0,1\}^{D\times D}$: adjacency matrix learned by the PC algorithm, serving as a structural prior.
    \item $V_i$: target variable.
    \item $\mathcal{S}_i^P$: set of direct parents of $V_i$.
    \item $\mathcal{S}_i^K$: set of direct children of $V_i$ that are not colliders.
    \item $\mathcal{S}_i^C$: set of collider variables associated with $V_i$.
    \item $\mathcal{S}_i^D := \mathcal{S}_i^P \cup \mathcal{S}_i^K$: direct causal set of $V_i$.
    \item $\mathcal{S}_i^{\text{SP}}$: spouse set of $V_i$ induced by collider structures.
    \item $\mathcal{S}_i^{\text{Spu}}$: spurious (non-causal) variables of $V_i$.
\end{itemize}

\item \textbf{Model Architecture and Dynamic Causal Adapter (Model Class)}
\begin{itemize}[label=$\bullet$]
    \item $\mathrm{Enc}(\cdot)$: variable-wise encoder mapping raw histories to latent representations.
    \item $H_j=\mathrm{Enc}(X_j)\in\mathbb{R}^{T\times d}$: encoded representation of the $j$-th variable.
    \item $g_d(\cdot), g_c(\cdot), g_s(\cdot)$: projection heads for direct-causal, collider-causal, and spouse channels, respectively.
    \item $W_{\text{DCS}}, W_{\text{CCS}}, W_{\text{SP}}\in\mathbb{R}^{D\times D}$: learnable relevance logit matrices initialized from $A_{\text{init}}$.
    \item $\pi_{\text{DCS}}^{(i)}(j)=\sigma(W_{\text{DCS}}[j,i])$: soft relevance weight of variable $j$ to target $V_i$ in the direct-causal channel.
    \item $\pi_{\text{CCS}}^{(i)}(j)=\sigma(W_{\text{CCS}}[j,i])$: soft relevance weight in the collider-causal channel.
    \item $\pi_{\text{SP}}^{(i)}(j)=\sigma(W_{\text{SP}}[j,i])$: soft relevance weight in the spouse channel.
    \item $H_{\text{DCS}}^i = \sum_{j=1}^{D}\pi_{\text{DCS}}^{(i)}(j)\cdot g_d(H_j)$: aggregated direct-causal representation for $V_i$.
    \item $H_{\text{CCS}}^i = \sum_{j=1}^{D}\pi_{\text{CCS}}^{(i)}(j)\cdot g_c(H_j)$: aggregated collider-causal representation for $V_i$.
    \item $H_{\text{ES}}^i$: endogenous representation of the target history.
    \item $H_{\text{SP}}^i = \sum_{j=1}^{D}\pi_{\text{SP}}^{(i)}(j)\cdot g_s(H_j)$: aggregated spouse representation for $V_i$.
    \item $S_i=[H_{\text{DCS}}^i;H_{\text{CCS}}^i;H_{\text{ES}}^i]$: serialized causal token stream for target $V_i$.
    \item $M$: segment constrained causal attention mask enforcing SCM-aligned information flow.
    \item $\hat{Y}_{\text{raw}}^i\in\mathbb{R}^{S}$: raw prediction for target $V_i$.
    \item $\hat{Y}^i\in\mathbb{R}^{S}$: final prediction after causal projection.
\end{itemize}

\item \textbf{Theoretical Analysis and Projection Operators (Theory Class)}
\begin{itemize}[label=$\bullet$]
    \item $\mathbb{E}[\cdot]$: expectation operator.
    \item $\Phi(\hat{Y}_{\text{raw}}^i \mid H_{\text{SP}}^i)$: conditional expectation operator approximated by a learnable projection head.
    \item $\Psi(\hat{Y}_{\text{raw}}^i; H_{\text{SP}}^i)
    = \hat{Y}_{\text{raw}}^i - \Phi(\hat{Y}_{\text{raw}}^i \mid H_{\text{SP}}^i)$: causal projection operator that suppresses collider induced spurious dependence.
    \item $\mathcal{F}_\Psi$: function class satisfying collider-consistency constraints.
    \item $L^2(V)$: space of square-integrable functions over variable space $V$.
    \item $\Delta(f,\Psi f)$: expected risk reduction induced by causal projection.
\end{itemize}

\end{itemize}

%这一小节，挪到appendix。
\section{Background In Causality} \label{background_causal}
Causal relationships among variables play a crucial role in MTSF. By constructing a structural causal model, we can better understand the dependencies and independencies among variables, enabling us to build more accurate forecasting models \citep{pearlCausality2009}. 
% 因果论的一个核心概念是条件独立关系, 其定义如下所示
One core concept of causality is conditional independence, which is defined as:
\begin{definition}[Conditional Independence \citep{dawid1979conditional}]\label{condin}
    Let $V=\left\{V_1,V_2,\cdots    ,V_D\right\}$ be a finite set of variables, $V_i$ is the $i$-th variable and $D$ is the number of variable, $P(\cdot)$ be a joint probability function over the variables in $V$, and $\mathcal{S}_X$, $\mathcal{S}_Y$, $\mathcal{S}_Z$ stand for three subsets of variables in $V$. Then, $\mathcal{S}_X$ and $\mathcal{S}_Y$ are said to be conditionally independent given $\mathcal{S}_Z$ if    \begin{equation}P(\mathcal{S}_X\mid\mathcal{S}_Y,\mathcal{S}_Z)=P(\mathcal{S}_X\mid\mathcal{S}_Z), \forall P(\mathcal{S}_Y,\mathcal{S}_Z)>0.
    \end{equation}
    That is, $\mathcal{S}_Y$ does not provide any additional information for predicting $\mathcal{S}_X$, once given $\mathcal{S}_Z$. $\mathcal{S}_X\Vbar \mathcal{S}_Y \mid \mathcal{S}_Z$ denotes the conditional independence of $\mathcal{S}_X$ and $\mathcal{S}_Y$ given $\mathcal{S}_Z$.
\end{definition}

% 变量间的条件独立关系是因果图模型的基础, 在因果图模型中，我们往往使用有向无环图$G=(V,E)$来描述变量之间的关系, 其中节点集合$V=\{V_1,V_2,\dots\}$代表随机变量, 边的集合$E$代表随机变量间的因果关系. 在因果图模型中， 
Conditional independence relationships among variables form the basis of the SCM. In these models, a DAG, denoted as $ G = (V, E) $, is typically used to represent the relationships among variables, where the node set $ V = \{V_1,V_2,\cdots,V_D\} $ corresponds to random variables, and the edge set $ E = \{(V_1, V_2), (V_2, V_3), \cdots\} $ represents causal relationships between variables. An SCM is built upon three fundamental structures: Chain, Fork and Collider. Any model containing at least three variables incorporates these key structures.

\begin{definition}[\textbf{Chain}]
A chain $ V_p \to  V_i \to  V_c $ is a graphical structure involving three variables $ V_p $, $ V_i $, and $ V_c $ in graph $ G $, where $ V_p $ has a directed edge to $ V_i $ and $ V_i $ has a directed edge to $ V_c $. Here, $ V_p $ causally influences $ V_i $, and $ V_i $ causally influences $ V_c $, making $ V_i $ a mediator.
\end{definition}

\begin{definition}[\textbf{Fork}]
A fork $ V_b \gets  V_p \to  V_i $ is a graphical structure involving $ V_b $, $ V_p $, and $ V_i $, where $ V_p $ is a common parent of both $ V_b $ and $ V_i $. $ V_p $ causally influences $ V_b $ and $ V_i $.
\end{definition}

\begin{definition}[\textbf{Collider}]
A collider, also known as a V-structure, $ V_i \to  V_c \gets  V_s $, is a graphical structure involving three variables $ V_i $, $ V_c $, and $ V_s $, where $ V_c $ is a common child of both $ V_i $ and $ V_s $, $ V_i $ and $ V_s $ are not directly connected. Here, $ V_i $ and $ V_s $ causally influence $V_c$.
% In this setup, $ V_i $ and $ V_s $ are conditionally independent given $ V_c $, but they are not independent without conditioning on $ V_c $.
\end{definition}

In a chain structure, $ V_p $ and $ V_c $ are conditionally independent given $ V_i $, formally, $ V_p \Vbar V_c \mid V_i $. In a fork structure, $ V_b $ and $ V_i $ are independent given $ V_p $, $ V_b $ provides no additional information about $ V_i $, and vice versa, i.e., $ V_b \Vbar V_i \mid V_p$. In a collider structure, $ V_i $ and $ V_s $ are marginally independent, knowing $ V_i $ does not provide information about $ V_s $ and vice versa. However, when conditioning on the collider $ V_c $, this independence is broken, making $ V_i $ and $ V_s $ dependent. Formally, $ V_i \Vbar V_s $ and $ V_i \nVbar V_s \mid V_c $. The related proofs are presented in Chapter Two of \citep{pearl2009causal}. The above independence relationships are fundamental for understanding the dependencies implied by an SCM, thereby facilitating tasks such as causal discovery and causal inference in MTSF.

\section{Graphical Causal Semantics and the Role of PC as a Structural Prior}\label{app:pc_semantics}
This appendix clarifies the causal semantics underlying our use of the Peter-Clark (PC) algorithm, and explains why we treat the discovered structure as a {structural prior} rather than a ground-truth causal graph. Our discussion follows the standard graphical framework in Pearl~\citep{pearl2009causal}, within which constraint-based discovery algorithms such as PC~\citep{spirtes2001causation} are theoretically grounded.

\subsection{Graphical prerequisites: Markov compatibility, d-separation, and equivalence}
Let $\mathcal{G}$ be a directed acyclic graph (DAG) over variables $\mathbf{V}=\{V_1,\ldots,V_D\}$ and let $P(\mathbf{V})$ denote their joint observational distribution.
The graphical framework relies on the following standard notions (indices refer to Pearl~\citep{pearl2009causal}):

\paragraph{Markov compatibility}
A distribution $P(\mathbf{V})$ is {compatible} with a DAG $\mathcal{G}$ if it factorizes according to $\mathcal{G}$ (Def.~1.2.2), i.e.,
\begin{equation}
P(\mathbf{V}) = \prod_{i=1}^D P\!\left(V_i \mid \mathrm{Pa}_{\mathcal{G}}(V_i)\right),
\end{equation}
where $\mathrm{Pa}_{\mathcal{G}}(V_i)$ denotes the parents of $V_i$ in $\mathcal{G}$.

\paragraph{d-separation and observational equivalence}
d-separation in $\mathcal{G}$ is sound and complete for the conditional independencies in $P(\mathbf{V})$ (Def.~1.2.3). Moreover, two DAGs are observationally equivalent if they share the same skeleton and the same set of v-structures (Thm.~1.2.8). Consequently, purely observational data can identify a causal structure only {up to} a Markov equivalence class.

\subsection{Minimality and faithfulness: what it means to be identifiable from observations}
\label{app:pc_semantics:identifiable}

Within Pearl's framework, the causal content that is identifiable from observational data is characterized by additional regularity conditions:

\paragraph{Minimality}
A causal graph is {minimal} if no edge can be removed without contradicting the conditional independencies encoded by $P(\mathbf{V})$ (Sec.~2.3, ``minimal potential structure''). Intuitively, minimality rules out redundant edges that do not correspond to necessary dependencies.

\paragraph{Stability/Faithfulness.}
Only those independencies that persist under small parameter perturbations are treated as structural (Sec.~2.4). ``Accidental'' independencies caused by fine-tuned parameters are excluded. This assumption links the observed independence pattern to a sparse graphical explanation.

\paragraph{Conservative notion of causal effect.}
Pearl defines causal claims conservatively: $X$ is said to have a causal effect on $Y$ only if {every} minimal structure consistent with the observed distribution contains a directed path from $X$ to $Y$ (Def.~2.3.6). This emphasizes that causal conclusions from observational data should be understood as statements that are robust across the entire set of indistinguishable minimal DAGs.

\subsection{What PC returns: a CPDAG representing a Markov equivalence class}
The PC algorithm is the canonical constraint-based realization of the above notions.
It uses conditional independence tests to (i) remove edges until a minimal graph consistent with the estimated independencies is obtained, and (ii) orient as many edges as possible without violating the equivalence constraints implied by v-structures and acyclicity.
The output is a completed partially directed acyclic graph (CPDAG), which represents the entire Markov equivalence class of DAGs consistent with the recovered independence model.

From the standpoint of Pearl's theory, this CPDAG is precisely the maximal causal information that can be justified from observational data under Markov compatibility and faithfulness. No method can legitimately claim identification of a {unique} DAG without additional assumptions (e.g., interventions, experimental designs, or stronger functional restrictions).

\subsection{Why we treat the discovered graph as a structural prior}
Our method uses the PC-discovered structure as a structural prior, not as ground truth. This design choice reflects two principles:

\paragraph{Identifiable content is class-level, not edge-level}
Under Thm.~1.2.8, all DAGs in a Markov equivalence class share the same skeleton and v-structures and thus encode exactly the same set of conditional independencies.
Therefore, observationally identifiable causal semantics are determined primarily by the induced independence model, rather than by any particular orientation decision within the class.
Edge-level perturbations that do not alter the skeleton/v-structures do not correspond to a new observationally justified causal hypothesis; only changes that alter the underlying independencies represent genuinely different hypotheses.

\paragraph{Finite-sample noise affects tests, not the causal semantics}
Order-dependence and variability in PC arise from finite-sample estimation of conditional independencies (a statistical issue), not from the underlying graphical semantics.
Accordingly, we do not interpret $A_{\text{init}}$ as a definitive causal graph.
Instead, it initializes the model's relevance parameters, which are subsequently refined end-to-end using forecasting supervision, thereby allowing the learning dynamics to recover from independence-test errors.

\subsection{Causal sufficiency and latent confounding: the boundary of DAG-based discovery}
PC operates in the standard ``fully observed'' DAG regime: causal sufficiency (no unmeasured confounders among the modeled variables), together with the Markov and faithfulness assumptions, yields a well-defined mapping from independencies to a Markov equivalence class.
When causal sufficiency is violated, Pearl's framework moves to more general mixed graphs and algorithms such as FCI, reflecting a fundamental identifiability limit of observational data with hidden variables rather than a defect of any particular learner.
In practice, any discovery method operating only on observed time series inherits this limitation; hence we emphasize the use of the discovered structure as a prior and assess robustness empirically under controlled perturbations and alternative discovery procedures.

\subsection{Implications for robustness and interpretability in causal-guided forecasting}
The graphical theory above implies two practical takeaways relevant to our forecasting framework:

\paragraph{Robustness should be evaluated at the level of independence patterns.}
Since the identifiable causal content is determined by skeleton/v-structures (i.e., the independence model), robustness claims should not hinge on exact edge orientations that are observationally indistinguishable.
Empirically, this motivates perturbation studies and cross-algorithm checks that quantify how changes in the learned structure affect forecasting.

\paragraph{Interpretability should be conservative.}
Following Def.~2.3.6, causal relevance is justified only when it is stable across minimal structures compatible with the observed independencies.
Accordingly, our causal guidance is designed to be conservative: the discovered structure provides a principled inductive bias, while the forecasting objective and regularization determine how strongly the model relies on each proposed relation.

\subsection{PC implement details}
In this section, we detail the Peter-Clark (PC) algorithm (Algorithm~\ref{alg:PC}) used for causal discovery. The PC algorithm assumes that the data are sampled from a faithful joint distribution $\hat{P}$ over a variable set $V$; that is, every conditional independence in $\hat{P}$ corresponds to d-separation in the true causal DAG, and vice-versa.
The outcome is a completed partially directed acyclic graph (CPDAG), denoted $H(\hat{P})$. A CPDAG represents the entire Markov-equivalence class of the true (but unobserved) causal DAG: edges that are compelled in every member of the class appear directed, whereas reversible edges remain undirected. In this mixed graph, oriented edges encode compelled causal directions, whereas undirected edges denote Markov-equivalent ambiguities.

\begin{algorithm}[htpb]
\caption{Causal Discovery Algorithm-PC}
\label{alg:PC}
\textbf{Input}: $\hat{P}$, a stable distribution on a set $V$ of variables;\par
\textbf{Output}: A pattern $H(\hat{P})$ compatible with $\hat{P}$.\par
\begin{algorithmic}[1]
\STATE Initialize complete undirected graph $G$ on $V$
\STATE depth $\gets 0$
\REPEAT
   \FOR{each ordered pair $(a,b)$ adjacent in $G$}
      \FOR{each $\textit{S}\subseteq \mathrm{Adj}(a)\setminus\{b\}$ with $|\textit{S}|=$depth}
         \IF{$a \Vbar b \mid \textit{S}$ in $\hat P$}
             \STATE Remove edge $a{-}b$ from $G$;\quad $\mathrm{Sepset}[a][b]\gets\textit{S}$
             \STATE \textbf{break}
         \ENDIF
      \ENDFOR
   \ENDFOR
   \STATE depth $\gets$ depth+1
\UNTIL{no edge removed at current depth}
\FOR{each non-adjacent $a,b$ with common neighbor $c$}
   \IF{$c\notin\mathrm{Sepset}[a][b]$}
      \STATE Orient $a\to c\gets b$ \hfill(v-structure)
   \ENDIF
\ENDFOR
\WHILE{any Meek rule applies without creating a cycle}
   \STATE Orient the corresponding edge
\ENDWHILE
\STATE return $G$ as CPDAG $H(\hat P)$
\end{algorithmic}
\end{algorithm}

\section{Proofs}\label{appproof}
\paragraph{Proofs of Conditional Independence for Paths}\label{proof4Group}
Based on the analysis detailed in Appendix \ref{background_causal}, we exhaust all categories of relationships among other variables and the target variable in Figure \ref{f1d}, obtaining: 1) \textbf{Path a:} $V_1 \to V_i$. Contains only two variables, $V_1$ points to $V_i$, and there are no other variables connected to the left of $V_1$ forming a chain ($V_2 \to V_1 \to V_i$) or a fork ($V_2 \gets V_1 \to V_i$). In this case, $V_1 \nVbar V_i$; 2) \textbf{Path b:} ${V_D} - \cdots  - {V_{i + 1}} - {V_{i - 1}} -  \cdots  - {V_3} - {V_2} \to {V_1} \to {V_i}$. Here, ``$ - $'' represents that the casual relationship between two variables is unclear, e.g., ``$ - $'' can be either ``$ \to $'' or ``$  \leftarrow  $'', but we are not sure whether it is ``$ \to $'' or ``$  \leftarrow  $''. In this case, when given $\mathcal{Z} = \{{V_1}, \cdots ,{V_{i - 1}},{V_{i + 1}}, \cdots {V_D}\}$, we can obtain that $\{{V_1} \nVbar {V_i}, {V_j} \Vbar {V_i} \}\mid \mathcal{Z}$, where $V_j \in \mathcal{Z}\setminus V_1 $. Thus, \textbf{Path b} equals to ${V_1} \to {V_i}$; 3) \textbf{Path c:} ${V_D}- \cdots  - {V_{i + 1}} - {V_{i - 1}} -  \cdots  - {V_3} - {V_2} \gets {V_1} \to {V_i}$. In this case, when given $\mathcal{Z} = \{{V_1}, \cdots ,{V_{i - 1}},{V_{i + 1}}, \cdots {V_D}\}$, we can obtain that $\{{V_1} \nVbar {V_i}, {V_j} \Vbar {V_i} \}\mid \mathcal{Z}$, where $V_j \in \mathcal{Z}\setminus V_1 $. Thus, \textbf{Path c} equals to ${V_1} \to {V_i}$; 4) \textbf{Path d:} $V_i \to V_1$. Contains only two variables, $V_i$ points to $V_1$, and there are no other variables to the right of $V_1$  forming a chain ($V_i \to V_1 \to V_2$) or a collider ($V_i \to V_1 \gets V_2$). In this case, $V_1 \nVbar V_i$;  5) \textbf{Path e:} $V_i \to V_1 \to V_2 - {V_3} -  \cdots - {V_{i - 1}} - {V_{i + 1}} -  \cdots - {V_D}$. In this case, when given $\mathcal{Z} = \{{V_1}, \cdots ,{V_{i - 1}},{V_{i + 1}}, \cdots {V_D}\}$, we can obtain that $\{{V_1} \nVbar {V_i}, {V_j} \Vbar {V_i} \}\mid \mathcal{Z}$, where $V_j \in \mathcal{Z}\setminus V_1  $. Thus, \textbf{Path e} equals to ${V_i} \to {V_1}$; 6) \textbf{Path f:} $V_i \to V_1 \gets V_2 - {V_3} -  \cdots - {V_{i - 1}} - {V_{i + 1}} -  \cdots - {V_D}$. In this case, when given $\mathcal{Z} = \{ {V_1}, \cdots ,{V_{i - 1}},{V_{i + 1}}, \cdots {V_D}\}$, we can obtain that $\{{V_1} \nVbar {V_i}, {V_2} \nVbar {V_i}, {V_j} \Vbar {V_i} \}\mid \mathcal{Z}$, where $V_j \in \mathcal{Z} \setminus \{V_1,V_2\} $. Thus, \textbf{Path f} equals to ${V_i} \to {V_1} \gets {V_2}$.

Let $\mathcal{V} = \{V_1, V_2, \cdots, V_D\}$ denote the full set of variables. We focus on a target variable $V_i$ and consider its interactions with other variables via six typical causal paths (Path a to Path f). We use the notation $\mathcal{S}_A \Vbar \mathcal{S}_B \mid \mathcal{S}_C$ to represent that $\mathcal{S}_A$ is conditionally independent of $\mathcal{S}_B$ given $\mathcal{S}_C$.

\textbf{Path a:} $V_1 \to V_i$. This is a direct causal link from $V_1$ to $V_i$. The joint distribution factorizes as $P(V_1, V_i) = P(V_1) P(V_i \mid V_1)$. Then the marginal and conditional probabilities are $P(V_i) = \int P(V_i \mid V_1) P(V_1) dV_1$. $P(V_i \mid V_1) \ne P(V_i)$ for any value of $V_1$, then $ P(V_i \mid V_1) \ne P(V_i) \Rightarrow  V_1 \nVbar V_i$.

\textbf{Path b:} ${V_D} - \cdots  - {V_{i + 1}} - {V_{i - 1}} -  \cdots  - {V_3} - {V_2} \to {V_1} \to {V_i}$. This structure represents a causal chain beginning at $V_2$ and ending at $V_i$, where $V_3, \cdots,V_{i-1},V_{i+1},\cdots, V_D$ are unclear path direction variables in the left of $V_2$.

% The conditional distribution of $V_i$ given all other variables is: $P(V_i \mid V_1, V_2,\cdots,V_{i-1},V_{i+1} \dots, V_D) = P(V_i \mid V_1)$, since $V_1$ d-separates $V_i$ from all upstream variables of $V_1$. For any $j \in \{2,3,\dots,i-1,i+1,\dots,D\}$, the conditional distribution satisfies $P(V_i \mid V_1, V_j) = P(V_i \mid V_1)$, because the path $V_j \to \cdots \to V_1 \to V_i$ is blocked by $V_1$. Therefore, we obtain $\{V_1 \nVbar V_i,V_j\Vbar V_i \}\mid V1,\cdots,V_{i-1},V_{i+1},\cdots,V_D$, where $j \in \{2,3,\cdots,i-1,i+1,\cdots,D\}$. This path simplifies to the direct influence $V_1 \to V_i$.
Let $\mathcal{Z}=\mathcal{V}\setminus\{V_1,V_i\}$. By d-separation, $V_1$ blocks all paths from $V_j$ to $V_i$, so $P(V_i \mid V_1, V_j, \mathcal{Z}\setminus\{V_j\})
= P(V_i \mid V_1)$. Hence $V_i \Vbar V_j \mid \{V_1\}\cup (\mathcal{Z}\setminus\{V_j\})$ and $ V_i \nVbar V_1 \mid \mathcal{Z}$. Thus the entire structure simplifies to the direct influence $V_1\to V_i$.

\textbf{Path c:} ${V_D} - \cdots - V_3 - V_2 \gets V_1 \to V_i$. This is a fork structure where $V_1$ is a common cause of both $V_2$ and $V_i$, and the path may include $V_3, \cdots,V_{i-1},V_{i+1},\cdots, V_D$ connected to $V_2$ from the left with unclear direction paths. 

% Conditioning on $V_1$ blocks the dependency between $V_2$ and $V_i$: $P(V_i \mid V_1,V_2,\cdots,V_{i-1},V_{i+1},\cdots,V_D) = P(V_i \mid V_1)$. Thus $\{V_1 \nVbar V_i,V_j\Vbar V_i \}\mid V1,\cdots,V_{i-1},V_{i+1},\cdots,V_D$, where $j \in \{2,3,\cdots,i-1,i+1,\cdots,D\}$. This path simplifies to the direct influence $V_1 \to V_i$.
Let $\mathcal{Z}=\mathcal{V}\setminus\{V_1,V_i\}$. Conditioning on $V_1$ d-separates $V_2$ (and all its upstream nodes) from $V_i$, so $P(V_i\mid V_1, V_2, \mathcal{Z}\setminus\{V_2\})= P(V_i\mid V_1)$. Thus $V_i \Vbar V_2 \mid \{V_1\}\cup(\mathcal{Z}\setminus\{V_2\})$ and $V_i \nVbar V_1 \mid \mathcal{Z}$. Hence Path c also reduces to $V_1\to V_i$.

\textbf{Path d:} $V_i \to V_1$. This is a direct causal link from $V_i$ to $V_1$. The joint distribution factorizes as $P(V_1, V_i) = P(V_i) P(V_1 \mid V_i)$. Then the marginal and conditional probabilities are $P(V_i) = \int P(V_i \mid V_1) P(V_1) dV_1$. $P(V_i \mid V_1) \ne P(V_i)$ for any value of $V_1$, then $ P(V_i \mid V_1) \ne P(V_i) \Rightarrow  V_1 \nVbar V_i$.

\textbf{Path e:} $V_i \to V_1 \to V_2 - V_3 - \cdots - V_D$
This is a chain structure originating from $V_i$ and propagating through intermediate nodes. let $\mathcal{Z}=\mathcal{V}\setminus\{V_i\}$. $P(V_1 \mid V_i, \mathcal{Z} \setminus \{V_1\}) \ne P(V_1 \mid \mathcal{Z} \setminus \{V_1\})  \Rightarrow V_1 \nVbar V_i \mid \mathcal{Z}$.
For $j \ge 2$, the dependency is blocked by known $V_1$:
$P(V_j \mid V_i, \mathcal{Z} \setminus \{V_j\}) = P(V_j \mid \mathcal{Z} \setminus \{V_j\}) \Rightarrow V_j \Vbar V_i \mid \mathcal{Z}$.
Thus, the path reduces to $V_i \to V_1$.

\textbf{Path f:} $V_i \to V_1 \gets V_2 - V_3 - \cdots - V_D$. This is a collider structure where $V_1$ is a common result of $V_i$ and $V_2$, and $V_2$ may be influenced by the right variables with unclear direction paths. let $\mathcal{Z}=\mathcal{V}\setminus\{V_i\}$.

Since $V_1$ is observed, the dependency between $V_i$ and $V_2$ is activated: $P(V_i \mid \mathcal{Z}) \ne P(V_i \mid \mathcal{Z} \setminus \{V_1, V_2\}) \Rightarrow V_2\nVbar V_i\mid\mathcal{Z}$. Meanwhile, $P(V_i\mid \mathcal{Z})= P(V_i\mid \mathcal{Z}\setminus\{V_j\})\Rightarrow V_j\Vbar V_i\mid\mathcal{Z} \forall j \in \{3,\cdots,i-1,i+1,\cdots,D\}$. Hence both $V_1$ and $V_2$ are relevant under conditioning, confirming the collider structure $V_i\to V_1\gets V_2$.

\paragraph{Proof of Theorem \ref{generalize}}\label{proof}
\begin{proof}
The conditional expectation $\Pi:Z \in L^2(\Omega) \mapsto \mathbb{E}[Z \vert \mathcal{S}_{V_s}]$ defines an orthogonal projection onto the space of $\mathcal{S}_{V_s}$-measurable random variables with finite variance $L^2(\Omega, \sigma(\mathcal{S}_{V_s}), P)$. Thus, its range and null space are orthogonal in $L^2(\Omega)$.
Let $f \in L^2(V)$. We have $\Phi f(\mathcal{S}_{V_c},\mathcal{S}_{V_s})= \mathbb{E}[f(\mathcal{S}_{V_c},\mathcal{S}_{V_s}) \vert \mathcal{S}_{V_s}] - C = \Pi f(\mathcal{S}_{V_c},\mathcal{S}_{V_s})-C$ hence $\Phi f(\mathcal{S}_{V_c},\mathcal{S}_{V_s})$ is in the range of $\Pi$. On the other hand,
\begin{equation}
    \resizebox{0.8\linewidth}{!}{$
    \begin{array}{l}
     \mathbb{E}[\Psi f(\mathcal{S}_{V_c},\mathcal{S}_{V_s}) \vert \mathcal{S}_{V_s}] = \mathbb{E}[f(\mathcal{S}_{V_c},\mathcal{S}_{V_s}) \vert \mathcal{S}_{V_s}]- \mathbb{E}[\Phi f(\mathcal{S}_{V_c},\mathcal{S}_{V_s}) \vert \mathcal{S}_{V_s}] \\
     = \mathbb{E}[f(\mathcal{S}_{V_c},\mathcal{S}_{V_s}) \vert \mathcal{S}_{V_s}]- \mathbb{E}[f(\mathcal{S}_{V_c},\mathcal{S}_{V_s}) \vert \mathcal{S}_{V_s}]=0.
    \end{array}
    $}
\end{equation}
Therefore, $\Psi f(\mathcal{S}_{V_c},\mathcal{S}_{V_s})$ is in the null space of $\Pi$. Finally, because $V_i \Vbar \mathcal{S}_{V_s}$ we have $\mathbb{E}[V_i \vert \mathcal{S}_{V_s}]= \mathbb{E}[V_i]=C$ by assumption, therefore $V_i$ is also in the null space of $\Pi$.

Hence, adopting this random variable view, the desired result simply follows from $L^2(\Omega)$ orthogonality:
\begin{equation}
\resizebox{0.8\linewidth}{!}{$
    \begin{array}{l}
    \Delta(f,\Psi f) = \mathbb{E}[(V_i-f(\mathcal{S}_{V_c},\mathcal{S}_{V_s}))^2]- \mathbb{E}[(V_i-\Psi f(\mathcal{S}_{V_c},\mathcal{S}_{V_s}))^2] \\
    = \Vert V_i-f(\mathcal{S}_{V_c},\mathcal{S}_{V_s}) \Vert_{L^2(\Omega)}^2- \Vert V_i-\Psi f(\mathcal{S}_{V_c},\mathcal{S}_{V_s}) \Vert_{L^2(\Omega)}^2 \\ 
    = \Vert V_i-\Psi f(\mathcal{S}_{V_c},\mathcal{S}_{V_s})-\Phi f(\mathcal{S}_{V_c},\mathcal{S}_{V_s}) \Vert_{L^2(\Omega)}^2- \Vert V_i-\Psi f(\mathcal{S}_{V_c},\mathcal{S}_{V_s}) \Vert_{L^2(\Omega)}^2 \\
    = \Vert V_i-\Psi f(\mathcal{S}_{V_c},\mathcal{S}_{V_s}) \Vert_{L^2(\Omega)}^2+ \Vert \Phi f(\mathcal{S}_{V_c},\mathcal{S}_{V_s}) \Vert_{L^2(\Omega)}^2      - \Vert V_i-\Psi f((\mathcal{S}_{V_c},\mathcal{S}_{V_s}) \Vert_{L^2(\Omega)}^2 \\
    = \mathbb{E}[\Phi f(\mathcal{S}_{V_c},\mathcal{S}_{V_s})^2] 
    = \Vert \Phi f \Vert_{L^2(\Omega)}^2.
\end{array}
$}
\end{equation}
\end{proof}

\section{Implementation Details}\label{appimple}
\textbf{Environment and Optimization.} All experiments are implemented in PyTorch \citep{Pytorch} and conducted on NVIDIA V100 32GB GPUs. We use the Adam optimizer \citep{Adam} with a learning rate selected from $\{10^{-3}, 10^{-4}\}$ and a batch size from $\{32, 64, 128\}$. Hyperparameters are tuned via grid search based on the validation set performance.

\textbf{Model Configuration.} For the Causal Decomposition Transformer (CDT), the backbone consists of a stack of Transformer encoder layers, with the depth selected from $\{1, 2, 3\}$ and the hidden dimension $d$ selected from $\{64, 128, 256, 512\}$. The dynamic causal adapter is initialized using the PC algorithm implemented in the causal-learn package \citep{causal-learn}, with a significance level of $\alpha \in \{0.5, 1,2,3\}$. The regularization coefficient $\lambda$ in the loss function is set to $0.2$. The pseudo code is Algorithm \ref{alg:PseudoCode}.

\textbf{Training Protocol.} Models are trained for up to 10 epochs. The training objective is the composite loss, while the evaluation metric for model selection is the Mean Squared Error (MSE) on the validation set. We employ an early stopping strategy that terminates training if the validation loss does not improve for three consecutive epochs. The checkpoint with the lowest validation loss is then evaluated on the test set. To ensure reproducibility and robustness, all experiments are repeated five times with different random seeds, and we report the average MSE and Mean Absolute Error (MAE).

\begin{algorithm}[htpb]
\caption{Pseudo-Code of Causal Decomposition Transformer (CDT)}
\label{alg:PseudoCode}
\small
\textbf{Input}: Multivariate time series $X \in \mathbb R^{T\times D}$, Ground truth $Y \in \mathbb R^{S\times D}$. \\
\textbf{Components}: Feature Encoder $\text{Enc}(\cdot)$, Transformer Backbone $\mathcal{T}$, Projection Heads $g_d, g_c, g_s$, Prediction Head $f_{\text{pred}}$, Learnable Logits $W_{\text{DCS}}, W_{\text{CCS}}, W_{\text{SP}}$. \par
\textbf{Initialization}: Derive $A_{\text{init}}$ via PC algorithm; Initialize $W_{\text{DCS}}, W_{\text{CCS}}, W_{\text{SP}}$ using $A_{\text{init}}$ as prior.

\begin{algorithmic}[1]
\STATE \textbf{Step 1: Dynamic Causal Adapter (DCA)}
\STATE $H_{\text{enc}} = \text{Enc}(X)$ \hfill $\triangleright$ \textit{Encode all variable histories independently}
\FOR{each target variable $i \in \{1, \dots, D\}$}
    \STATE \textbf{Relevance Weighting}:
    \STATE $\pi_{\text{DCS}}^{(i)} = \sigma(W_{\text{DCS}}[:, i]), \quad \pi_{\text{CCS}}^{(i)} = \sigma(W_{\text{CCS}}[:, i])$
    \STATE \textbf{Feature Aggregation}:
    \STATE $H_{\text{DCS}}^i = \sum_{j} \pi_{\text{DCS}}^{(i)}(j) \cdot g_d(H_{\text{enc}}[j])$
    \STATE $H_{\text{CCS}}^i = \sum_{j} \pi_{\text{CCS}}^{(i)}(j) \cdot g_c(H_{\text{enc}}[j])$
    \STATE $H_{\text{ES}}^i = H_{\text{enc}}[i]$
\ENDFOR

\STATE \textbf{Step 2: segment constrained Transformer Processing}
\FOR{each target variable $i \in \{1, \dots, D\}$}
    \STATE \textbf{Serialization}: $S_i = \text{Concat}([H_{\text{DCS}}^i, H_{\text{CCS}}^i, H_{\text{ES}}^i], \text{dim}=\text{time})$
    \STATE \textbf{Mask Construction}: Construct segment constrained mask $M$ to enforce independence between DCS and CCS segments.
    \STATE \textbf{Forward Pass}: $Z_i = \mathcal{T}(S_i, \text{mask}=M)$
    \STATE \textbf{Raw Prediction}: $\hat{Y}_{\text{raw}}^i = f_{\text{pred}}(Z_i[\text{endogenous\_tokens}])$
\ENDFOR

\STATE \textbf{Step 3: Causal-Constrained Output Projection}
\STATE Assemble raw predictions: $\hat{Y}_{\text{raw}} = \{\hat{Y}_{\text{raw}}^1, \dots, \hat{Y}_{\text{raw}}^D\}$
\FOR{each target variable $i \in \{1, \dots, D\}$}
    \STATE \textbf{Spouse Aggregation}: $H_{\text{SP}}^i = \sum_{j} \sigma(W_{\text{SP}}[j, i]) \cdot g_s(H_{\text{enc}}[j])$
    \STATE \textbf{Bias Estimation}: $B^i = \Phi(\hat{Y}_{\text{raw}}^i, H_{\text{SP}}^i)$ \hfill $\triangleright$ \textit{Estimate conditional expectation}
    \STATE \textbf{Final Correction}: $\hat{Y}^i = \hat{Y}_{\text{raw}}^i - B^i$
\ENDFOR

\STATE \textbf{Optimization}: 
\STATE Compute Loss: $\mathcal{L} = \text{MSE}(\hat{Y}, Y) + \lambda \sum_{W \in \{W_{\text{DCS}}, \dots\}} \mathcal{L}_{\text{prior}}(W, A_{\text{init}})$
\STATE Update all learnable parameters via backpropagation.
\STATE \textbf{Output:} Predicted future values $\hat{Y}$
\end{algorithmic}
\end{algorithm}

\section{Dataset Descriptions}\label{appendixdatasets}
In this paper, we conduct tests using eight real-world datasets. These datasets include:\par
(1) The ETT dataset contains 7 factors of electricity transformer from July 2016 to July 2018, consists of two sub-datasets, ETT1 and ETT2, collected from electricity transformers at two different stations. Each sub-dataset is available in two resolutions (15 minutes and 1 hour), containing multiple load series and a single oil temperature series.
(2) Weather covers 21 meteorological variables recorded at 10-minute intervals throughout the year 2020. The data was collected by the Max Planck Institute for Biogeochemistry's Weather Station, providing valuable meteorological insights.
(3) Exchange Rate contains daily currency exchange rates for eight countries, spanning from 1990 to 2016.
(4) ECL contains the electricity consumption of 370 clients for short-term forecasting while it contains the electricity consumption of 321 clients for long-term forecasting. It is collected since 01/01/2011. The data sampling interval is every 15 minutes;
(5) Traffic gathers hourly road occupancy rates from 862 sensors on San Francisco Bay area freeways, covering the period from January 2015 to December 2016.
% (4) Solar-Energy documenting the solar energy output from 137 photovoltaic (PV) plants in 2006, with data points every 10 minutes.
% (5) ECL records the hourly electricity consumption data of 321 clients.
% (6) PEMS contains the public traffic network data in California collected by 5-minute windows. We use the same two public subsets (PEMS04 and PEMS08).
\par
                 
We follow the same data processing and chronological train-validation-test split protocol as used in iTransformer \citep{iTransformer} to avoid data leakage issues. The details of the datasets are provided in Table \ref{tab:dataset}.

\begin{table*}[htpb]
  \centering
  \caption{Detailed descriptions of datasets. \emph{Dim} denotes the number of variables in each dataset. \emph{Prediction Length} denotes the number of future time points to predict; each dataset includes four different forecasting horizons. \emph{Time steps} represents the number of time points. \emph{Percentage} indicates the proportions of the dataset allocated to Train, Validation, and Test splits. \emph{Frequency} specifies the sampling interval between consecutive time points.}
  \vspace{0.1in}
  \label{tab:dataset}
 \resizebox{0.99\linewidth}{!}{
  \begin{tabular}{l|c|c|c|c|c|c}
    \toprule
    Dataset & Dim & Prediction Length & Time steps & Percentage & Frequency& Information \\
    \toprule
     {ETTh1,ETTh2} & 7 & \scalebox{0.8}{\{96, 192, 336, 720\}} & 17420 & (60\%, 20\%, 20\%) & Hourly & Electricity\\
     \midrule
     {ETTm1,ETTm2} & 7 & \scalebox{0.8}{\{96, 192, 336, 720\}} & 69680 & (60\%, 20\%, 20\%) & 15min & Electricity\\
     \midrule
    {Exchange} & 8 & \scalebox{0.8}{\{96, 192, 336, 720\}} & 7588 & (70\%, 10\%, 20\%) & Daily & Economy \\
    \midrule
    Weather & 21 & \scalebox{0.8}{\{96, 192, 336, 720\}} & 52560& (70\%, 10\%, 20\%) & 10min & Weather\\
    % \midrule
    % Solar-Energy & 137 & \scalebox{0.8}{\{96, 192, 336, 720\}} & 52560& (70\%, 10\%, 20\%) & 10min & Energy\\
    \midrule
    ECL & 321 & \scalebox{0.8}{\{96, 192, 336, 720\}}  &26304& (70\%, 10\%, 20\%) & Hourly & Electricity\\
    \midrule
    Traffic & 862 & \scalebox{0.8}{\{96, 192, 336, 720\}}  &17451& (70\%, 10\%, 20\%) & Hourly & Transportation\\
    % \midrule
    % PEMS04 & 307 & \scalebox{0.8}{\{12, 24, 72, 144\}}  &16992 & (60\%, 20\%, 20\%) & 5min & Transaction\\
    % \midrule
    % PEMS08 & 170 & \scalebox{0.8}{\{12, 24, 72, 144\}}  &17856 & (60\%, 20\%, 20\%) & 5min & Transaction\\
    \bottomrule
    \end{tabular}}
    
\end{table*}

\section{Additional Experiments}\label{app_exp}
We provide additional experiments to evaluate CDT.

\subsection{Full results}
The full multivariate forecasting results are provided in Table \ref{tab:main_result}. All results are averaged over five random seeds.

\begin{table*}[t]
  % \vspace{-5pt}
  \renewcommand{\arraystretch}{1} %行间距
  \centering
      \caption{Multivariate time series forecasting results with prediction lengths $S\in\left \{96, 192, 336, 720\right \}$ and fixed lookback length $T = 96$. The best results in \textbf{bold}, the second \underline{underlined}, and “–” denote metrics not reported in the original papers. The lower MSE/MAE indicates a more accurate prediction result. }
 \resizebox{\linewidth}{!}{
      \begin{tabular}{cc|cc|cc|cc|cc|cc|cc|cc|cc|cc|cc|cc|cc|cc|cc|cc|cc}
    \toprule
    \multicolumn{2}{c}{\multirow{1}{*}{Models}} & 
    \multicolumn{2}{c}{\rotatebox{0}{CDT}} &
    \multicolumn{2}{c}{\rotatebox{0}{Numerion}} &
    \multicolumn{2}{c}{\rotatebox{0}{DistDF}} &
    \multicolumn{2}{c}{\rotatebox{0}{GTR}} &
    \multicolumn{2}{c}{\rotatebox{0}{TimePro}} &
    \multicolumn{2}{c}{\rotatebox{0}{SEMPO}} &
    \multicolumn{2}{c}{\rotatebox{0}{TFPS}} &
    \multicolumn{2}{c}{\rotatebox{0}{{iTransformer}}} &
    \multicolumn{2}{c}{\rotatebox{0}{{PatchTST}}} &
    \multicolumn{2}{c}{\rotatebox{0}{{RLinear}}}  &
    \multicolumn{2}{c}{\rotatebox{0}{{Crossformer}}}  &
    \multicolumn{2}{c}{\rotatebox{0}{{TiDE}}} &
    \multicolumn{2}{c}{\rotatebox{0}{{{TimesNet}}}} &
    \multicolumn{2}{c}{\rotatebox{0}{{DLinear}}}&
    % \multicolumn{2}{c}{\rotatebox{0}{{SCINet}}} &
    \multicolumn{2}{c}{\rotatebox{0}{{FEDformer}}} &
    % \multicolumn{2}{c}{\rotatebox{0}{{Stationary}}} &
    \multicolumn{2}{c}{\rotatebox{0}{{Autoformer}}} \\
    \multicolumn{2}{c}{} &
    \multicolumn{2}{c}{{\textbf{(Ours)}}} &
    \multicolumn{2}{c}{{(\citeyear{Numerion}})} &
    \multicolumn{2}{c}{{(\citeyear{DistDF}})} &
    \multicolumn{2}{c}{{(\citeyear{GTR}})} &
    \multicolumn{2}{c}{{(\citeyear{timepro}})} &
    \multicolumn{2}{c}{{(\citeyear{SEMPO}})} &
    \multicolumn{2}{c}{{(\citeyear{TFPS}})} &
    \multicolumn{2}{c}{{(\citeyear{iTransformer})}} & 
    \multicolumn{2}{c}{{(\citeyear{PatchTST})}} & 
    \multicolumn{2}{c}{{(\citeyear{RLinear})}} &
    \multicolumn{2}{c}{{(\citeyear{Crossformer})}}  & 
    \multicolumn{2}{c}{{(\citeyear{TiDE})}} & 
    \multicolumn{2}{c}{{(\citeyear{TimesNet})}} & 
    \multicolumn{2}{c}{{(\citeyear{DLinear})}}& 
    % \multicolumn{2}{c}{{\citep{SCINet}}} &
    \multicolumn{2}{c}{{(\citeyear{FEDformer})}} &
    % \multicolumn{2}{c}{{\citep{NSTransformer}}} &
    \multicolumn{2}{c}{{(\citeyear{Autoformer})}} \\
     \cmidrule(lr){3-4} \cmidrule(lr){5-6}\cmidrule(lr){7-8} \cmidrule(lr){9-10}\cmidrule(lr){11-12}\cmidrule(lr){13-14} \cmidrule(lr){15-16} \cmidrule(lr){17-18} \cmidrule(lr){19-20}  \cmidrule(lr){21-22} \cmidrule(lr){23-24} \cmidrule(lr){25-26} \cmidrule(lr){27-28} \cmidrule(lr){29-30} \cmidrule(lr){31-32} \cmidrule(lr){33-34}
    \multicolumn{2}{c}{Metric}  & {MSE} & {MAE}  & MSE & MAE & MSE & MAE & MSE & MAE & {MSE} & {MAE}  & {MSE} & {MAE}  & {MSE} & {MAE}  & {MSE} & {MAE} & {MSE} & {MAE} & {MSE} & {MAE} & {MSE} & {MAE} & {MSE} & {MAE} & {MSE} & {MAE} & {MSE} & {MAE} & {MSE} & {MAE} & {MSE} & {MAE} \\
    \toprule

    \multirow{5}{*}{{\rotatebox{90}{{ETTm1}}}}
    &  {96} & \underline{{0.307}} & \textbf{{0.334}} & \textbf{0.305}& \underline{0.337}  & {0.316} & {0.357} & {0.316} & {0.357} & {0.326} & {0.364} & {0.466} & {0.443} & \underline{0.327} & {0.367} & {{0.334}} & {{0.368}} &{0.329} &{0.367} & {0.355} & {0.376} & {0.404} & {0.426} & {0.364} & {0.387} &{{0.338}} &{{0.375}} &{{0.345}} &{{0.372}}  &{0.379} &{0.419} &{0.505} &{0.475} \\ 
    & {192} & \textbf{0.351} & {\textbf{0.357}} & \underline{0.356} & \underline{0.367} & {0.358} & {0.380} & {0.358} & {0.380} & {0.367} & \underline{0.383} & {0.484} & {0.455} & {0.374} & {0.395} & {0.377} & {0.391} &{0.367} &{0.385} & {0.391} & {0.392} & {0.450} & {0.451} &{0.398} & {0.404} &{{0.374}} &{{0.387}}  &{{0.380}} &{{0.389}}  &{0.426} &{0.441} &{0.553} &{0.496} \\ 
    & {336} & \textbf{{0.374}} & \textbf{{0.382}}  & \underline{0.380} & \underline{0.386} & {0.392} & {0.404} & {0.392} & {0.404}& {0.402} & {0.409} & {0.506} & {0.469} & {0.401} & {0.408} & {0.426} & {0.420} &{0.399} &{0.410} & {0.424}  &{0.415} & {0.532}  &{0.515} & {0.428} & {0.425} &{{0.410}} &{{0.411}}  &{{0.413}} &{{0.413}}   &{0.445} &{0.459} &{0.621} &{0.537} \\ 
    & {720} & \textbf{{0.431}} &\underline{{0.427}} & \underline{0.439}& \textbf{0.423}  & {0.448} & {0.437} & {0.448} & {0.437} & {0.469} & {0.446} & {0.557} & {0.498} & {0.479} & {0.456} & {0.491} & {0.459} &{0.454} &{0.439} & {0.487} & {0.450} & {0.666} & {0.589} & {0.487} & {0.461} &{{0.478}} &{{0.450}} &{{0.474}} &{{0.453}} &{0.543} &{0.490} &{0.671} &{0.561} \\ 
    \cmidrule(lr){2-34}
    & {Avg} &   \textbf{{0.365}} & \textbf{{0.375}} & \underline{0.370} & \underline{0.378} & {0.378} & {0.394} & {0.378} & {0.394} & 0.391 & {0.400} & 0.503 & {0.466} & 0.395 & {0.406} & {0.407} & {0.410} &{0.387} &\underline{0.400} & {0.414} & {0.407} & {0.513} & {0.496} & {0.419} & {0.419} &{{0.400}} &{{0.406}}  &{{0.403}} &{{0.407}} &{0.448} &{0.452}  &{0.588} &{0.517} \\  
    \midrule
    
    \multirow{5}{*}{\rotatebox{90}{{ETTm2}}}
    &  {96} & \textbf{{0.168}} & \underline{{0.255}} & \underline{0.170} & \textbf{0.249} & {0.174} & {0.256} & {0.174} & {0.256} & {0.178} & {0.260} & {0.196} & {0.286} & \underline{0.170} & \underline{0.255} & {0.180} & {0.264} &{0.175} &{0.259} & {0.182} & {0.265} & {0.287} & {0.366} & {0.207} & {0.305} &{{0.187}} &{0.267} &{0.193} &{0.292} &{0.203} &{0.287} &{0.255} &{0.339} \\ 
    & {192} & \textbf{{0.230}} & \underline{{0.296}} & \underline{0.231} & \textbf{0.292} & {0.239} & {0.298} & {0.239} & {0.298} & {0.242} & {0.303} & {0.252} & {0.323} & {0.235} & {0.296} & {0.250} & {0.309} &{0.241} &{0.302} & {0.246} & {0.304} & {0.414} & {0.492} & {0.290} & {0.364} &{{0.249}} &{{0.309}} &{0.284} &{0.362}&{0.269} &{0.328} &{0.281} &{0.340} \\ 
    & {336} & \underline{{0.290}} & \underline{{0.333}} & \textbf{0.289} & \textbf{0.330} & {0.300} & {0.338} & {0.300} & {0.338} & {0.303} & {0.342} & {0.306} & {0.354} & {0.297} & {0.335} & {{0.311}} & {{0.348}} &{0.305} &{0.343} & {0.307} & {0.342} & {0.597} & {0.542}  & {0.377} & {0.422} &{{0.321}} &{{0.351}} &{0.369} &{0.427} &{0.325} &{0.366} &{0.339} &{0.372} \\ 
    & {720} & \textbf{{0.387}} & \textbf{{0.383}} & \underline{0.390}& \underline{0.387}  & {0.397} & {0.394} & {0.397} & {0.394} & {0.400} & {0.399} & {0.391} & {0.404} & {0.401} & {0.397}& {{0.412}} & {{0.407}} &{0.402} &{0.400} & {0.407} & {0.398} & {1.730} & {1.042} & {0.558} & {0.524} &{{0.408}} &{{0.403}} &{0.554} &{0.522} &{0.421} &{0.415} &{0.433} &{0.432} \\ 
    \cmidrule(lr){2-34}
    & {Avg} & \textbf{{0.268}} & \underline{{0.316}} & \underline{0.270} & \textbf{0.315} & {0.277} & {0.321} & {0.277} & {0.321}  & {0.281} & {0.326}  & {0.286} & {0.341}  & {0.276} & {0.321}  & {{0.288}} & {{0.332}} &\underline{0.281} &{0.326} & {0.286} & {0.327} & {0.757} & {0.610} & {0.358} & {0.404} &{{0.291}} &{{0.333}} &{0.350} &{0.401} &{0.305} &{0.349} &{0.327} &{0.371} \\  
    \midrule

    \multirow{5}{*}{\rotatebox{90}{{{ETTh1}}}}
    &  {96} & \textbf{0.352} & \textbf{{0.379}} & \underline{0.359} & \underline{0.380} & {0.373} & {0.393} & {0.373} & {0.393} & {0.375} & {0.398} & {0.384} & {0.408} & {0.398} & {0.413}& {{0.386}} & {{0.405}} &0.414 &0.419 & {0.386} & \textbf{0.395} & {0.423} & {0.448} & {0.479}& {0.464}  &{{0.384}} &{{0.402}} & {0.386} &{{0.400}} &{{0.376}} &{0.419} &{0.449} &{0.459}  \\ 
    & {192} & \textbf{{0.402}} & \textbf{{0.406}} & \underline{0.407} & \underline{0.409} & {0.428} & {0.425} & {0.428} & {0.425} & {0.427} & {0.429} & {0.409} & {0.426} & {0.423} & {0.423}& {0.441} & {0.436} &0.460 &0.445 & {0.437} &{0.424} & {0.471} & {0.474}  & {0.525} & {0.492} &{{0.436}} &{{0.429}}  &{{0.437}} &{{0.432}} &{{0.420}} &{0.448} &{0.500} &{0.482} \\ 
    & {336} & \underline{{0.434}} & \underline{{0.430}} & {0.444} & \textbf{0.429} & {0.466} & {0.445} & {0.466} & {0.445} & {0.472} & {0.450} & \textbf{0.417} & {0.433} & {0.484} & {0.461}& {{0.487}} & {{0.458}} &0.501 &0.466 & {0.479} & {0.446} & {0.570} & {0.546} & {0.565} & {0.515} &{0.491} &{0.469} &{{0.481}} & {{0.459}} &{{0.459}} &{{0.465}} &{0.521} &{0.496} \\ 
    & {720} & \underline{{0.439}} & \textbf{{0.448}} & {0.447} & \underline{0.449} & {0.453} & {0.453} & {0.453} & {0.453} & {0.476} & {0.474} & \textbf{0.432} & {0.454} & {0.488} & {0.476}& {{0.503}} & {{0.491}} &0.500 &0.488 & {0.481} & {0.470} & {0.653} & {0.621} & {0.594} & {0.558} &{0.521} &{{0.500}} &{0.519} &{0.516} &{{0.506}} &{{0.507}} &{{0.514}} &{0.512}  \\ 
    \cmidrule(lr){2-34}
    & {Avg} & \textbf{{0.406}} & \textbf{{0.415}} & {0.414} & \underline{0.417} & {0.430} & {0.429} & {0.430} & {0.429} & {0.438} & {0.438}  & \underline{0.410} & {0.430}  & {0.448} & {0.443}  & {{0.454}} & {{0.447}} &0.469 &0.454 & {0.446} & {0.434} & {0.529} & {0.522} & {0.541} & {0.507} &{0.458} &{{0.450}} &{{0.456}} &{{0.452}} &{{0.440}} &{0.460} &{0.496} &{0.487}  \\
    \midrule

    \multirow{5}{*}{\rotatebox{90}{{ETTh2}}}
    &  {96} & \underline{{0.281}} & \textbf{0.324} & \textbf{0.279}& \underline{0.326}  & {0.287} & {0.336} & {0.290} & {0.342} & {0.293} & {0.345} & {0.282} & {0.342} & {0.313} & {0.355}& {{0.297}} & {{0.349}} &0.302 &0.348 & {0.288} & {0.338} & {0.745} & {0.584} &{0.400} & {0.440}  & {{0.340}} & {{0.374}} &{{0.333}} &{{0.387}} &{0.358} &{0.397} &{0.346} &{0.388} \\
    & {192} & \underline{{0.337}} & \textbf{{0.376}} & {0.359} & \underline{0.379} & {0.358} & {0.381} & {0.362} & {0.389} & {0.367} & {0.394} & \textbf{0.334} & {0.384} & {0.405} & {0.410}& {{0.380}} & {{0.400}} &0.388 &0.400 & {0.374} & {0.390} & {0.877} & {0.656} & {0.528} & {0.509} & {{0.402}} & {{0.414}} &{0.477} &{0.476} &{{0.429}} &{{0.439}} &{0.456} &{0.452} \\ 
    & {336} & \textbf{{0.405}} & {{0.416}} & \underline{0.406} & {0.416} & {0.408} & {0.421} & {0.414} & {0.429}& {0.419} & {0.431} & {0.355} & \textbf{0.403} & {0.392} & \underline{0.415}& {{0.428}} & {{0.432}} &0.426 &0.433 & {{0.415}} & {{0.426}} & {1.043} & {0.731} & {0.643} & {0.571}  & {{0.452}} & {{0.452}} &{0.594} &{0.541} &{0.496} &{0.487} &{{0.482}} &{0.486}\\ 
    & {720} & \underline{{0.412}} & \textbf{{0.429}}  & {0.414} & \underline{0.433}& {0.416} & {0.435} & {0.423} & {0.442} & {0.427} & {0.445} & {0.395} & {0.435} & \textbf{0.410} & \underline{0.433}& {{0.427}} & {{0.445}} &0.431 &0.446 & {{0.420}} & {{0.440}} & {1.104} & {0.763} & {0.874} & {0.679} & {{0.462}} & {{0.468}} &{0.831} &{0.657} &{{0.463}} &{{0.474}} &{0.515} &{0.511} \\ 
    \cmidrule(lr){2-34}
    & {Avg} & \underline{{0.358}} & \underline{{0.386}}& {0.364} & {0.388} & {0.367} & {0.393} & {0.372} & {0.400} & {0.377} & {0.403}  & \textbf{0.341} & \textbf{0.391}  & {0.380} & {0.403}  & {{0.383}} & {{0.407}} &0.387 &0.407 & {{0.374}} & {{0.398}} & {0.942} & {0.684} & {0.611} & {0.550}  &{{0.414}} &{{0.427}} &{0.559} &{0.515} &{{0.437}} &{{0.449}} &{0.450} &{0.459} \\  
    \midrule

    \multirow{5}{*}{\rotatebox{90}{{Weather}}} 
    &  {96} & \textbf{{0.147}} & \textbf{{0.200}}& {0.159} & {0.203} & {0.164} & {0.209} & {0.164} & {0.209} & {0.166} & {0.207} & {0.171} & {0.228} & \underline{0.154} & \underline{0.202}& {0.174} & {0.214} &0.177 &0.218 &0.192 &0.232 & {{0.158}} & {0.230}  & {0.202} & {0.261} &{{0.172}} &{{0.220}} & {0.196} &{0.255} & {0.217} &{0.296} & {0.266} &{0.336} \\ 
    & {192} & \textbf{{0.195}} & \textbf{{0.243}} & {0.211} & {0.250} & {0.212} & {0.252} & {0.212} & {0.252}& {0.216} & {0.254} & {0.218} & {0.269} & \underline{0.205} & \underline{0.249}& {0.221} & {{0.254}} &0.225 &0.259 &0.240 &0.271 & {{0.206}} & {0.277} & {0.242} & {0.298} &{{0.219}} &{{0.261}}  & {0.237} &{0.296} & {0.276} &{0.336} & {0.307} &{0.367} \\
    & {336} & \underline{{0.269}} & \textbf{{0.285}}& {0.270} & \underline{0.289} & {0.270} & {0.295} & {0.270} & {0.295} & {0.273} & {0.296} & \textbf{0.267} & {0.304} & {0.262} & \underline{0.289}& {0.278} & {{0.296}} &0.278 &0.297 &0.292 &0.307 & {{0.272}} & {0.335} & {0.287} & {0.335} &{{0.280}} &{{0.306}} & {0.283} &{0.335} & {0.339} &{0.380} & {0.359} &{0.395}\\ 
    & {720} & {{0.345}} & \textbf{{0.340}}& {0.345} & \textbf{0.340} & {0.348} & {0.345} & {0.348} & {0.345} & {0.351} & {0.346} & \textbf{0.336} & {0.350} & \underline{0.344} & \underline{0.342}& {0.358} & {0.349} &{0.354} &{0.348} & {0.364} & {0.353} & {0.398} & {0.418} & {{0.351}} & {0.386} &{0.365} &{{0.359}} & {{0.345}} &{{0.381}} & {0.403} &{0.428} & {0.419} &{0.428} \\ 
    \cmidrule(lr){2-34}
    & {Avg} & \textbf{{0.239}} & \textbf{{0.267}} & {0.246} & \underline{0.271} & {0.248} & {0.275} & {0.248} & {0.275} & {0.251} & {0.276}  & {0.248} & {0.287}  & \underline{0.241} & \underline{0.271}  & {{0.258}} & {{0.279}} &0.259 &0.281 &0.272 &0.291 & {0.259} & {0.315} & {0.271} & {0.320} &{{0.259}} &{{0.287}} &{0.265} &{0.317} &{0.309} &{0.360} &{0.338} &{0.382} \\  
    \midrule
      \multirow{5}{*}{\rotatebox{90}{{ECL}}} 
    &  {96} & \underline{{0.135}} & \textbf{{0.225}}& {0.152} & {0.239} & {0.137} & {0.235} & \textbf{0.134} & \underline{0.229} & {0.139} & {0.234} & {0.168} & {0.271} & {0.149} & {0.236}& \underline{{0.148}} & \underline{{0.240}}& {0.181} & {{0.270}} & {0.201} & {0.281}  & {0.219} & {0.314} & {0.237} & {0.329} &{{0.168}} &{0.272} &{0.197} &{0.282} &{0.193} &{0.308} &{0.201} &{0.317}  \\ 
    & {192} & \underline{{0.153}} & \textbf{{0.245}}& {0.167} & {0.251} & {0.159} & {0.257} & \textbf{0.152} & \textbf{0.245} & {0.156} & \underline{0.249} & {0.183} & {0.283} & {0.162} & {0.253}& {{0.162}} & {{0.253}} & {0.188} & {{0.274}}& {0.201} & {0.283}  & {0.231} & {0.322} & {0.236} & {0.330} &{{0.184}} &{0.289} &{0.196} &{{0.285}} &{0.201} &{0.315} &{0.222} &{0.334} \\ 
    & {336} & \textbf{{0.171}} & \textbf{{0.262}}& {0.182} & {0.270} & {0.178} & {0.272} & \textbf{0.171} & \underline{0.264} & {0.172} & {0.267} & {0.198} & {0.297} & {0.200} & {0.310} & {{0.178}} & {{0.269}}& {0.204} & {{0.293}}  & {0.215} & {0.298} & {0.246} & {0.337} & {0.249} & {0.344} &{{0.198}} &{{0.300}} &{0.209} &{{0.301}} &{0.214} &{0.329}&{0.231} &{0.338}  \\ 
    & {720} & \textbf{{0.204}} & \underline{{0.300}}& {0.224} & {0.308} & \underline{0.208} & \underline{0.300} & {0.212} & {0.302} & {0.209} & \textbf{0.299} & {0.238} & {0.329} & {0.220} & {0.320} & {{0.225}} & {{0.317}} & {0.246} & {0.324} & {0.257} & {0.331} & {0.280} & {0.363} & {0.284} & {0.373} &{{0.220}} &{{0.320}} &{0.245} &{0.333} &{0.246} &{0.355} &{0.254} &{0.361} \\
    \cmidrule(lr){2-34}
    & {Avg}  & \textbf{{0.165}} & \textbf{{0.258}} & {0.181} & {0.267} & {0.172} & {0.267} & \underline{0.166} & \underline{0.260} & {0.169} & {0.262}  & {0.196} & {0.295}  & {0.183} & {0.280}  & {{0.178}} & {{0.270}} & {0.205} & {{0.290}} & {0.219} & {0.298} & {0.244} & {0.334} & {0.251} & {0.344} &{{0.192}} &{0.295} &{0.212} &{0.300} &{0.214} &{0.327} &{0.227} &{0.338} \\  
    \midrule

    \multirow{5}{*}{\rotatebox{90}{{Exchange}}}
    &  {96} & \textbf{{0.083}} & \underline{{0.201}} & \underline{0.083}& \textbf{0.200} & {-} & {-} & {-} & {-} & \underline{0.085} & {0.204} & {-} & {-} & \textbf{0.083} & {0.205}& {{0.086}} &{0.206} &0.088 &0.205  & {0.093} & {0.217} & {0.256} & {0.367} & {0.094} & {0.218} & {0.107} & {0.234} & {0.088} & {0.218} & {0.148} & {0.278} & {0.197} & {0.323} \\ 
    &  {192} & \textbf{{0.169}} & \underline{{0.295}}& \underline{0.172} & \textbf{0.293} & {-} & {-} & {-} & {-} & {0.178} & {0.299} & {-} & {-} & {0.174} & {0.297}& {0.177} & {0.299} &{0.176} &{0.299} & {0.184} & {0.307} & {0.470} & {0.509} & {0.184} & {0.307} & {0.226} & {0.344} & {{0.176}} & {0.315} & {0.271} & {0.315} & {0.300} & {0.369} \\  
    &  {336} & \textbf{0.292} & \textbf{{0.395}}& {0.322} & {0.410} & {-} & {-} & {-} & {-} & {0.328} & {0.414} & {-} & {-} & {0.310} & {0.398}& {0.331} & {0.417} &\underline{0.301} &\underline{0.397} & {0.351} & {0.432} & {1.268} & {0.883} & {0.349} & {0.431} & {0.367} & {0.448} & {0.313} & {0.427} & {0.460} & {0.427} & {0.509} & {0.524} \\ 
    &  {720} & \underline{{0.832}} & \underline{{0.688}}& {0.855} & {0.693} & {-} & {-} & {-} & {-} & \textbf{0.817} & \textbf{0.679} & {-} & {-} & {1.011} & {0.756}& {0.847} & {{0.691}} &0.901 &0.714 &0.886 &0.714 & {1.767} & {1.068} & {0.852} & {0.698} & {0.964} & {0.746} & {{0.839}} & {0.695} & {1.195} & {{0.695}}& {1.447} & {0.941} \\ 
    \cmidrule(lr){2-34}
    &  {Avg} &  \textbf{{0.344}} & \textbf{{0.395}}& {0.358} & \underline{0.399} & {-} & {-} & {-} & {-} & \underline{0.352} & \underline{0.399} & - & -  & {0.395} & {0.414}  & {0.360} & {{0.403}} &0.367 &0.404 & {0.378} & {0.417} & {0.940} & {0.707} & {0.370} & {0.413} & {0.416} & {0.443} & {{0.354}} & {0.414} & {0.519} & {0.429} & {0.613} & {0.539} \\  
    \midrule
    
    \multirow{5}{*}{\rotatebox{90}{{Traffic}}} 
    & {96} & \textbf{{0.378}} & \textbf{{0.261}} & {0.441} & {0.264} & \underline{0.380} & \underline{0.262} & {0.440} & {0.263} & {-} & {-}& {0.441} & {0.333}& {-} & {-} & {{0.395}} & {{0.268}} & {{0.462}} & {0.295} & {0.649} & {0.389} & {0.522} & {{0.290}} & {0.805} & {0.493} &{{0.593}} &{{0.321}} &{0.650} &{0.396} &{{0.587}} &{0.366}  &{0.613} &{0.388} \\
    & {192} & \textbf{{0.401}} & \textbf{{0.269}}& {0.457} & {0.276} & \underline{0.407} & {0.275} & {0.454} & \underline{0.274} & {-} & {-} & {0.456} & {0.339} & {-} & {-} & {{0.417}} & {{0.276}} & {{0.466}} & {0.296} & {0.601} & {0.366} & {0.530} & {{0.293}} & {0.756} & {0.474} &{0.617} &{{0.336}} &{{0.598}} &{0.370} &{0.604} &{0.373} &{0.616} &{0.382}  \\ 
    & {336} & \textbf{{0.424}} & \textbf{{0.276}}& {0.465} & \underline{0.282} & \underline{0.429} & {0.284} & {0.472} & \underline{0.282} & {-} & {-} & {0.467} & {0.344} & {-} & {-}  & {{0.433}} & {{0.283}}& {{0.482}} & {{0.304}} & {0.609} & {0.369}  & {0.558} & {0.305}  & {0.762} & {0.477} &{0.629} &{{0.336}}  &{{0.605}} &{0.373} &{0.621} &{0.383} &{0.622} &{0.337} \\ 
    & {720} & \textbf{{0.441}} & \textbf{{0.292}} & {0.507} & {0.301} & \underline{0.452} & \underline{0.297} & {0.514} & {0.301}& {-} & {-} & {0.503} & {0.360} & {-} & {-}  & {{0.467}} & {{0.302}} & {{0.514}} & {{0.322}} & {0.647} & {0.387} & {0.589} & {0.328}  & {0.719} & {0.449} &{0.640} &{{0.350}} &{0.645} &{0.394} &{{0.626}} &{0.382}  &{0.660} &{0.408} \\ 
    \cmidrule(lr){2-34}
    & {Avg} & \textbf{{0.411}} & \textbf{{0.274}} & {0.468} & {0.281} & \underline{0.417} & \underline{0.279} & {0.470} & {0.280} & {-} & {-}  & {0.466} & {0.344}  & - & -  & {{0.428}} & {{0.282}} & {{0.481}} & {{0.304}} & {0.626} & {0.378}& {0.550} & {{0.304}} & {0.760} & {0.473} &{{0.620}} &{{0.336}} &{0.625} &{0.383} &{{0.610}} &{0.376} &{0.628} &{0.379} \\

    \bottomrule
  \end{tabular}}
  \label{tab:main_result}
\end{table*}

\subsection{Additional Ablation Studies}

\paragraph{Validation of Linear vs. Nonlinear PC}\label{nonlinearPC}
A potential limitation of the standard PC algorithm is its reliance on partial correlation tests, which assume linear dependencies. To investigate whether this restricts CDT's ability to capture complex dynamics, we implemented a variant using the PC algorithm with non-linear independence tests (kernel-based CI tests). Table~\ref{tab:nonlinearPC_comparison} compares the performance of CDT initialized with the standard Linear PC versus the Nonlinear PC across six datasets. We observe that the performance gap is negligible (within 0.01 on average). This result strongly supports the efficacy of our Dynamic Causal Adapter (DCA): although Linear PC may initially miss non-linear causal edges (False Negatives), the learnable nature of the DCA allows the model to recover these missing links end-to-end during training. Thus, CDT effectively combines the efficiency of linear causal discovery with the representational power of deep learning, making it robust to the choice of the initialization algorithm.

\begin{table}[h]
\centering
\caption{Comparison between CDT initialized with Linear PC (Ours) and Nonlinear PC. The results (Avg MSE on horizon 96/192/336/720) show that CDT is robust to the initialization choice, as the learnable adapter refines the structure during training.}
\label{tab:nonlinearPC_comparison}
\begin{tabular}{l|c|cccccc}
\toprule
Dataset & Horizon & MSE & MSE \\
&  & \textbf{Linear PC (Ours)} & Nonlinear PC &  \\
\midrule
ETTm1 & Avg & 0.365 & 0.369 & +0.004 \\
ETTm2 & Avg & 0.268 & 0.266 & -0.002 \\
ETTh1 & Avg & 0.406 & 0.410 & +0.004 \\
ETTh2 & Avg & 0.358 & 0.359 & +0.001 \\
Exchange & Avg & 0.344 & 0.339 & -0.005 \\
Weather & Avg & 0.239 & 0.240 & +0.001 \\
ECL & Avg & 0.165 & 0.169 & +0.004 \\
Traffic & Avg & 0.411 & 0.414 & +0.003 \\
\bottomrule
\end{tabular}
\end{table}

\paragraph{Causal sufficiency stability}  
The PC algorithm assumes causal sufficiency, i.e., all relevant confounders are observed. When relying on finite-sample conditional independence tests, the resulting graph may vary across data subsets, especially when latent confounding or sampling noise is present. Since CDT treats the PC output as a structural prior (rather than ground truth) and refines it end-to-end, our main conclusions do not hinge on this assumption. Here we provide a {stress test} that examines whether the estimated structures remain broadly stable across splits on a representative dataset. Concretely, on ETTh1, we run PC separately on the training split, the test split, and the full dataset, and compute the Jaccard similarity between the resulting (undirected) edge sets.We compute similarity on edge sets to avoid over-interpreting directions that are known to be less stable under finite-sample tests. We obtain 80.9\% similarity between train vs.\ test and 90.4\% between train vs.\ full, suggesting that the extracted structures are reasonably consistent under different splits in this setting. For transparency, we report the corresponding adjacency matrices in Table~\ref{tab:sufficiency}.

\begin{table}[h]
\centering
\caption{Adjacency matrices extracted by PC from training, test, and full ETTh1 datasets. 
Here $-1$ denotes a directed edge from the row variable to the column variable, 
$0$ indicates no edge, and $1$ denotes the opposite direction.}
\label{tab:sufficiency}
\vspace{0.1in}
\resizebox{0.7\linewidth}{!}{
\begin{tabular}{c|ccccccc}
\toprule
Train/Test/All & 1 & 2 & 3 & 4 & 5 & 6 & 7 \\
\midrule
1 & 0/0/0 & -1/0/1 & 0/-1/1 & -1/-1/-1 & 1/-1/-1 & 1/-1/0 & 0/0/0 \\
2 & 1/0/-1 & 0/0/0 & 0/-1/0 & -1/-1/1 & 1/1/1   & 1/0/1  & 1/1/1 \\
3 & 0/1/-1 & 0/1/0 & 0/0/0  & -1/-1/1 & 1/-1/1  & 1/-1/1 & -1/0/0 \\
4 & 1/1/1  & 1/1/-1& 1/1/-1 & 0/0/0   & 1/0/0   & 1/1/0  & 1/0/1  \\
5 & -1/1/1 & -1/-1/-1& -1/1/-1& -1/0/0& 0/0/0   & 0/-1/1 & -1/-1/-1 \\
6 & -1/1/0 & -1/0/-1& -1/1/-1& -1/-1/0& 0/-1/-1 & 0/0/0  & 0/-1/0  \\
7 & 0/0/0  & -1/-1/-1& 1/0/0 & -1/0/-1& 1/1/1   & 0/1/0  & 0/0/0   \\
\bottomrule
\end{tabular}}
\end{table}

\paragraph{Robustness to causal discovery algorithms.}  
To further assess whether CDT is overly sensitive to a specific causal discovery procedure, we compare several mainstream algorithms on ETTh1. We estimate graphs using PC, FCI, and PCMCI, and report their adjacency matrices (encoded by $-1/0/1$ for edge directions) in Table~\ref{tab:pc_fci_pcmci}. Based on the resulting edge sets, the Jaccard similarities are 95\% between PC and FCI, and 90\% between PC and PCMCI, indicating high overlap at the level of discovered connections.

We then initialize CDT with each estimated graph and run forecasting on ETTh1 under input length 96. As shown in Table~\ref{tab:etth1_causal_alg}, the MSE/MAE differences across PC-/FCI-/PCMCI-initialized variants are within 0.01 for all horizons (96/192/336/720). Overall, these results suggest that, for this dataset and setting, CDT's forecasting accuracy is not materially affected by switching among these commonly used causal discovery backends; This empirical finding aligns with our theoretical discussion in Appendix \ref{app:theoretical_alignment}, which argues that structural differences within the same Markov equivalence class do not alter the fundamental causal semantics required for our decomposition.

\begin{table}[h]
\centering
\caption{Adjacency matrices estimated by PC, FCI, and PCMCI on the ETTh1 dataset.
Here $-1$ denotes a directed edge from the row variable to the column variable,
$0$ indicates no edge, and $1$ denotes the opposite direction.}
\label{tab:pc_fci_pcmci}
\vspace{0.1in}
\resizebox{0.8\linewidth}{!}{
\begin{tabular}{c|ccccccc}
\toprule
PC/FCI/PCMCI & 1 & 2 & 3 & 4 & 5 & 6 & 7 \\
\midrule
1 & 0/0/0   & -1/1/1   & 0/-1/1   & -1/-1/-1 & 1/1/-1   & 1/-1/1   & 0/0/0   \\
2 & 1/-1/-1 & 0/0/0    & 0/-1/0   & -1/-1/1  & 1/1/1    & 1/0/1    & 1/1/-1  \\
3 & 0/1/-1  & 0/1/0    & 0/0/0    & -1/-1/1  & 1/1/1    & 1/1/1    & -1/-1/0 \\
4 & 1/1/1   & 1/1/-1   & 1/1/-1   & 0/0/0    & 1/0/0    & 1/1/0    & 1/0/1   \\
5 & -1/-1/1 & -1/-1/-1 & -1/-1/-1 & -1/0/0   & 0/0/0    & 0/-1/1   & -1/-1/-1\\
6 & -1/1/-1 & -1/0/-1  & -1/-1/-1 & -1/-1/0  & 0/-1/-1  & 0/0/0    & 0/0/-1  \\
7 & 0/0/0   & -1/-1/1  & 1/1/0    & -1/0/-1  & 1/1/1    & 0/0/1    & 0/0/0   \\
\bottomrule
\end{tabular}}
\end{table}

\begin{table}[h]
\centering
\caption{Forecasting performance of CDT on ETTh1 (input-96) using DAGs estimated by PC, FCI, and PCMCI. The differences across causal discovery algorithms are within 0.01 for all horizons.}
\label{tab:etth1_causal_alg}
\vspace{0.1in}
\begin{tabular}{c|cc|cc|cc}
\toprule
\multirow{2}{*}{Horizon} & \multicolumn{2}{c|}{PC-based} & \multicolumn{2}{c|}{FCI-based} & \multicolumn{2}{c}{PCMCI-based} \\
 & MSE & MAE & MSE & MAE & MSE & MAE \\
\midrule
96  & 0.352 & 0.379 & 0.354 & 0.379 & 0.352 & 0.380 \\
192 & 0.402 & 0.296 & 0.401 & 0.295 & 0.403 & 0.296 \\
336 & 0.434 & 0.430 & 0.434 & 0.431 & 0.433 & 0.429 \\
720 & 0.439 & 0.448 & 0.440 & 0.450 & 0.438 & 0.448 \\
\bottomrule
\end{tabular}
\end{table}

\paragraph{Ensemble and Bootstrapping of DAGs}\label{app:ensemble_bootstrap}
We additionally explore whether simple aggregation strategies can yield more reliable structural priors for CDT. On ETTh1, we consider: (i) an \emph{ensemble} prior obtained by running PC, FCI, and PCMCI on the full training set and applying majority voting over their edge sets; and (ii) a \emph{bootstrap} prior that assesses sampling variability by running PC on ten bootstrap resamples (each formed by sampling 10\% of the training set with replacement) and retaining edges that appear in at least six of the ten graphs.

We report the adjacency matrices of the single PC prior, the ensemble prior, and the bootstrap prior in Table~\ref{tab:ensemble_bootstrap_adj}, and evaluate CDT initialized with each prior in Table~\ref{tab:ensemble_bootstrap_perf}. Compared with the single PC prior, the ensemble prior yields a slight degradation, while the bootstrap prior leads to a more noticeable degradation across all horizons. A plausible explanation is that, under our bootstrap setting, each causal discovery run effectively operates on a much smaller sample, which can increase statistical noise in conditional-independence testing; voting/thresholding may further remove weak-but-useful edges and distort the prior sparsity pattern. These observations suggest that, in this setting, using a single graph estimated from the full training split provides a sufficiently informative prior for CDT, and naive aggregation does not necessarily improve downstream forecasting. This observation reinforces the motivation behind our Dynamic Causal Adapter: rather than relying on complex offline ensemble techniques to "fix" the graph, it is more effective to treat a single, noisy graph as a starting point and allow the downstream forecasting objective to drive the refinement end-to-end.

\begin{table}[h]
\centering
\caption{Adjacency matrices estimated by PC, ensemble voting, and bootstrap aggregation on ETTh1. Entries $-1/0/1$ denote a directed edge from row to column, no edge, and the opposite direction, respectively.}
\label{tab:ensemble_bootstrap_adj}
\resizebox{0.75\linewidth}{!}{
\begin{tabular}{c|ccccccc}
\toprule
PC / Ensemble / Bootstrap & 1 & 2 & 3 & 4 & 5 & 6 & 7 \\
\midrule
1 & 0/0/0   & -1/1/0   & 0/0/1   & -1/-1/-1 & 1/1/-1   & 1/1/-1   & 0/0/0   \\
2 & 1/-1/0  & 0/0/0    & 0/0/0   & -1/-1/1  & 1/1/0    & 1/1/0    & 1/1/-1  \\
3 & 0/0/-1  & 0/0/0    & 0/0/0   & -1/-1/1  & 1/1/1    & 1/1/0    & -1/-1/0 \\
4 & 1/1/1   & 1/1/-1   & 1/1/-1  & 0/0/0    & 1/0/0    & 1/1/0    & 1/1/1   \\
5 & -1/-1/1 & -1/-1/0  & -1/-1/-1& -1/0/0   & 0/0/0    & 0/0/1    & -1/-1/0 \\
6 & -1/-1/1 & -1/-1/0  & -1/-1/0 & -1/-1/0  & 0/-1/-1  & 0/0/0    & 0/0/-1  \\
7 & 0/0/0   & -1/-1/1  & 1/1/0   & -1/-1/-1 & 1/1/0    & 0/0/1    & 0/0/0   \\
\bottomrule
\end{tabular}}
\end{table}

\begin{table}[h]
\centering
\caption{Forecasting performance of CDT on ETTh1 when using PC-based, ensemble-based, or bootstrap-based DAGs (input length = 96). Ensemble and bootstrap aggregation do not improve performance and instead introduce noticeable degradation.}
\label{tab:ensemble_bootstrap_perf}
\vspace{0.1in}
\begin{tabular}{c|cc|cc|cc}
\toprule
\multirow{2}{*}{Horizon} &
\multicolumn{2}{c|}{PC-based} &
\multicolumn{2}{c|}{Ensemble-based} &
\multicolumn{2}{c}{Bootstrap-based} \\
& MSE & MAE & MSE & MAE & MSE & MAE \\
\midrule
96  & 0.352 & 0.379 & 0.362 & 0.385 & 0.382 & 0.401 \\
192 & 0.402 & 0.406 & 0.410 & 0.414 & 0.438 & 0.439 \\
336 & 0.434 & 0.430 & 0.450 & 0.442 & 0.480 & 0.473 \\
720 & 0.439 & 0.448 & 0.455 & 0.458 & 0.503 & 0.497 \\
\bottomrule
\end{tabular}
\end{table}

\subsection{Interpretability Experiments}\label{app:interpretability}
To further demonstrate that CDT uncovers meaningful causal structures and improves interpretability, we conduct two complementary experiments.

\paragraph{Ablation on All-to-one Forecasting Paradigm}  
To isolate the benefit of the proposed all-to-one SCM decomposition from the specific Transformer architecture, we replace the CDT backbone with a minimal MLP forecaster. We compare two training paradigms on the ETTh1 dataset (input-96, predict-96): (a) All-to-all MLP: Each target is predicted using the history of all $D$ variables; (b) All-to-one MLP (Ours): For each target, an independent MLP consumes only its ES, DCS, and CCS subsegments as defined by the PC-discovered DAG. As shown in Table~\ref{tab:mlp_causal}, the All-to-one MLP achieves equal or better performance across all targets (strictly better on 6/7 targets), reducing the average MSE from $0.395$ to $0.392$. This result suggests that the SCM-based decomposition strategy effectively filters out spurious interference and enhances predictive performance, proving its value even within a lightweight, architecture agnostic framework.

\begin{table}[h]
\centering
\caption{MLP on ETTh1 (input-96, predict-96): all-to-all vs. all-to-one (ES+DCS+CCS).}
\vspace{0.1in}
\label{tab:mlp_causal}
\begin{tabular}{lccccccc}
\toprule
Target & X1 & X2 & X3 & X4 & X5 & X6 & X7 \\
\midrule
All-to-all (All vars) & 0.392 & 0.403 & 0.382 & 0.386 & 0.388 & 0.391 & 0.420 \\
All-to-one (ES+DCS+CCS) & 0.390 & 0.403 & 0.380 & 0.383 & 0.385 & 0.386 & 0.419 \\
\bottomrule
\end{tabular}
\end{table}

\paragraph{Attribution-based Variable Removal}  
We then conduct a removal-and-retrain test by masking the variables in set (b) and set (c), respectively. The results are reported in Table~\ref{tab:interpretability}. When removing CDT-specific variables (Set b), CDT suffers a significant performance degradation (MSE rises from $0.105$ to $0.217$), whereas iTransformer is less affected ($0.132$ to $0.152$). This implies that the variables identified by CDT contain unique, non-redundant causal information essential for prediction. Conversely, removing Baseline-specific variables (Set c) has a negligible impact on CDT ($0.105$ to $0.114$), while iTransformer suffers a comparable drop. These findings suggest that CDT relies on robust causal precursors (which are hard to substitute), whereas baseline models like iTransformer often rely on spurious or redundant correlations that can be easily substituted by other variables. This confirms that CDT enhances interpretability by focusing on meaningful causal links. We use IG for fair comparison with attention-based baselines; additionally, CDT provides intrinsic relevance weights $\sigma(W)$, visualized in Fig.~\ref{fig:attn_heatmap}

\begin{table}[h]
\centering
\caption{MSE on Weather dataset under variable removal.}
\vspace{0.1in}
\label{tab:interpretability}
\begin{tabular}{lcc}
\toprule
Model & CDT & iTransformer \\
\midrule
w/ all & 0.105 & 0.132 \\
w/o set (b) & 0.217 & 0.152 \\
w/o set (c) & 0.114 & 0.147 \\
\bottomrule
\end{tabular}
\end{table}

We then remove the variables in set (b) or set (c) and retrain both models. Table~\ref{tab:interpretability} shows the results. Removing CDT-specific variables (\emph{set b}) causes a large drop in CDT’s performance (MSE $0.105 \to 0.217$) and a smaller decline for iTransformer. In contrast, removing iTransformer-specific variables (\emph{set c}) has little effect on CDT ($0.105 \to 0.114$), while iTransformer suffers a comparable drop to the removal of set (b). 

These findings suggest that CDT relies more on variables with genuine causal relevance, whereas iTransformer is more sensitive to variables that can be substituted by correlations. This demonstrates that CDT not only improves forecasting accuracy but also enhances interpretability by uncovering meaningful causal links.

\subsection{Robustness Evaluation}\label{robust}
We provide additional robustness diagnostics beyond the main perturbation study in Section~\ref{Experiment}.

\paragraph{Random seed sensitivity.}  
We repeat experiments five times with different random seeds $\{2021, 2022, 2023, 2024, 2025\}$ on the ETTm1, ETTh1, and Exchange-rate datasets. Table \ref{tab:std} reports the standard deviations of CDT’s performance, which are consistently low, demonstrating stable and reproducible results.

\vspace{-0.1in}
\begin{table*}[htbp]
\renewcommand{\arraystretch}{0.65} %行间距
  \caption{Robustness evaluation of CDT. The reported results (MSE and MAE) reflect the mean and standard deviation computed over five independent runs with different random seeds.}
  \vspace{0.1in}
  % \vspace{1.5pt}
  \label{tab:std}
  \centering
  % \begin{threeparttable}
  \begin{small}
  \renewcommand{\multirowsetup}{\centering}
  \setlength{\tabcolsep}{6pt}
\resizebox{0.8\linewidth}{!}{
  \begin{tabular}{c|cc|cc|cc}
    \toprule
    Dataset & \multicolumn{2}{c}{ETTm1} & \multicolumn{2}{c}{ETTh1} & \multicolumn{2}{c}{Exchange}   \\
    \cmidrule(lr){2-3} \cmidrule(lr){4-5}\cmidrule(lr){6-7}
    Horizon & MSE & MAE & MSE & MAE & MSE & MAE  \\
    \toprule
    $96$ & 0.307\scalebox{0.9}{$\pm$0.002} & 0.334\scalebox{0.9}{$\pm$0.001} & 0.352\scalebox{0.9}{$\pm$0.002} & 0.379\scalebox{0.9}{$\pm$0.003} & 0.083\scalebox{0.9}{$\pm$0.000} & 0.201\scalebox{0.9}{$\pm$0.002}  \\
    $192$ & 0.351\scalebox{0.9}{$\pm$0.002} & 0.357\scalebox{0.9}{$\pm$0.003} & 0.402\scalebox{0.9}{$\pm$0.003} & 0.406\scalebox{0.9}{$\pm$0.001} & 0.169\scalebox{0.9}{$\pm$0.001} & 0.295\scalebox{0.9}{$\pm$0.001}    \\
    $336$ & 0.374\scalebox{0.9}{$\pm$0.003} & 0.382\scalebox{0.9}{$\pm$0.001} & 0.434\scalebox{0.9}{$\pm$0.002} & 0.430\scalebox{0.9}{$\pm$0.002} & 0.292\scalebox{0.9}{$\pm$0.002} & 0.395\scalebox{0.9}{$\pm$0.001}    \\
    $720$ & 0.431\scalebox{0.9}{$\pm$0.002} & 0.427\scalebox{0.9}{$\pm$0.004} & 0.439\scalebox{0.9}{$\pm$0.003} & 0.448\scalebox{0.9}{$\pm$0.004} & 0.832\scalebox{0.9}{$\pm$0.005} & 0.688\scalebox{0.9}{$\pm$0.003}    \\
    \midrule
    Dataset & \multicolumn{2}{c}{Traffic} & \multicolumn{2}{c}{ECL} & \multicolumn{2}{c}{Weather}   \\
    \cmidrule(lr){2-3} \cmidrule(lr){4-5}\cmidrule(lr){6-7}
    Horizon & MSE & MAE & MSE & MAE & MSE & MAE  \\
    \toprule
    $96$ & 0.378\scalebox{0.9}{$\pm$0.003} & 0.261\scalebox{0.9}{$\pm$0.002} & 0.135\scalebox{0.9}{$\pm$0.001} & 0.225\scalebox{0.9}{$\pm$0.001} & 0.147\scalebox{0.9}{$\pm$0.000} & 0.200\scalebox{0.9}{$\pm$0.003}  \\
    $192$ & 0.401\scalebox{0.9}{$\pm$0.007} & 0.269\scalebox{0.9}{$\pm$0.004} & 0.153\scalebox{0.9}{$\pm$0.004} & 0.245\scalebox{0.9}{$\pm$0.002} & 0.195\scalebox{0.9}{$\pm$0.001} & 0.243\scalebox{0.9}{$\pm$0.002}    \\
    $336$ & 0.424\scalebox{0.9}{$\pm$0.002} & 0.276\scalebox{0.9}{$\pm$0.001} & 0.171\scalebox{0.9}{$\pm$0.002} & 0.262\scalebox{0.9}{$\pm$0.003} & 0.269\scalebox{0.9}{$\pm$0.002} & 0.285\scalebox{0.9}{$\pm$0.002}    \\
    $720$ & 0.441\scalebox{0.9}{$\pm$0.001} & 0.292\scalebox{0.9}{$\pm$0.002} & 0.204\scalebox{0.9}{$\pm$0.005} & 0.300\scalebox{0.9}{$\pm$0.004} & 0.345\scalebox{0.9}{$\pm$0.005} & 0.340\scalebox{0.9}{$\pm$0.004}    \\
    \bottomrule
  \end{tabular}
  }
  \end{small}
  % \end{threeparttable}
\end{table*}

\paragraph{Variable ordering sensitivity.}  
Variables in multivariate time series (e.g., temperature, pressure) have no inherent order, and altering their positions should not affect causal semantics. To verify this, we randomly shuffled all variables three times on the Weather dataset. The PC algorithm consistently produced the same causal graph, and retraining CDT (input-96, predict-96) resulted in negligible performance changes ($\Delta$MSE < 0.001, $\Delta$MAE < 0.001). This confirms that CDT is robust to variable ordering.

\subsection{Parameter sensitivity analysis}
\begin{figure}[h]
\centering
\subfloat[look-back window]{
    \includegraphics[width=0.24\linewidth]{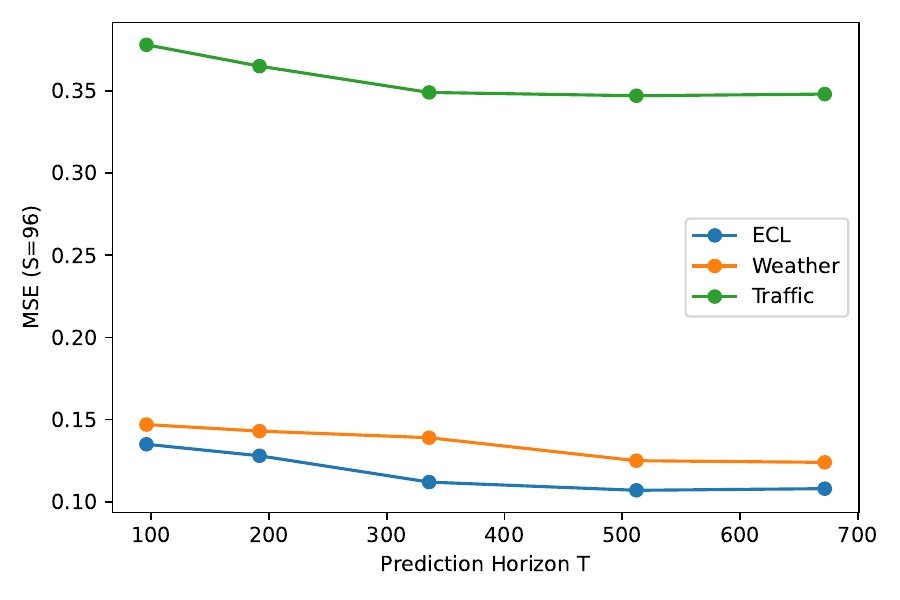}
}
\subfloat[$\lambda$ sensitivity]{
    \includegraphics[width=0.24\linewidth]{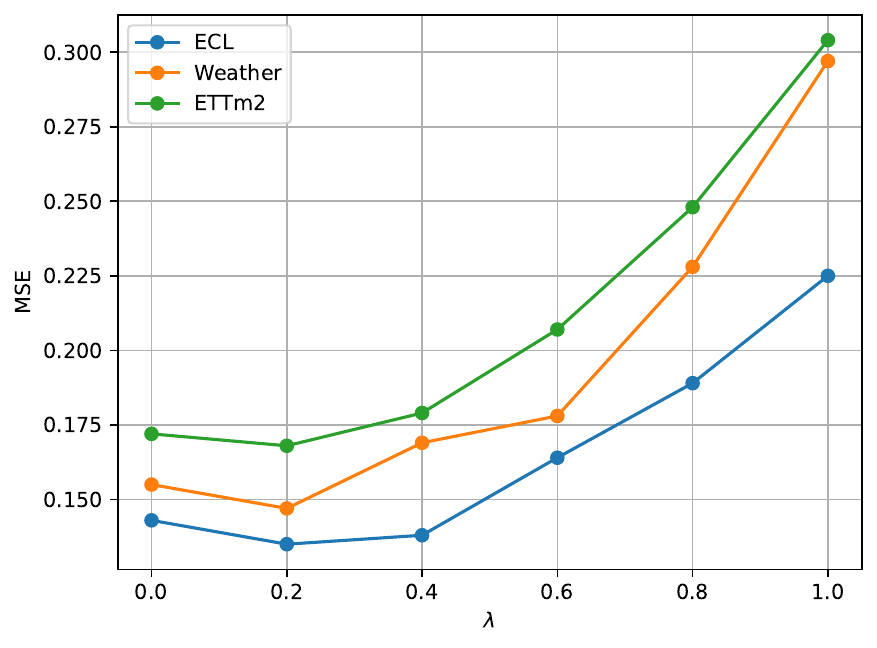}
}
\subfloat[$\alpha$ sensitivity]{
    \includegraphics[width=0.24\linewidth]{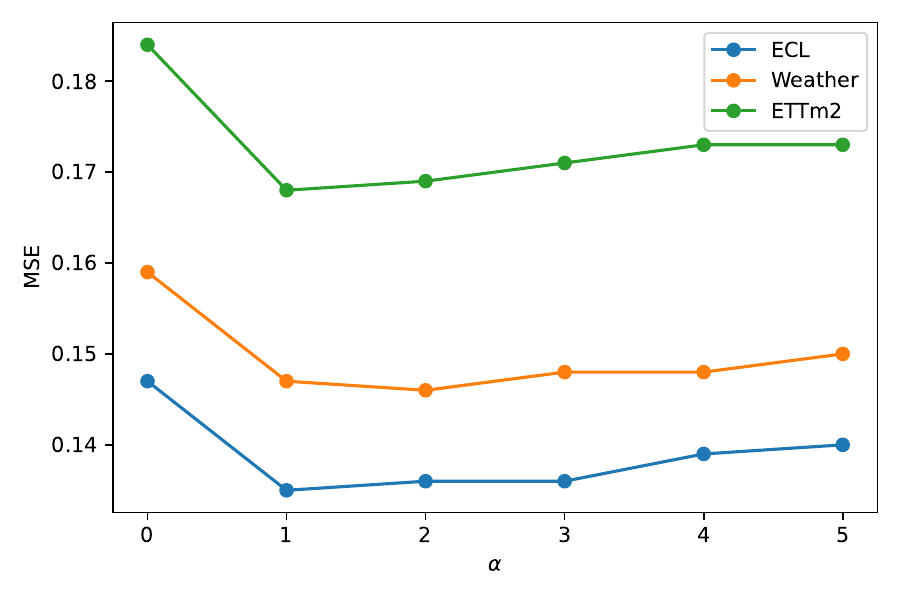}
}
\subfloat[$\beta$ sensitivity]{
    \includegraphics[width=0.24\linewidth]{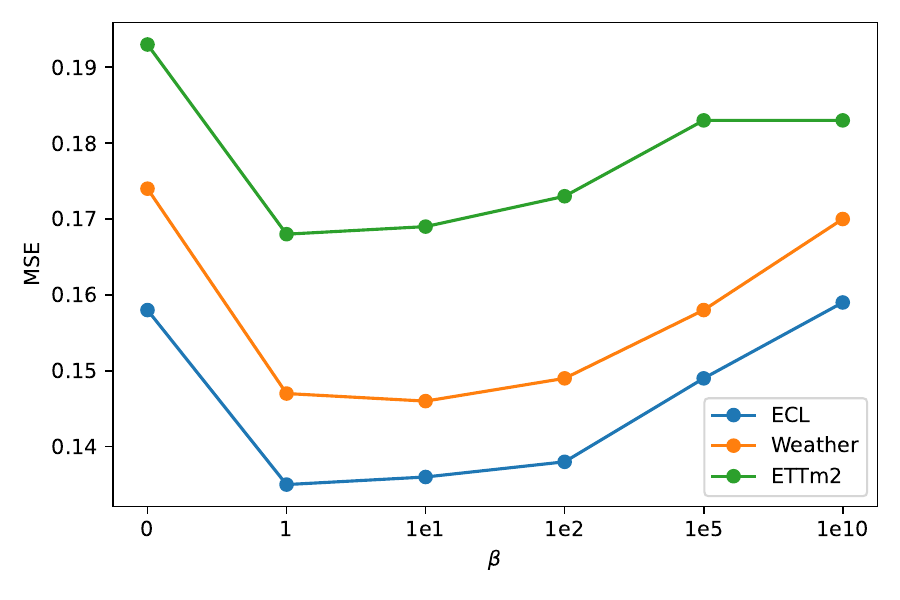}
}
\caption{Parameter sensitivity analysis.}
\label{fig:sensa}
\end{figure}

\paragraph{Effect of Look-back Window Length}\label{hisL}
Following prior observations that longer historical contexts can benefit Transformer-based forecasters~\citep{iTransformer, PatchTST}, we evaluate CDT under different input lengths while fixing the prediction horizon to $S=96$ on the ECL, Weather, and Traffic dataset. Specifically, we vary the look-back window $T\in\{96,192,336,512,672\}$. As shown in Fig~\ref{fig:sensa}[a]. We observe a clear and monotonic improvement as $T$ increases, suggesting that CDT can effectively exploit extended temporal contexts. The gain begins to saturate when $T\ge 512$, which is consistent with the diminishing returns commonly reported in long-context forecasting.

\paragraph{Regularization Coefficient $\lambda$}
The coefficient $\lambda$ controls the strength of the sparsity and prior-alignment regularization in the training objective. We examine the effect of $\lambda \in [0,1]$ on ETTm2, Weather, and ECL for input 96 output 96. As shown in Figure~\ref{fig:sensa}[b], we observe a "U-shaped" performance curve, with the optimal performance achieved around 0.2, striking a balance between sparsity and adaptability.

\paragraph{Regularization Coefficient $\alpha$ and $\beta$}
The initialization of the learnable logit matrices $W$ is defined as $W_{ij} = \alpha$ if $A_{\text{init}}[i,j] \neq 0$, and $W_{ij} = -\beta$ otherwise. This configuration represents our prior belief strength.  This parameter determines the initial activation strength of edges discovered by the PC algorithm. We examine the performance in the range of $[0,5]$. As shown in Figure~\ref{fig:sensa}[c] find that the performance is relatively robust to $\alpha$ in the range of $[1.0, 3.0]$. We attribute this robustness to the saturation property of the sigmoid function in the adapter, which ensures that once a causal link is initialized with sufficient strength, the exact magnitude becomes less critical. We fix $\alpha=1.0$ for the main experiments. Similarly, we examine the performance of $\beta$ in the range of $[0,1e10]$. As shown in Figure~\ref{fig:sensa}[d], we fix $\beta=1.0$ for the main experiments.

\subsection{Adapting CDT to Pre-trained LLMs}
While the main implementation of the Causal Decomposition Transformer (CDT) utilizes a Transformer encoder trained from scratch, the proposed structural decomposition and dynamic refinement mechanisms are fundamentally model-agnostic. To demonstrate the scalability and adaptability of our framework, we extend the architecture to incorporate pre-trained Large Language Models (LLMs) as the predictive backbone, denoted here as Causal-LLM.

In this variant, we freeze the parameters of a pre-trained LLM (e.g., LLaMA-7B) to leverage its robust reasoning capabilities and few-shot generalization potential. The core modification lies in the alignment of feature spaces. Specifically, the variable histories are first processed by the Dynamic Causal Adapter (DCA) described in Section~\ref{Sec:Method} to obtain the causal context representations $H_{\text{DCS}}^i$ and $H_{\text{CCS}}^i$. Instead of feeding these directly into a standard Transformer, we project them into the LLM's embedding dimension $d_{\text{model}}$ via a linear alignment layer. To enable the LLM to comprehend the causal structure without fine-tuning, we construct a structured input sequence for each target variable $V_i$ by concatenating a task instruction prefix, the aligned causal contexts, and the endogenous history.

Crucially, rather than relying on the LLM's internal attention to implicitly discern causal roles, we inherit the segment constrained causal mask mechanism from CDT. This mask is applied to the LLM's self-attention layers (via soft-prompting or direct attention bias, depending on the LLM implementation), strictly enforcing that the endogenous reasoning stream only aggregates information from the identified direct causes and collider structures. The final prediction is obtained by projecting the LLM's last hidden state back to the target dimension, followed by the spouse projection operator $\Psi$ to enforce conditional independence. This design effectively transforms the frozen LLM into a causal-aware forecaster, where the lightweight adapter refines the noisy structural priors and the LLM performs the complex temporal inference. The result as shown in Table \ref{tab:llm_ablation}

\begin{table}[htbp]
    \centering
    \caption{Performance of Causal-LLM with different frozen LLM backbones under a fixed context length $C = 672$. Consistent improvements across GPT-2 (124M) and LLaMA-7B demonstrate that the proposed residual feedback correction is model-agnostic and independent of the underlying LLM architecture.}
  \label{tab:llm_ablation}
  \footnotesize
  \begin{threeparttable}
  \begin{small}
  \renewcommand{\multirowsetup}{\centering}
  \setlength{\tabcolsep}{1pt}
  \resizebox{0.8\textwidth}{!}{
  \begin{tabular}{c|c|cccc|cccc|cccc|cccc}
    \toprule
    \multirow{2}{*}{LLM} & \multirow{2}{*}{Metric} & \multicolumn{4}{c|}{ETTh1} & \multicolumn{4}{c|}{ECL} & \multicolumn{4}{c}{Traffic} & \multicolumn{4}{c}{Weather} \\
    \cmidrule(lr){3-6} \cmidrule(lr){7-10} \cmidrule(lr){11-14} \cmidrule(lr){15-18} 
     && 96 & 192 & 336 & 720 & 96 & 192 & 336 & 720 & 96 & 192 & 336 & 720 & 96 & 192 & 336 & 720 \\
    \midrule
    GPT-2 \cite{gpt2} & MSE & 0.352 & 0.382 & 0.399 & 0.420 & 0.131 & 0.138 & 0.169 & 0.201 & 0.352 & 0.381 & 0.407 & 0.438 & 0.134 & 0.189 & 0.249 & 0.331 \\
    & MAE & 0.390 & 0.407 & 0.425 & 0.445 & 0.224 & 0.242 & 0.261 & 0.295 & 0.242 & 0.257 & 0.264 & 0.285 & 0.201 & 0.247 & 0.291 & 0.345 \\
    \midrule
    LLaMA \cite{Llama2} & MSE & 0.339 & 0.368 & 0.374 & 0.401 & 0.119 & 0.124 & 0.153 & 0.185 & 0.343 & 0.369 & 0.387 & 0.415 & 0.119 & 0.168 & 0.231 & 0.312 \\
    & MAE & 0.372 & 0.375 & 0.412 & 0.431 & 0.208 & 0.223 & 0.240 & 0.281 & 0.229 & 0.236 & 0.247 & 0.269 & 0.192 & 0.221 & 0.273 & 0.324 \\
    \bottomrule
  \end{tabular}}
  \end{small}
  \end{threeparttable}
\end{table}

\section{Causal Discovery Algorithm Analysis}\label{pc}
For the graphical causal semantics (Markov compatibility, d-separation, Markov equivalence, and faithfulness) that justify PC as an observationally identifiable structural prior, see Appendix~\ref{app:pc_semantics}. Here we focus on the concrete implementation and runtime characteristics.

\subsection{Causal Discovery Visualization}\label{visual}
\begin{figure*}[htbp]
\centering
\subfloat[Weather]{
    \includegraphics[width=0.55\linewidth]{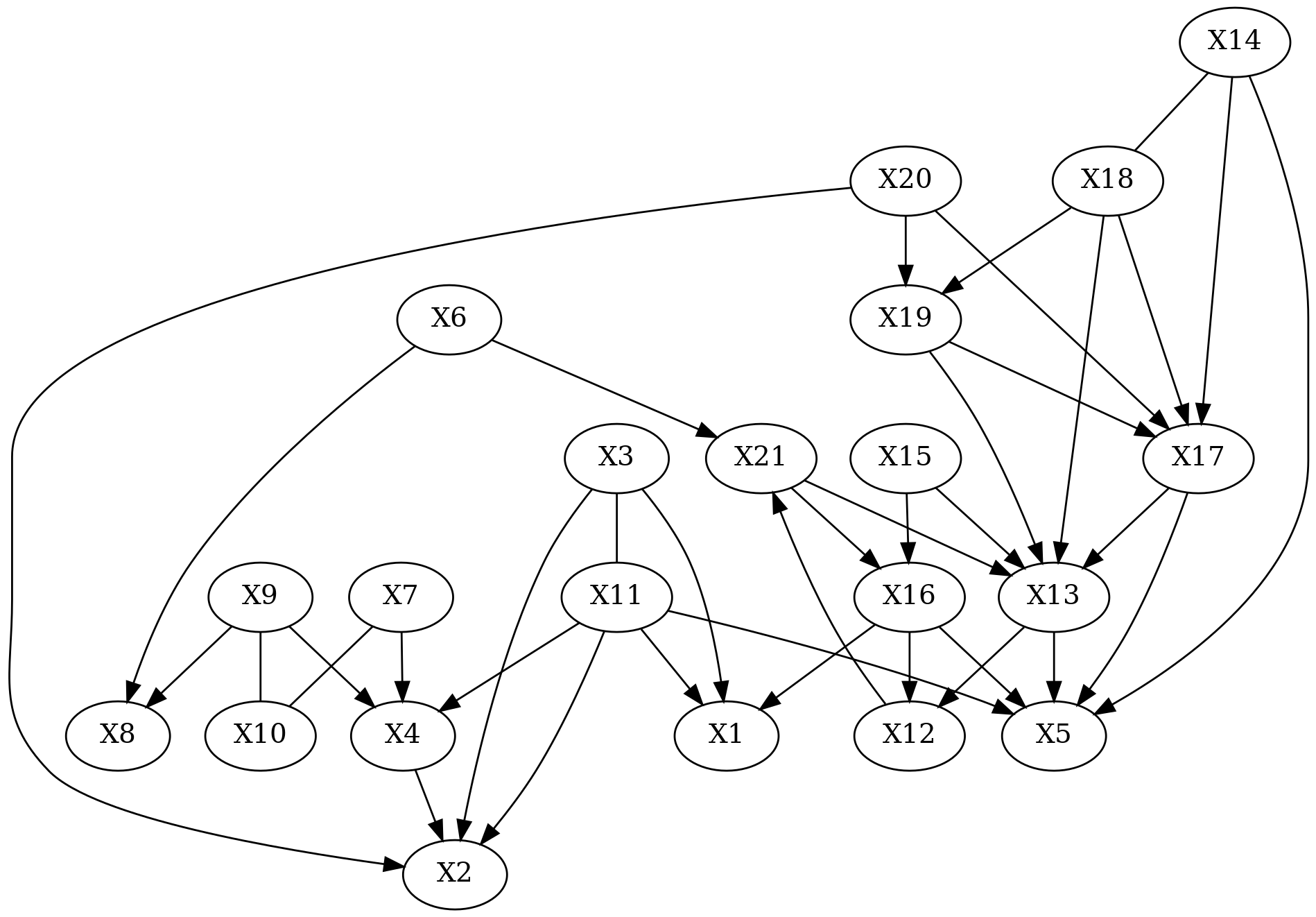}
    \label{fig:weather}
}
\subfloat[Exchange]{
    \includegraphics[width=0.25\linewidth]{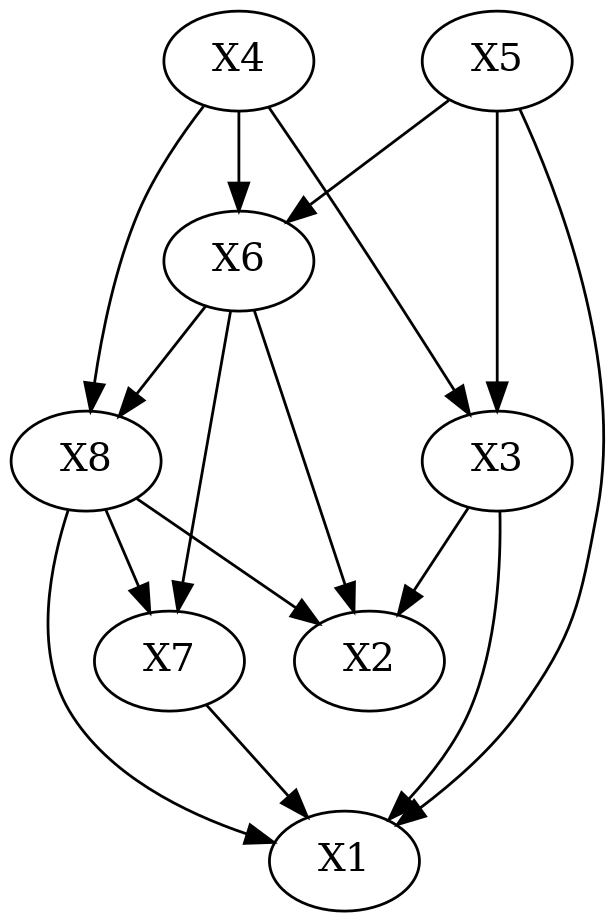}
    \label{fig:exchange}
}
\\
\subfloat[ETTh1]{
    \includegraphics[width=0.17\linewidth]{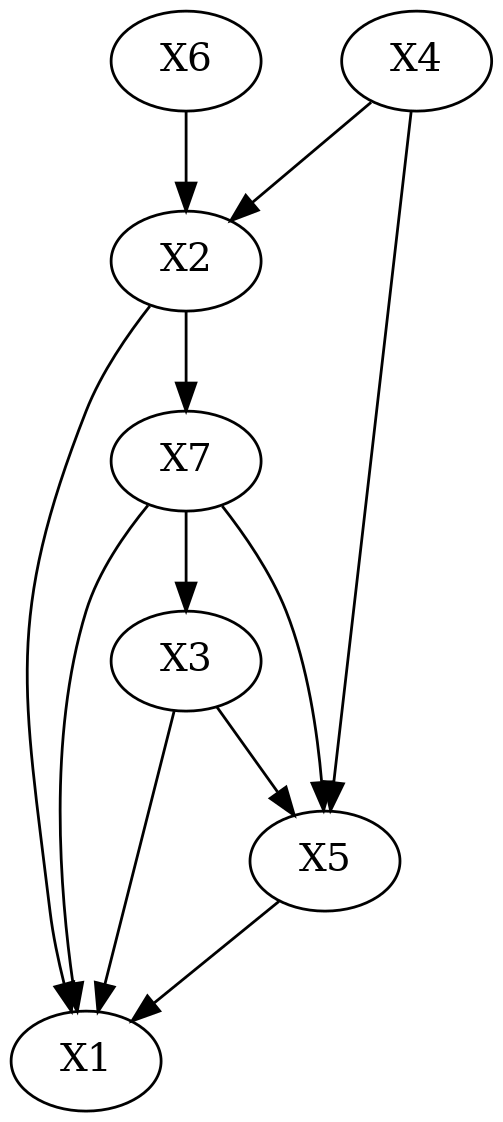}
    \label{fig:etth1}
}
\subfloat[ETTh2]{
    \includegraphics[width=0.25\linewidth]{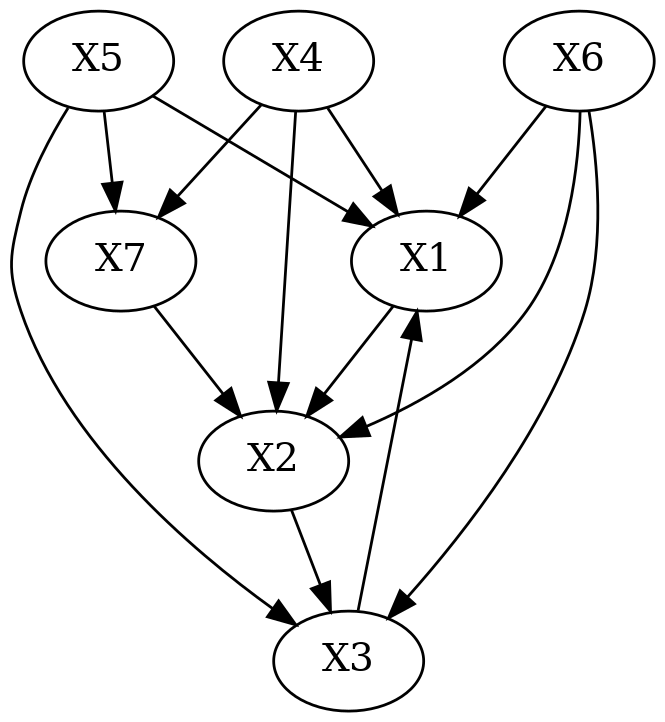}
    \label{fig:etth2}
}
\subfloat[ETTm1]{
    \includegraphics[width=0.17\linewidth]{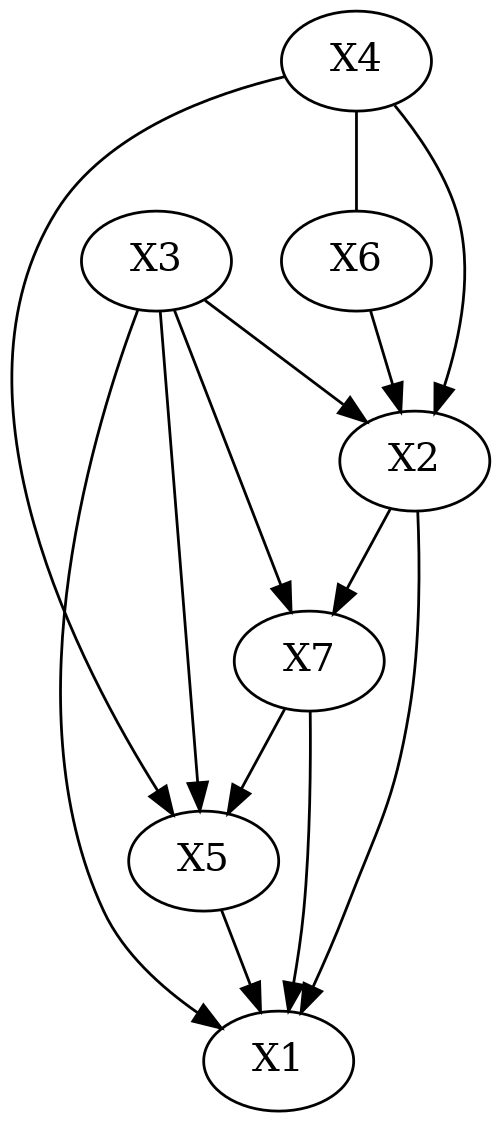}
    \label{fig:ettm1}
}
\subfloat[ETTm2]{
    \includegraphics[width=0.32\linewidth]{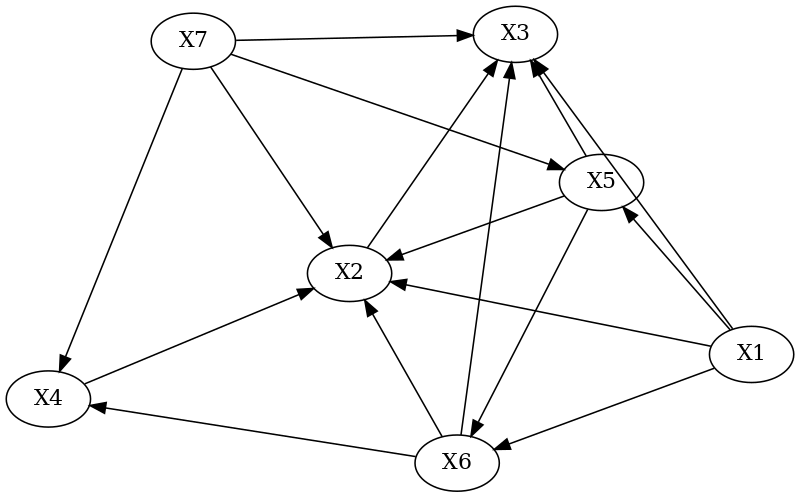}
    \label{fig:ettm2}
}
\caption{Visualization of causal DAGs discovered by the PC algorithm across different datasets. Directed edges indicate inferred causal relationships between variables, while undirected edges indicate uncertainty regarding causal direction. The results cover six datasets: (a) Weather,(b) Exchange, (c) ETTh1, (d) ETTh2, (e) ETTm1, and (f) ETTm2.}
\label{fig:dag}
\end{figure*}

Figure \ref{fig:dag} displays the DAGs discovered by the PC algorithm on six datasets. Directed edges denote compelled causal relations, whereas undirected edges mark orientational ambiguity. We primarily use ETTh1, ETTh2, Weather, and Exchange datasets to illustrate the semantic alignment.

Beyond the statistical visualization in Figure~\ref{fig:dag}, the discovered structures also align with variable semantics. For example, in the ETT datasets we consistently observe $X_4 \to X_2 \leftarrow X_6$, where medium and low loads jointly influence high load[cite: 58, 67, 343]. This reflects the known interactions among load levels in power systems. In the Exchange dataset, a structure $X_4 \to X_3 \leftarrow X_5$ indicates that two USD–currency exchange rates jointly affect a third one, capturing triangular relations commonly seen in foreign exchange markets. These examples demonstrate that the extracted DAGs are consistent with practical domain knowledge.

\begin{figure*}[htpb]
\centering
    \subfloat[$D_\text{mask}$ on ETTh1]{
        \includegraphics[width=0.15\linewidth]{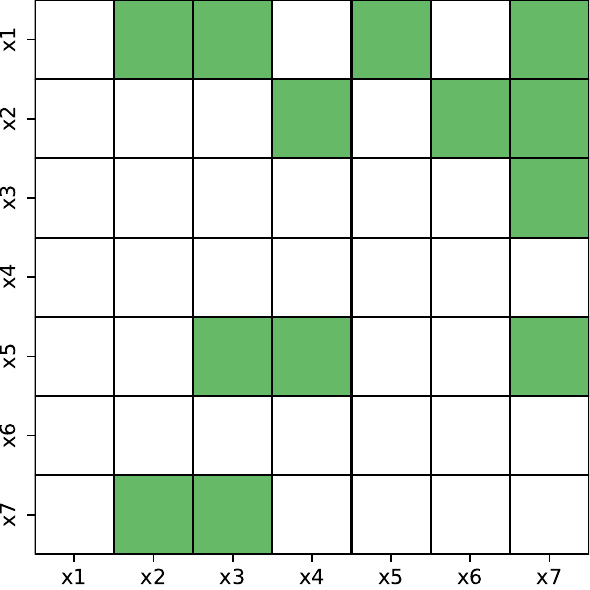}
    }
    % \hspace{-0.2in}
    \subfloat[$CS_\text{mask}$ on ETTh1]{
        \includegraphics[width=0.15\linewidth]{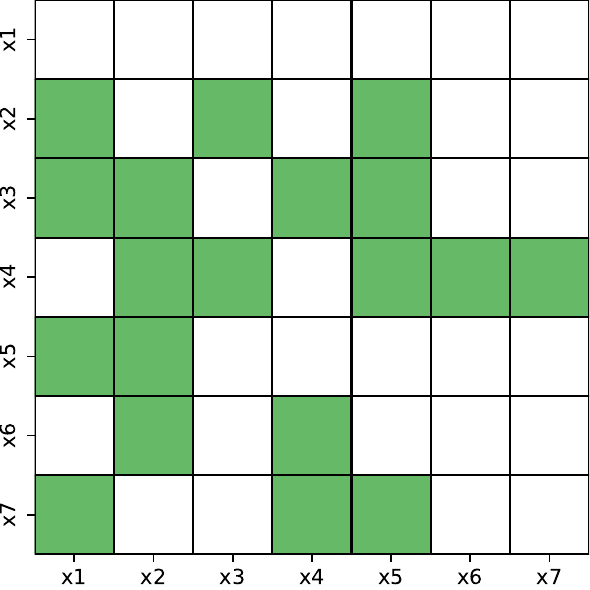}
    }
    % \hspace{-0.2in}
    \subfloat[$S_\text{mask}$ on ETTh1]{
        \includegraphics[width=0.15\linewidth]{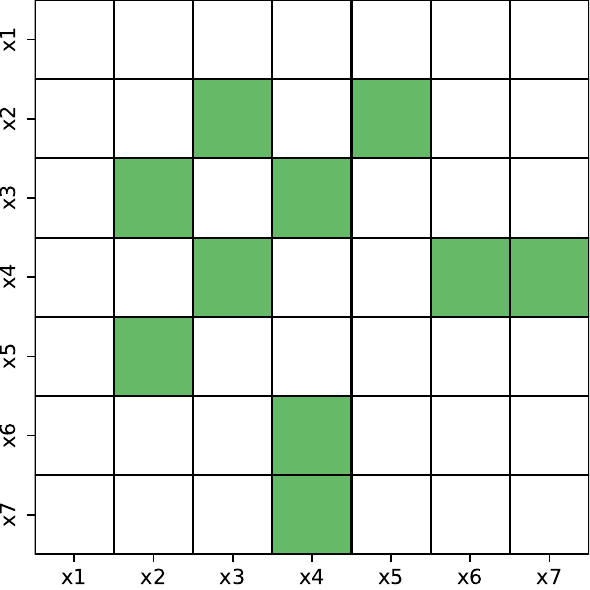}
    }
    \\
    \subfloat[$D_\text{mask}$ on ETTh2]{
        \includegraphics[width=0.15\linewidth]{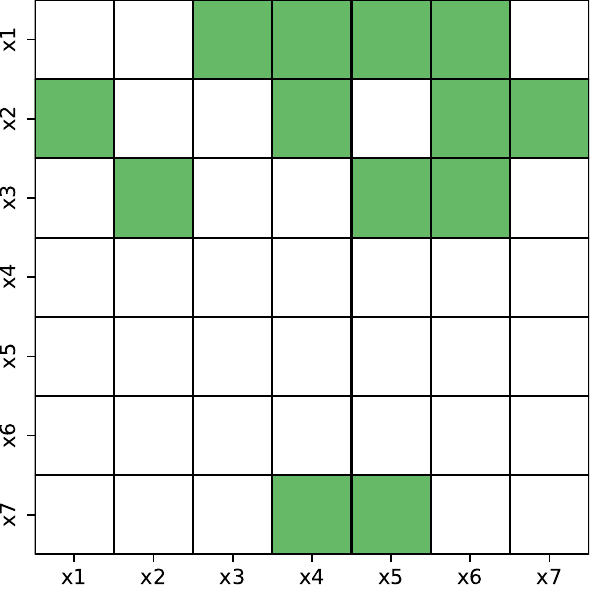}
    }
    % \hspace{-0.2in}
    \subfloat[$CS_\text{mask}$ on ETTh2]{
        \includegraphics[width=0.15\linewidth]{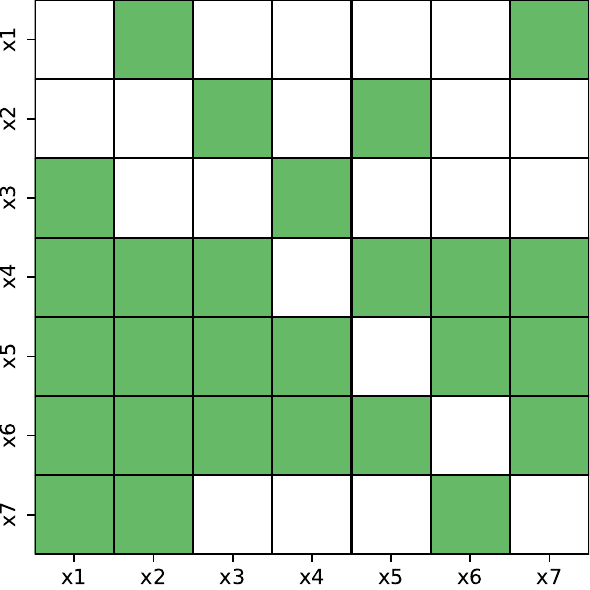}
    }
    % \hspace{-0.2in}
    \subfloat[$S_\text{mask}$ on ETTh2]{
        \includegraphics[width=0.15\linewidth]{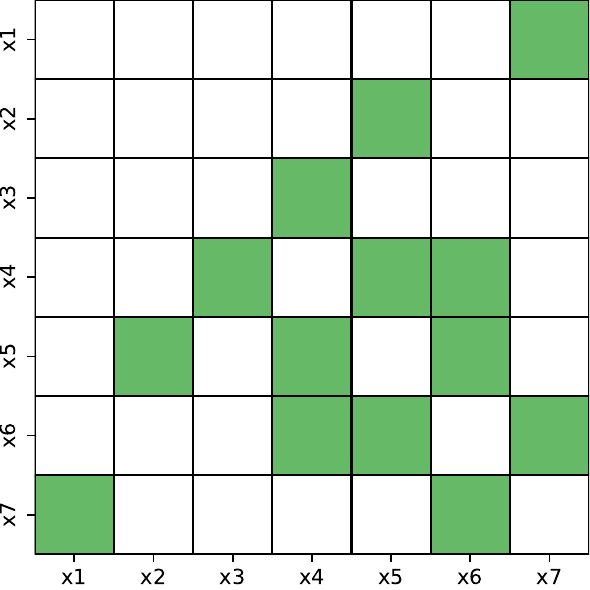}
    }
    \\
    \subfloat[$D_\text{mask}$ on Weather]{
        \includegraphics[width=0.2\linewidth]{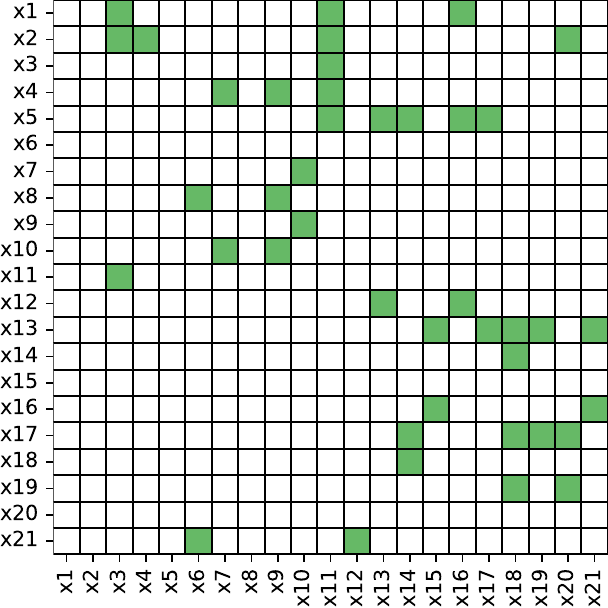}
    }
    % \hspace{-0.2in}
    \subfloat[$CS_\text{mask}$ on Weather]{
        \includegraphics[width=0.2\linewidth]{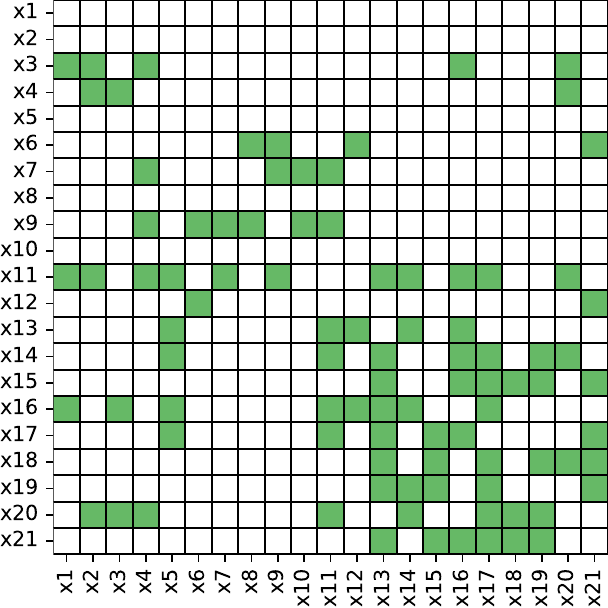}
    }
    % \hspace{-0.2in}
    \subfloat[$S_\text{mask}$ on Weather]{
        \includegraphics[width=0.2\linewidth]{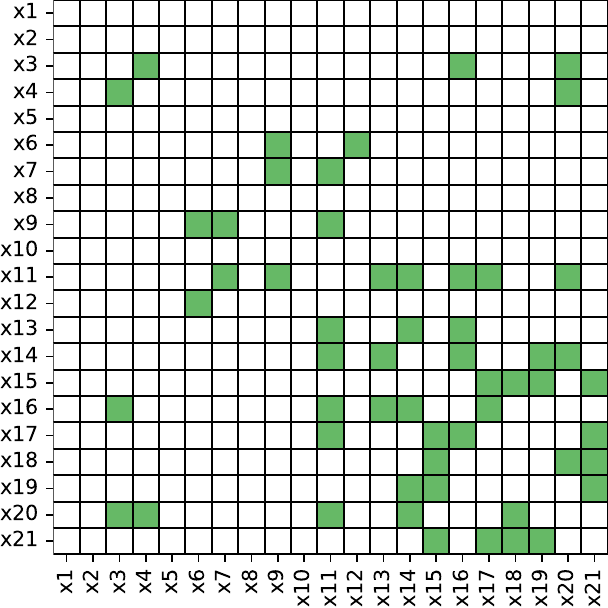}
    }
    \caption{Visualization of the {structural priors} (adjacency matrices) derived from the PC algorithm on ETTh1, ETTh2, and Weather. These binary maps serve as the \textbf{sparse initialization} for the learnable weights in the Dynamic Causal Adapter.}
    \label{fig:Visualization—mask}
\end{figure*}

Based on these causal graphs, we derive the {structural priors} used to initialize the Dynamic Causal Adapter in CDT (corresponding to the initialized states of $W_{\text{DCS}}$, $W_{\text{CCS}}$, and $W_{\text{SP}}$). Their visualizations in Figure~\ref{fig:Visualization—mask} show clear structural patterns: some variables lack direct causal parents, while others do not have indirect auxiliary variables. These priors provide a sparse initialization that guides the model optimization, explaining why disabling the causal initialization leads to performance degradation.

\begin{figure}[h]
    \centering
    \subfloat[DCS learned relevance weights]{
        \includegraphics[width=0.35\linewidth]{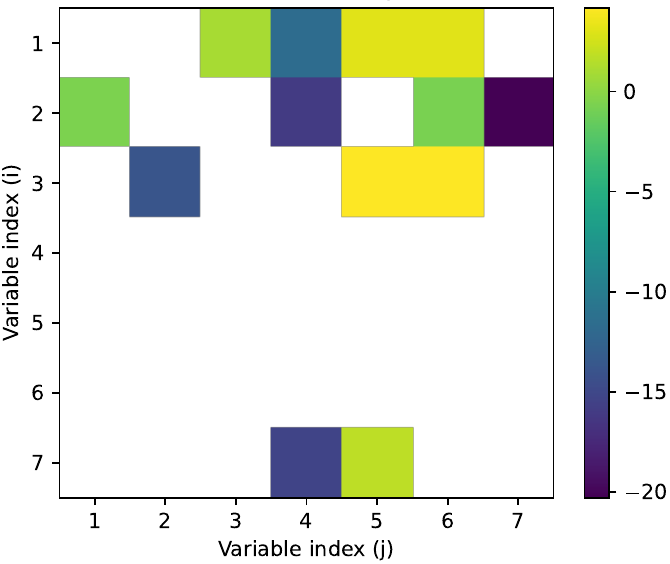}
        \label{fig:attn_dcs}
    }
    \subfloat[CCS learned relevance weights]{
        \includegraphics[width=0.35\linewidth]{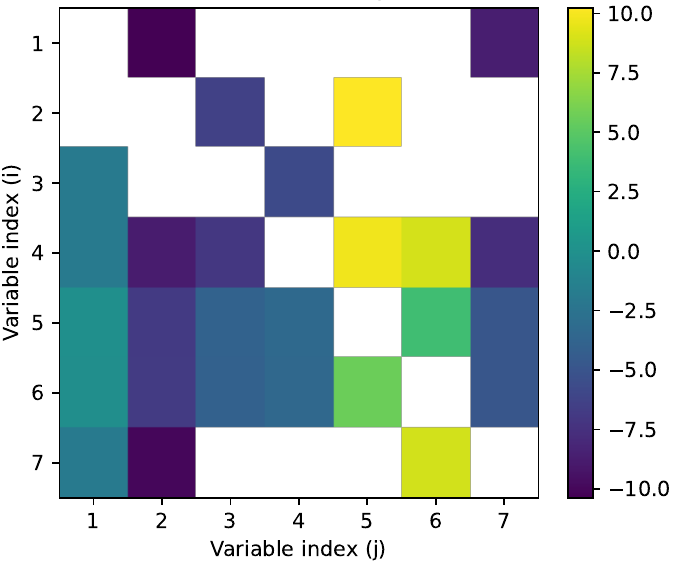}
        \label{fig:attn_ccs}
    }
    \caption{Visualization of the learned relevance weights for DCS and CCS segments on ETTh2. Warmer colors indicate that the Dynamic Causal Adapter assigns higher importance to these variables for generating causal representations.}
    \label{fig:attn_heatmap}
\end{figure}

To further demonstrate the effectiveness of the Dynamic Causal Adapter, we visualize the {learned relevance weights} (i.e., the sigmoid-activated logits $\sigma(W)$) in Figure~\ref{fig:attn_heatmap}. Darker or warmer colors highlight stronger causal relevance. These visualizations confirm that CDT successfully learns to assign higher weights to variables with plausible causal impact while suppressing irrelevant ones, effectively refining the initial PC priors during end-to-end training.

\begin{figure}
    \centering
    \includegraphics[width=0.5\linewidth]{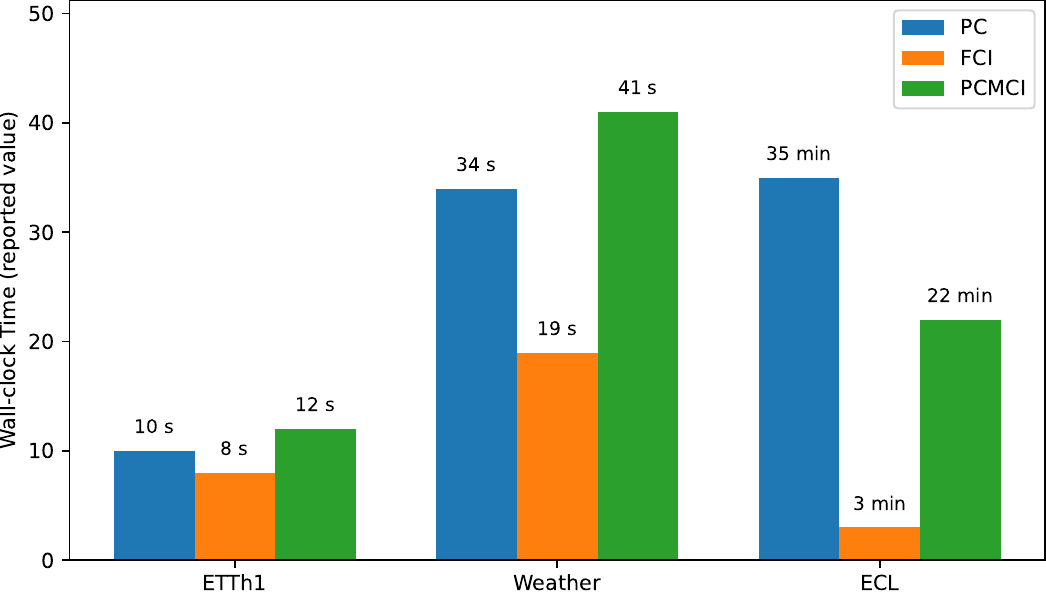}
    \caption{Wall-clock time of running different causal discovery algorithms on three representative datasets.}
    \label{fig:cd_time_appendix}
\end{figure}

\subsection{Runtime Analysis of Causal Discovery Methods}
\label{app:cd_runtime}
To complement the end-to-end efficiency results reported in the main text, we also measure the stand-alone runtime of common causal discovery algorithms on three representative datasets. Fig~\ref{fig:cd_time_appendix} reports the wall-clock time of running PC, FCI, and PCMCI on ETTh1, Weather, and ECL using our hardware. For small and medium-scale datasets such as ETTh1 and Weather, all methods finish within a few tens of seconds, making the preprocessing overhead negligible compared to model training. For larger and higher-dimensional datasets such as ECL, the runtime increases to several minutes, most notably for PC and PCMCI. However, this cost is incurred once as an offline preprocessing step and can be amortized over repeated training or deployment. Overall, the causal discovery overhead remains moderate in practice and does not limit the applicability of CDT.

\section{Forecasting Results Comparison}\label{compare}
\begin{figure*}[htpb]
\centering
    \subfloat[CDT for ETTh1]{
        \includegraphics[width=0.3\linewidth]{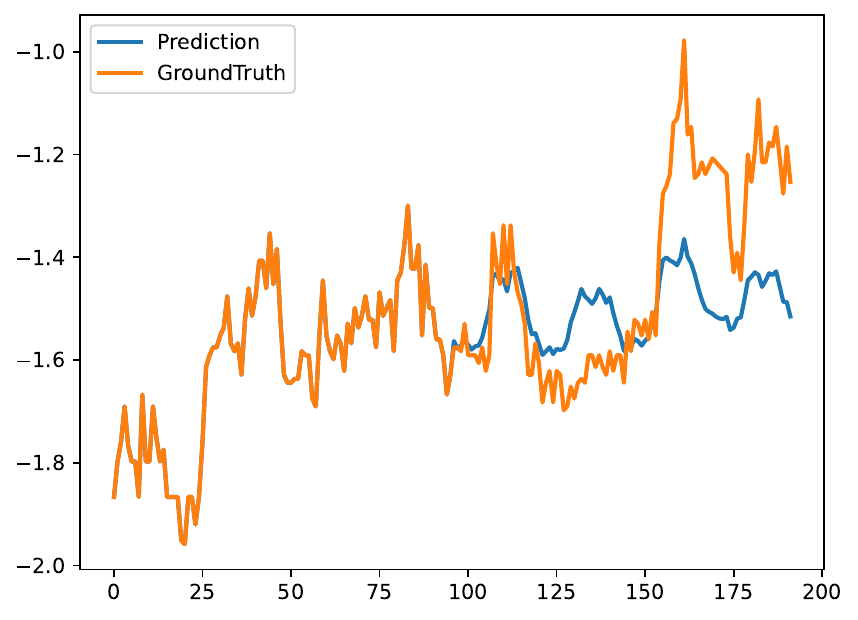}
    }
    % \hspace{-0.2in}
    \subfloat[iTransformer for ETTh1]{
        \includegraphics[width=0.3\linewidth]{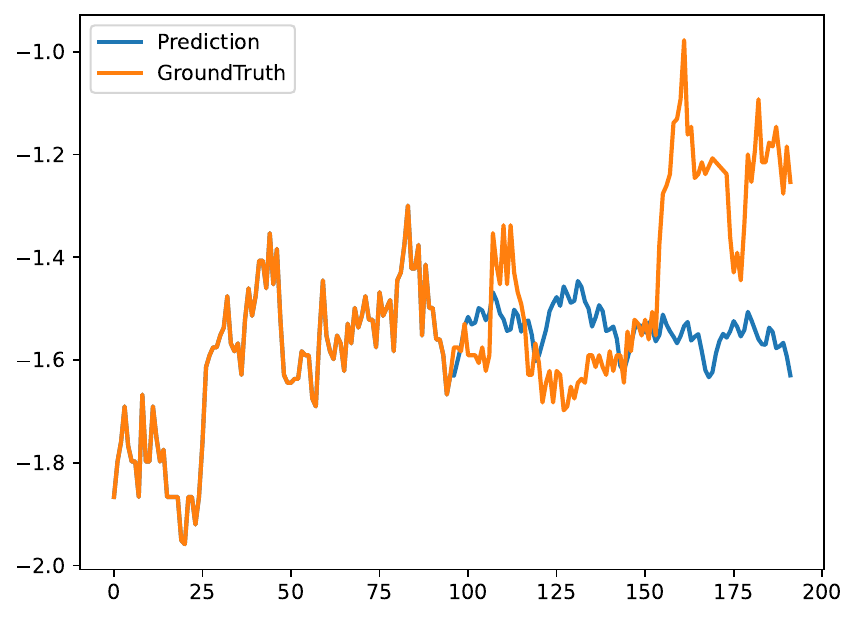}
    }
    % \hspace{-0.2in}
    \subfloat[PatchTST for ETTh1]{
        \includegraphics[width=0.3\linewidth]{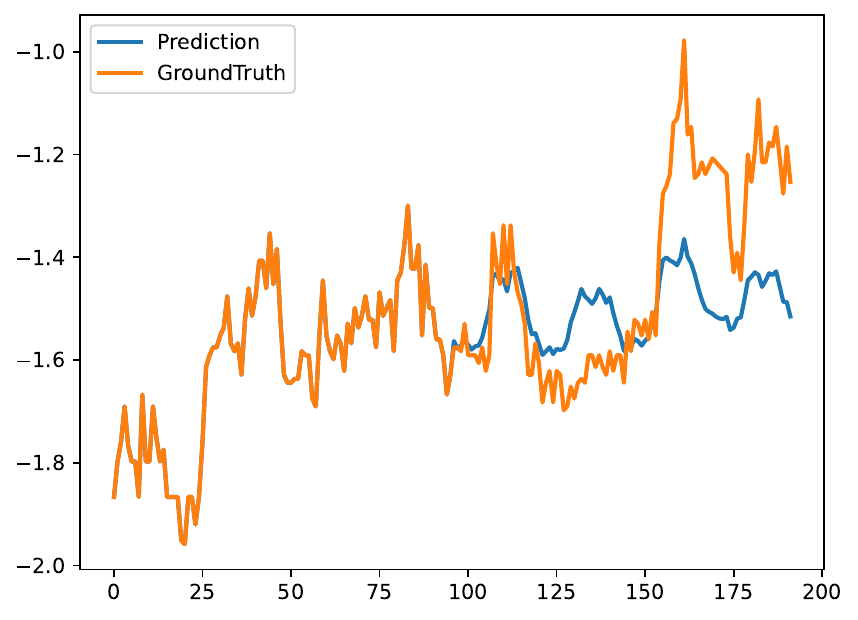}
    }
    \\
    \subfloat[CDT for ETTh2]{
        \includegraphics[width=0.3\linewidth]{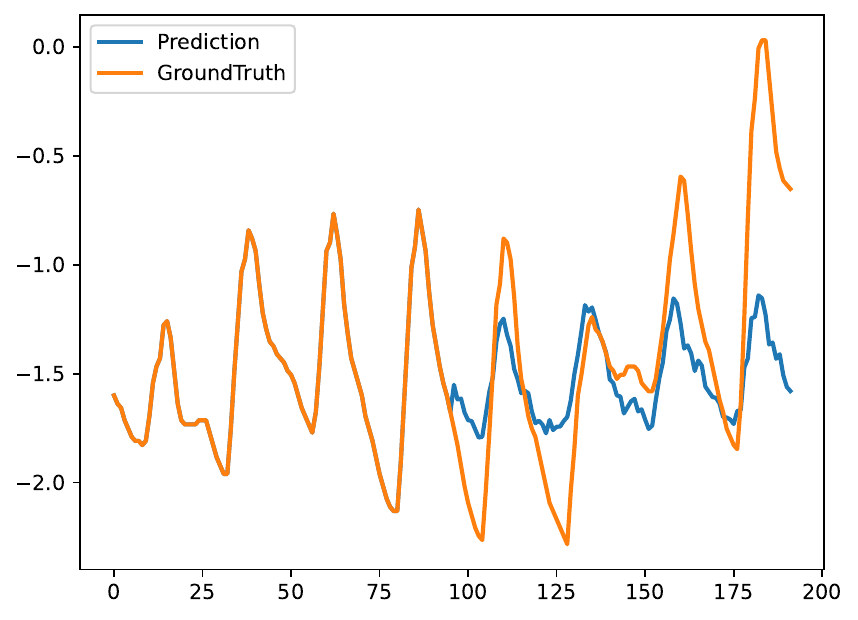}
    }
    % \hspace{-0.2in}
    \subfloat[iTransformer for ETTh2]{
        \includegraphics[width=0.3\linewidth]{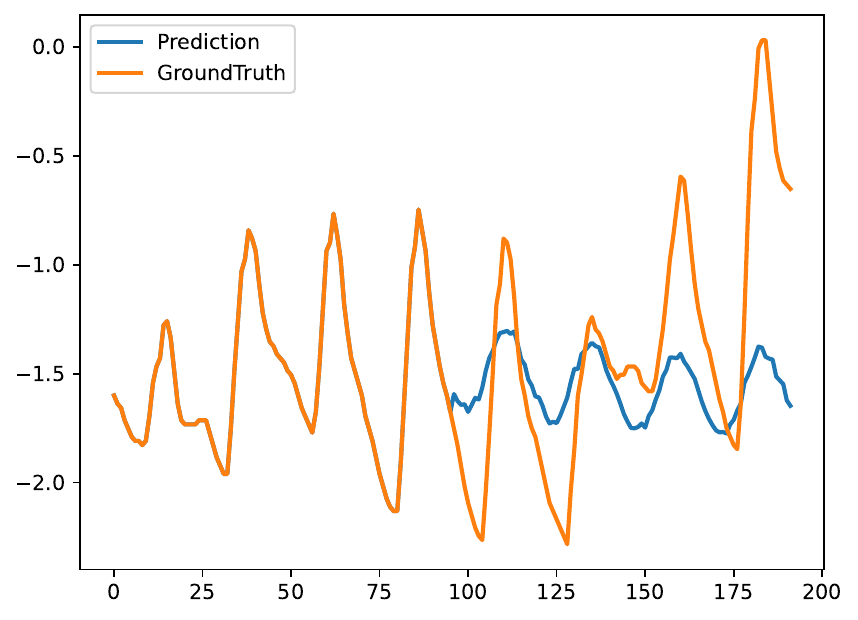}
    }
    % \hspace{-0.2in}
    \subfloat[PatchTST for ETTh2]{
        \includegraphics[width=0.3\linewidth]{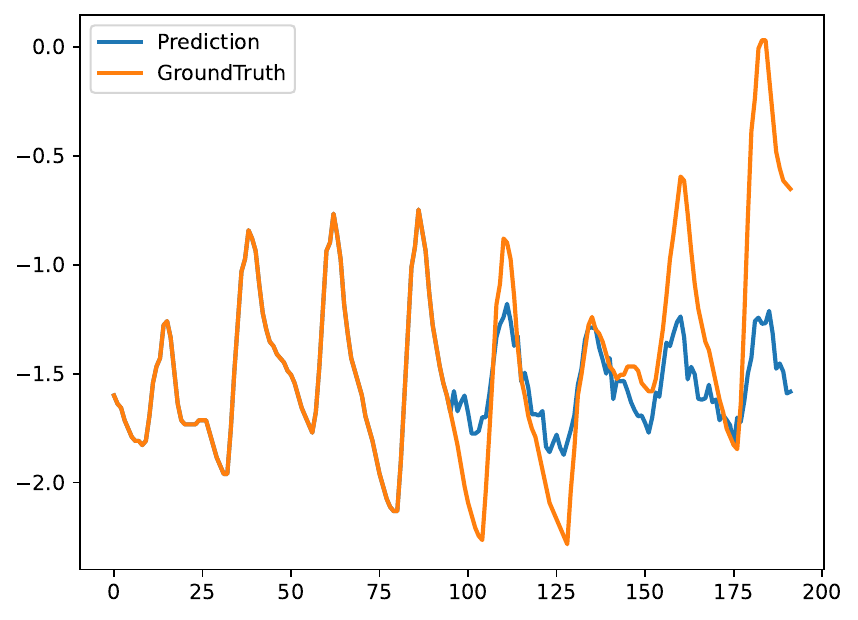}
    }
    \\
    \subfloat[CDT for Weather]{
        \includegraphics[width=0.3\linewidth]{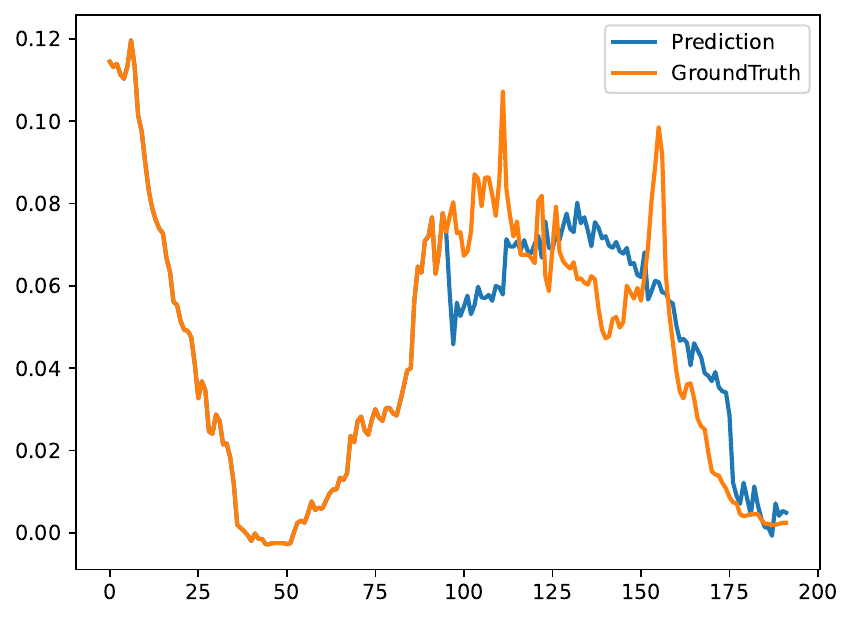}
    }
    % \hspace{-0.2in}
    \subfloat[iTransformer for Weather]{
        \includegraphics[width=0.3\linewidth]{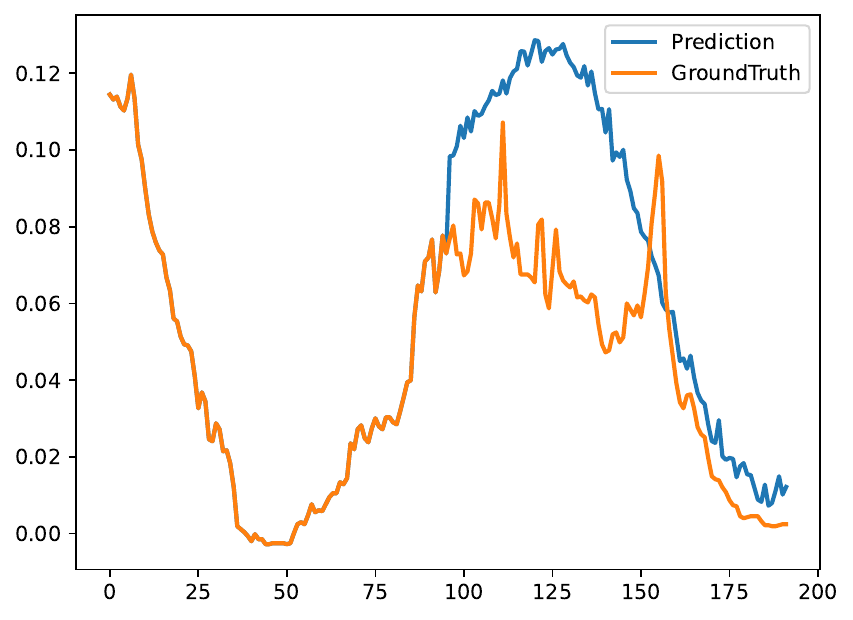}
    }
    % \hspace{-0.2in}
    \subfloat[PatchTST for Weather]{
        \includegraphics[width=0.3\linewidth]{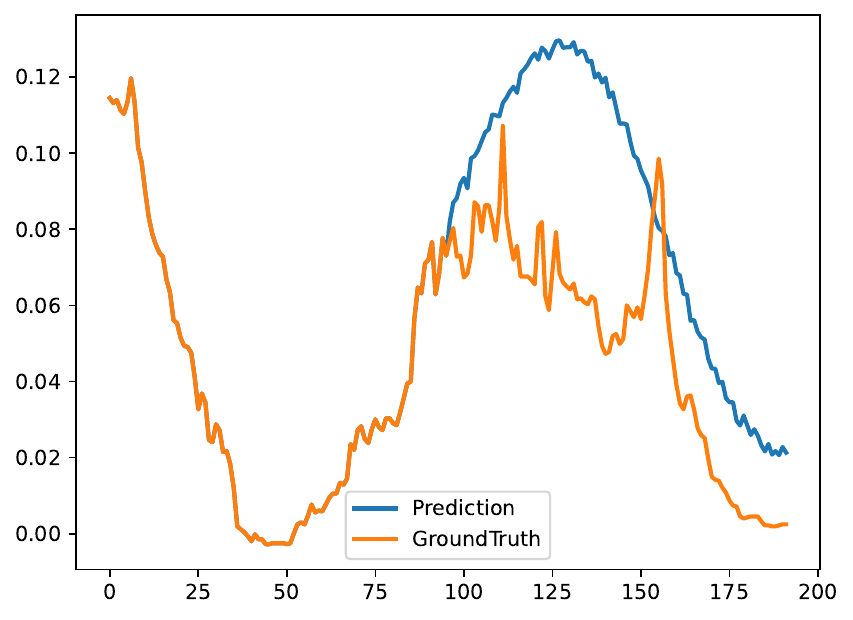}
    }
    \caption{Visualization of forecasting results for the ETTh1, ETTh2 and Weather dataset under the input-96-predict-96 setting.}
    \label{fig:Visualization—h2}
\end{figure*}

% We provide a visualization comparing the forecasting results of CDT (with lookback length 96 and prediction length 96) against iTransformer and PatchTST on the ETTh1, ETTh2, and Weather datasets, as depicted in Figure \ref{fig:Visualization—h2}.
Figure \ref{fig:Visualization—h2} compares CDT (input window 96, prediction 96) with iTransformer and PatchTST on ETTh1, ETTh2, and Weather.

\begin{figure}
   \centering
   \subfloat[\footnotesize ETTh2]{
    \includegraphics[width=0.3\linewidth]{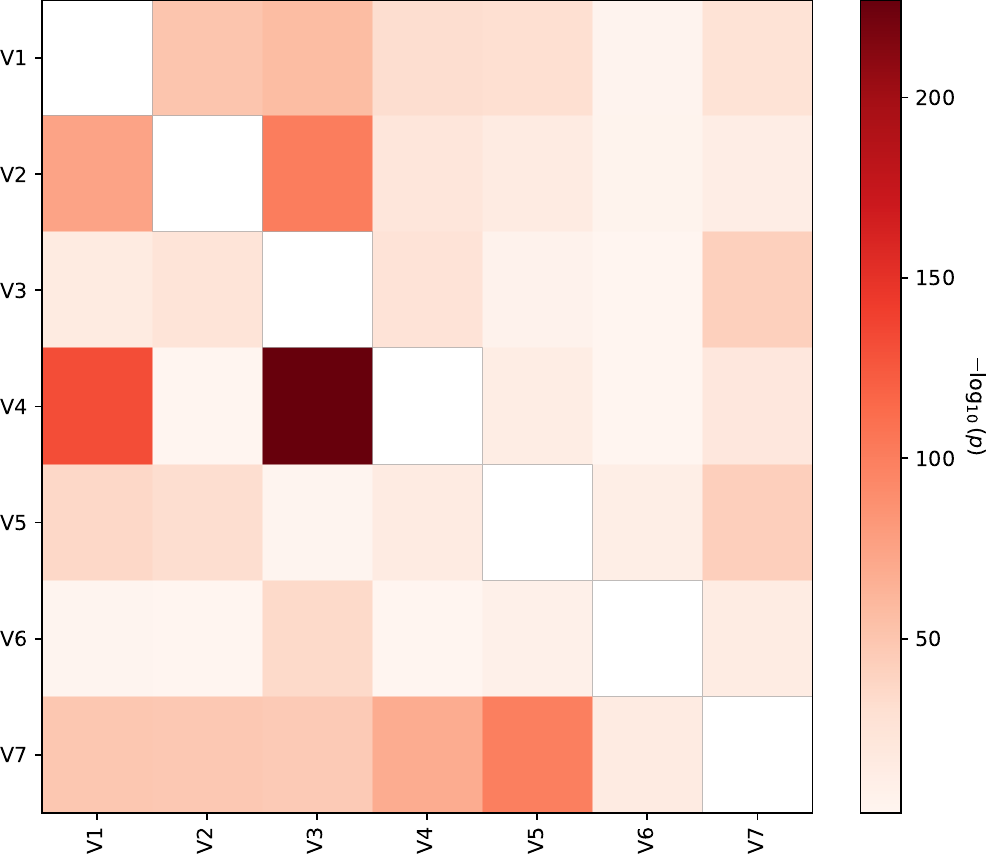}
    }
    \subfloat[\footnotesize ETTm2]{
        \includegraphics[width=0.3\linewidth]{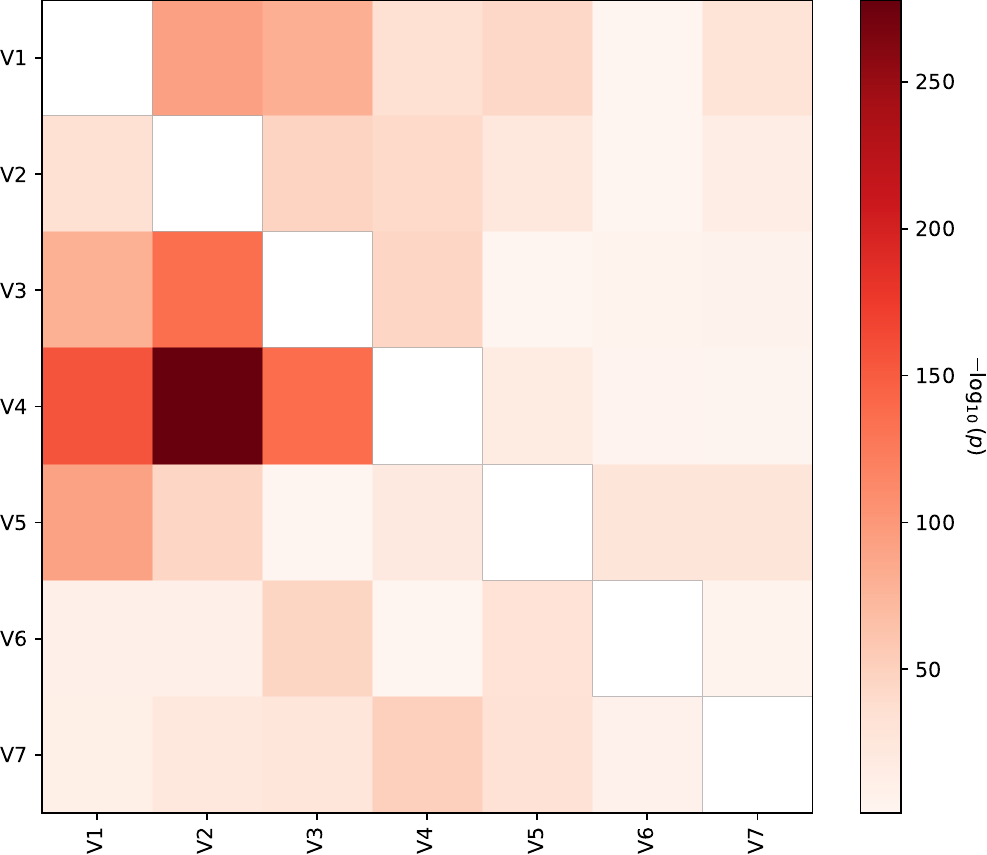}
    }
   \caption{Visualization of Granger causality across variables in ETTh2, ETTm2. Each heatmap shows the transformed causal strength matrix using $-\log(P)$ values, where a darker color indicates a stronger causal influence from the row variable to the column variable. Diagonal entries are masked.}
   \label{fig:motivation2}
\end{figure}

\end{document}